%% file: det6d.tex
\documentclass[lettersize,journal]{IEEEtran}
\usepackage{amsmath,amsfonts}
\usepackage{algorithmic}
\usepackage{algorithm}
\usepackage{array}
\ifCLASSOPTIONcompsoc
    \usepackage[caption=false, font=normalsize, labelfont=sf, textfont=sf,subrefformat=parens]{subfig}
\else
\usepackage[caption=false, font=footnotesize,subrefformat=parens]{subfig}
\fi
\usepackage{textcomp}
\usepackage{stfloats}
\usepackage{url}
\usepackage{verbatim}
\usepackage{graphicx}
\usepackage{cite}   
\usepackage{bbm}
\usepackage{autobreak}
\usepackage{multirow}
\usepackage[table]{xcolor}
\usepackage{tikz-imagelabels}

\makeatletter
\let\NAT@parse\undefined
\makeatother
\usepackage{hyperref}  
\def\etal\{\COMMAND{\textit{et al.}}

\begin{document}

\input{front}

\input{abstract}
\input{inctroduction}

\input{related_works}

\input{proposed}
\input{experiments}

\input{conclusion}

\bibliographystyle{IEEEtran}
\bibliography{det6d} 

\end{document}

%% file: front.tex
\title{Det6D: A Ground-Aware Full-Pose 3D Object Detector for Improving Terrain Robustness}

\author{Junyuan Ouyang, Haoyao Chen
\thanks{}
\thanks{J. Ouyang, and H. Chen* are with the School of Mechanical Engineering and Automation, Harbin Institute of Technology Shenzhen, P.R. China (e-mail: hychen5@hit.edu.cn).}
}

\markboth{Journal of \LaTeX\ Class Files, June~2022.}{}

\IEEEpubid{\scriptsize This work has been submitted to the IEEE for possible publication. Copyright may be transferred without notice, after which this version may no longer be accessible.}

\maketitle






%% file: abstract.tex
\begin{abstract}
Accurate 3D object detection with LiDAR is critical for autonomous driving. 
Existing research is all based on the flat-world assumption. 
However, the actual road can be complex with steep sections, which breaks the premise. 
Current methods suffer from performance degradation in this case due to difficulty correctly detecting objects on sloped terrain. 
In this work, we propose Det6D, the first full-degree-of-freedom 3D object detector without spatial and postural limitations, to improve terrain robustness. 
We choose the point-based framework by founding their capability of detecting objects in the entire spatial range. 
To predict full-degree poses, including pitch and roll, we design a ground-aware orientation branch that leverages the local ground constraints. 
Given the difficulty of long-tail non-flat scene data collection and 6D pose annotation, we present Slope-Aug, a data augmentation method for synthesizing non-flat terrain from existing datasets recorded in flat scenes. 
Experiments on various datasets demonstrate the effectiveness and robustness of our method in different terrains. 
We further conducted an extended experiment to explore how the network predicts the two extra poses. 
The proposed modules are plug-and-play for existing point-based frameworks. 
The code is available at \url{https://github.com/HITSZ-NRSL/De6D}.

\end{abstract}


%% file: inctroduction.tex
\section{Introduction}
\IEEEPARstart{I}{n} recent years, LiDAR sensors have been widely used for scene understanding in robotics and autonomous vehicles. 
3D object detection as an upstream task of these applications has attracted more researchers' attention. 
Most of them can be divided into grid-based methods \cite{pointpillars,voxelnet,yan2018second,yin2021center,deng2020voxel,zheng2020cia,ge2020afdet,hu2021afdetv2} and point-based methods \cite{qi2018frustum,qi2019deep,shi2019pointrcnn,yang2019std,yang20203dssd}. 
These methods are based on the flat-world assumption that objects are located on flat ground. 
Thus current approaches can only detect objects close to the xoy plane and their orientation along the gravity axis. 
Nevertheless, this assumption does not always hold in the real world, such as when driving through overpasses, underground parking lots, or other urban scenes with steep slopes. 
In these cases, objects have an extensive range of z-coordinates, and their pitch and roll can no longer be ignored. 
An example of failure cases is shown in Fig. \ref{fig:cover}, where previous methods suffer from wrong object detection and incorrect pose estimation. 
This kind of corner case should be considered, as severe performance degradation will lead to potential traffic accidents. 

In this work, we present a point-based one-stage detector to archive robust perception in complex terrains. 
The proposed method named Det6D is not dependent on the flat-world assumption, which fully considers the non-flat driving scene existing in the real world. 
It aims to improve the conventional 3D object detection, which is more like a 2.5D detection, to the full-space and full-pose 3D object detection. \IEEEpubidadjcol
\input{_data__general_view}

There are three main challenges. 
The {\bf first} is full-space 3D proposal generation, i.e., how to detect objects at any position in space to avoid missing objects on the slope or high places. 
This problem has not been considered previously because the assumption of existing methods limits the detection range to near the ground. 
For example, the classic voxel-based pipelines\cite{voxelnet,yan2018second} place anchor boxes and detect objects in bird's-eye view (BEV), ignoring height. 
A naive approach is to expand anchor boxes and voxelization range along height direction, but this makes the method less efficient. 
The {\bf second} is full pose prediction. 
Incorrect orientation prediction also causes the predicted object center to shift. 
However, all previous methods ignore the pitch and roll of objects. 
Note that full-pose prediction in this work is similar to 6D pose estimation. 
Methods in the 6D pose estimation task based on depth information can predict full pose well, typically capturing data with an RGB-D camera and having prior CAD models of objects. 
Unfortunately, these methods are unsuitable for detection due to the gaps in task domains, sensors, and real-time requirements. 
The {\bf third} challenge is the lack of data to train the network. 
Existing large-scale open datasets for autonomous driving are all recorded in flat scenes and have no 6D pose annotation of objects. 
Additionally, collecting long-tail sloped scenes and annotating 6D poses make it harder to construct a suitable dataset. 
Thus, the last problem to be solved is how to take advantage of a large number of existing datasets without much effort. 
It should be emphasized that the performance decline of previous methods is not only caused by the lack of similar non-flat scenes in the dataset but also because of inaccurate pose estimation. 

To address the above challenges, We first utilize the point-based framework and anchor-free center point generation module to detect objects without spatial limitation. 
Then the ground-aware orientation branch assists in full pose prediction by leveraging the results from the ground segmentation module. 
Finally, a new data augmentation method named Slope-Aug is proposed for non-flat scene generation and further alleviates the imbalanced distribution of pitch and roll. 

The main contributions are summarized as follows:
\begin{itemize}
  \item {
To the best of our knowledge, this work is the first attempt to consider a 6D pose prediction in 3D object detection, which is non-trivial. To improve the performance in complex terrains, we design Det6D with a ground-aware orientation branch to lift conventional detectors from 2.5D-limited into true 3D.}
 
  \item { 
The proposed data augmentation strategy, Slope-Aug, leverages existing datasets recorded in ordinary flat scenes to train the network, avoiding long-tail scene collection and laborious 6D pose annotations.}

  \item {
We benchmark our method against state-of-the-art methods in both flat and non-flat scenes datasets. 
The results show that our model maintains high performance on various terrains. 
Furthermore, extended experiments indicate that the network predicts object poses by learning the ground constraints on the pose.}
\end{itemize}

%% file: _data__general_view.tex
\begin{figure}[!t]
\begin{minipage}{0.65\linewidth}
    \subfloat[]{
    	\label{fig:cover:a}
    	\begin{minipage}{\linewidth}
    			\includegraphics[width=\linewidth]{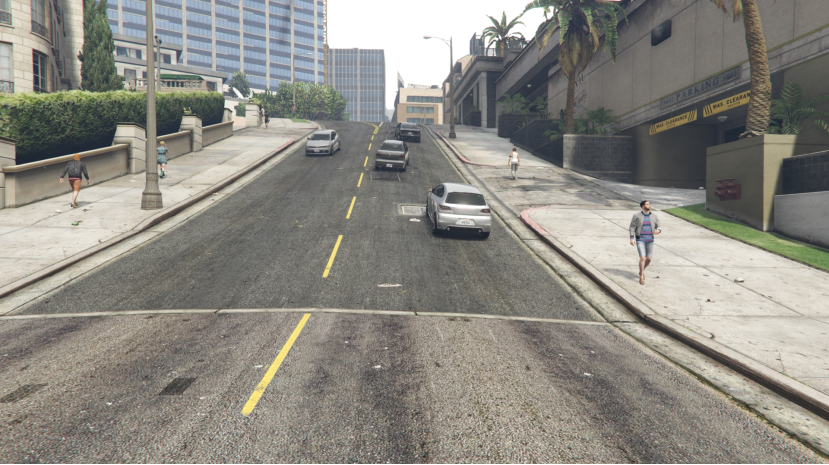}\vspace{7pt}
                \begin{annotationimage}{width=\linewidth}{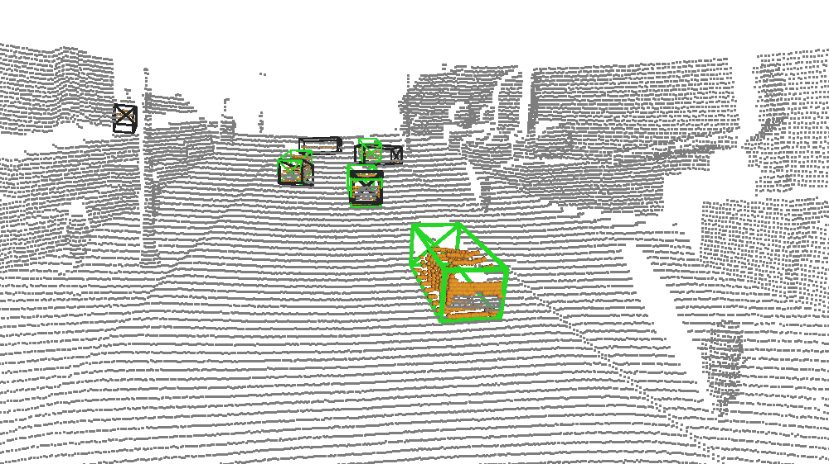}
                \imagelabelset{
                    border thickness = 0.0pt,
                    arrow thickness = 0.4pt,
                    tip size = 0.8mm,
                    outer dist = -1.5cm,
                    }
                    \draw[annotation right = {(d) at 0.45}] to (0.6,0.45);
                    \imagelabelset{ outer dist = -1.5cm}
                    \draw[annotation left = {(c) at 0.55}] to (0.34,0.65);
                    \imagelabelset{ outer dist = -0.6cm}
                    \draw[annotation above = {(b) at 0.38}] to (0.38,0.69);
                \end{annotationimage}
    	\end{minipage}
    }
\end{minipage}
\begin{minipage}{0.31\linewidth}
	\subfloat[]{
		\label{fig:cover:b}
		\includegraphics[width=\linewidth]{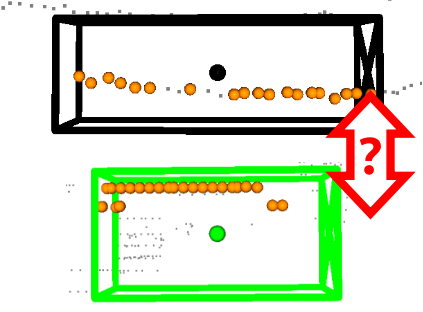}
	}\vfil\vspace{-6pt}
	\subfloat[]{
		\label{fig:cover:c}
		\includegraphics[width=\linewidth]{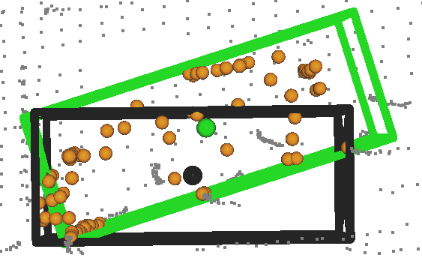}
	}\vfil\vspace{-6pt}
	\subfloat[]{
		\label{fig:cover:d}
		\includegraphics[width=\linewidth]{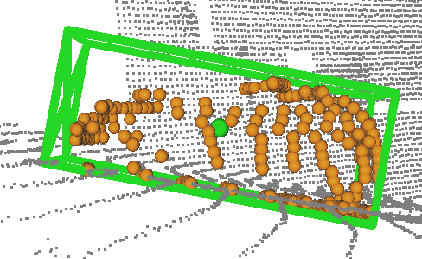}
	}
\end{minipage}

\caption{
A common non-flat urban scene. 
Some failure cases of the previous method are shown in the right column. 
The green and black boxes indicate the results of our method and previous method PointRCNN \cite{shi2019pointrcnn}, respectively.
(a) Scene overview.
(b) Regard the scan lines on slope as the side of a vehicle. 
(c) Incorrect position and pose prediction. 
(d) Fail to detect the car. 
}

\label{fig:cover}
\end{figure}

%% file: related_works.tex
\section{RELATED WORKS}
\subsection{Point Cloud 3D Object Detection}
The purpose of existing point cloud 3D object detection methods is to predict objects' position, dimension, and heading from point clouds. 
According to the data representation of the unordered 3D point cloud, the current 3D object detector can be mainly divided into grid-based methods and point-based methods. 
Some works \cite{chen2017mv3d,zhou2020mvf,fan2021rangedet} project clouds into front view, BEV, or others in grid-based methods and then are processed by highly optimized 2D CNNs. 
Other methods divide space into pillars \cite{pointpillars} or voxels \cite{voxelnet,yan2018second,shi2020pv,deng2020voxel,zheng2020cia,yin2021center,ge2020afdet}, called voxel-based methods. 
Zhou \textit{et al.} \cite{voxelnet} used learned voxel features instead of artificial ones. 
Yan \textit{et al.} \cite{yan2018second} stacked sparse convolution and submanifold convolution \cite{graham2017submanifold} for efficient feature extraction by utilizing the sparsity of point cloud. 
Most voxel-based methods compressed the feature volume along the z-axis into a BEV feature map. 
And then, an anchor-based \cite{voxelnet,yan2018second,shi2020pv,deng2020voxel,zheng2020cia} or anchor-free \cite{yin2021center,ge2020afdet,hu2021afdetv2} detection head is employed to predict boxes. 

The point-based methods use points directly as input without any representation conversion. 
The pioneering work \cite{qi2017pointnet} learned order-invariant global features from points set via multi-layer perceptron (MLP) and channel-wise max pooling. 
PointNet++ \cite{qi2017pointnet++} is a hierarchical network that can learn local and global features from points. 
As an fundamental module in point-based detectors, it is widely used to extract semantic features in various networks \cite{qi2018frustum,qi2019deep,shi2019pointrcnn}. 
Shi \textit{et al.} \cite{shi2019pointrcnn} first proposed a bottom-up proposal generator to get proposals from points directly.
Yang \textit{et al.} \cite{yang20203dssd} removes the feature propagation layer to reduce the time consumption. 
Chen \textit{et al.} \cite{chen2022sasa} introduces learnable parameters in the downsampling layer to retain more foreground points adaptively. 

\subsection{Object Pose Estimation from Depth Information }
These methods \cite{guo2021efficient,chen2020pointposenet,gao20206d,he2020pvn3d,hagelskjaer2020pointvotenet,gao2021cloudaae} aim to recover the position and orientation of objects from depth information captured by RGB-D cameras. 
Since CAD models are known, there is no need to predict the dimension of objects. 
Guo \textit{et al.} \cite{guo2021efficient} proposed a method based on point pair features (PPF) to estimate object 6D poses from points in the depth map. 
Gao \textit{et al.} \cite{gao20206d,gao2021cloudaae} utilized two network branches to estimate position and orientation and proposed a novel geodesic loss function for more accurate rotation regression. 
Other voting-based \cite{he2020pvn3d} and keypoint-based \cite{chen2020pointposenet} methods fed the point clouds into a PointNet-like network to predict key points. 
They recovered 6D poses by estimating the transformation between predicted keypoint sets and predefined point sets. 
Most of these methods perform iterated closest point (ICP) based on the known CAD model to refine the estimation results further. 

In the detection task, existing methods only consider predicting the horizontal orientation, i.e., yaw, from a point cloud. 
Zhou \textit{et al.} \cite{voxelnet} directly regresses the residual between the proposal and its corresponding anchor box. 
Yan \textit{et al.} \cite{voxelnet} encodes the yaw into sin-cos code for better regression. 
Other works \cite{shi2019pointrcnn,qi2018frustum,yang20203dssd} divide the yaw into several bins, then predict which bin it belonged to and the residual in the bind.

\subsection{Limitations of Existing Methods }
Existing detection methods assume the world to be flat, resulting in only detecting objects near the xoy plane and only predicting yaw. 
Projection-based methods have no detection range limitation but lower accuracy. 
While expanding the detection range in voxel-based methods will lead to less efficiency. 
The 6D pose estimation methods based on depth information cannot be used to predict full pose in the detection task. 
These methods utilize the depth data from the RGB-D camera, which is dense and small-scale. 
While in the detection task, the point cloud from LiDAR is non-uniform, sparse, and large-scale. 
Moreover, in the 6D pose estimation task, networks can only handle an instance point cloud obtained by image segmentation at a time and involve time-consuming post processes. 
Thus multiple objects need to be processed sequentially. 
However, the detection task needs to process multiple objects quickly in parallel. 
Furthermore, these 6D pose estimates methods get accurate predictions because the CAD models are known, which are not available in the detection task\looseness=-1.


%% file: proposed.tex
\input{_data__pipeline}
\section{PROPOSED APPROACH}
In this section, we first analyze the method selection, introduce our networks, and then describe the differences between yaw, pitch, and roll in pose prediction. 
The overall pipeline is illustrated in Fig. \ref{fig:pipeline}. 
It consists of three main parts: 1) a center point generator for full-space proposal generation; 2) a multi-task detection head with the ground-aware branch for full orientation prediction; 3) an augmentation method Slope-Aug for synthesizing non-flat scenes.
\subsection{Problem Definition and Method Analysis}
We denote $\mathcal{P}=\left\{\boldsymbol{p}_i\right\}_{i=1}^n$ as the input point clouds, where each point $\boldsymbol{p}_i=\left(x_i,y_i,z_i,\ldots\right) \in\mathbb{R}^{3+c}$ consists of three coordinates and optional features like intensity. 
Existing methods parameterize the bounding box of detected object with $\boldsymbol{b}_{2.5\text{D}}=\left(x_c,y_c,z_c,l,w,h,\theta_z\right)$, which stands for position, dimension, and yaw, respectively. 
Nevertheless, the $z_c$ of $\boldsymbol{b}_{2.5\text{D}}$ is limited to a small value, i.e., near the ground. 
To describe the object in non-flat scenes, we denote the full-pose bounding box by $\boldsymbol{b}_{3\text{D}}=(x_c,y_c,z_c,l,w,h,\theta_x,\theta_y,\theta_z)$, where $z_c$ is unconstrained and $(\theta_x,\theta_y,\theta_z)$ are Euler angles in x-y-z order. 
The aim of this work is to detect objects in $\mathcal{P}$ represented by $\boldsymbol{b}_{3\text{D}}$ instead of $\boldsymbol{b}_{2.5\text{D}}$ used in previous works.

{\bf Method Analysis.} We explore what limits current methods to detect objects with full pose in the entire space. 
In voxel-based methods, the voxelization range of the z-axis tends to be no more than 3m, while the height of target objects in non-flat scenes can be up to 10m, causing points outside the range to be ignored. 
A simple approach is to expand the z-axis range, but this greatly increases the time and memory consumption. 
In addition, since non-flat scenes are corner scenes, such a large z range is meaningless in most cases and results in lots of empty voxels. 
Although sparse convolution alleviates this problem, the sparse features need to be converted into a dense presentation and compressed along the z-axis to obtain BEV feature maps before dense detection. 
This causes the dimension of BEV features to grow and introduces many zero values in features. 
Anchor boxes with different categories, sizes, and orientations need to be placed at each location in the BEV maps for anchor-based methods. 
Even for flat scenes, regardless of pitch, roll, and z, the number of anchors can reach a million. Thus anchor-based methods are not feasible for our task\looseness=-1. 

\subsection{Feature Encoding and Center Point Generation}
Based on the above analysis, for efficient feature encoding and full-space full-pose object detection in non-flat scenes, we design a point-based backbone and an anchor-free proposal generator similar to previous works \cite{yang20203dssd,chen2022sasa}. 

{\bf Point Backbone.} 
The backbone only remains the encoder part of the U-shaped network \cite{qi2017pointnet++} to reduce overhead. 
The input points $\mathcal{P}$ are gradually aggregated into a small set of representative points $\mathcal{P}_r=\left\{\boldsymbol{p}_i\right\}_{i=1}^{n_1}$, each with semantic feature $\boldsymbol{f}_r\in\mathbb{R}^{c_1}$ by multiple furthest point sampling (FPS) layers and multi-scale grouping set abstraction (MSG-SA in \cite{qi2017pointnet++}) layers. 
The SA layer first groups sampled points and then extracts features at each grouping center by PointNet\cite{qi2017pointnet} as:
\begin{equation}
\label{pointnet}
\boldsymbol{f}=\boldsymbol{\gamma}\left(\underset{i=1, \ldots, k}{\textrm{MAX}}\left\{\boldsymbol{h}\left(\boldsymbol{x}_{i}\right)\right\}\right)
\end{equation}
where $\boldsymbol{f}$ is the extracted feature from grouped points $\{\boldsymbol{x}_i\}_{i=1}^k$ and $\boldsymbol{\gamma}\left(\cdot\right)$, $\boldsymbol{h}\left(\cdot\right)$ are multilayer perceptron. 
In order to preserve more foreground points in downsampling for improving recall, the distance metric of FPS is weighted by feature distance, foreground score, and local point density. 

{\bf Center Point Generator.} 
The offset from each point in $\mathcal{P}_r$ to the center of its corresponding object is predicted and used to obtain coarse centers $\mathcal{P}_{cc}$ (the golden balls shown in Fig. \ref{fig:pipeline}) for further refinement. 
To extract the feature $\boldsymbol{f}_{cc}\in\mathbb{R}^{c_2}$ of each coarse center, an SA layer is also applied on $\mathcal{P}_{cc}$. 

\subsection{Box Prediction via Full Pose Detection Head}
Our detection head consists of multiple branches to predict the attributes of objects corresponding to coarse centers $\mathcal{P}_{cc}$ with their features $\mathcal{F}_{cc}=\left\{ \boldsymbol{f}_{cc} \right\}$. 
Concretely, these branches predict the class label $c$, offset to fine center $(\Delta{x},\Delta{y},\Delta{z})$, dimensions $(l,w,h)$, and orientations $(\theta_{x},\theta_{y},\theta_{z})$ of objects. 
Finally, apply non-maximum suppression to remove redundancy\looseness=-1.

{\bf Ground-aware Orientation Head.} 
Unlike existing methods, our orientation head predicts both the yaw~$\theta_z$ and the two previously ignored poses, i.e., pitch~$\theta_y$ and roll~$\theta_x$. 
It is worth noting that this extension is non-trivial. 
Since the data distribution difference, copying the method for yaw to these two poses does not work. 
For better explanation, Fig. \ref{fig:distribution} shows the distribution of object attributes in SlopedKITTI containing 6D pose annotations. 
Both $(l,w,h)$ and $\theta_z$ approximately obey the mixed Gaussian distribution, which makes them easy to be regressed. 
However, despite this dataset containing up to 10\% of non-flat scenes, higher than most cases, the points with zero $\theta_x$ and $\theta_y$ are still 3\~{}4 orders of magnitude more than the non-zero ones. 
If we directly predict $\theta_x$ and $\theta_y$ like $\theta_z$, the results tend to be zero since the highly imbalanced distribution leads to non-zero values being ignored like noise. 
Furthermore, it is intuitive from the BEV that the shape of points varies with $\theta_z$. 
While as the extended experiment indicated, the shape changes imperceptibly with $\theta_x$ and $\theta_y$, making the network hard to capture the explicit relationship between shape and these two poses. 
The above reasons explain why the prediction methods for different poses are not the same.

\input{_data__distribution}
A novel view is that objects in autonomous driving scenes all lie on the ground, which well constrains $\theta_x$ and $\theta_y$. 
Thus, awareness of the ground can help the regression of these two poses. 
With this prior, we introduce a lightweight ground segmentation module composed of only several full connection layers. 
It classifies the terrain of each point located into two categories: flat and sloped, as represented by blue and red balls in Fig. \ref{fig:pipeline}, respectively. 
Specifically, we feed the feature $\boldsymbol{f}_{cc}$ of each coarse center in $\mathcal{P}_{cc}$ to the segmentation module and predict the probability $s_g$ of being on sloped terrain via
\begin{equation}
\label{ground_segmentation}
s_g={\textrm{Sigmoid}}({\textrm{MLPs}}(\boldsymbol{f}_{cc})).
\end{equation}
By doing this, the unbalanced regression problem is decomposed into two well-addressed problems: unbalanced classification and balanced regression. 
We use focal loss \cite{lin2017focal} with the default setting to handle this imbalanced classification since there are fewer points belonging to sloped terrain. 
And smooth-L1 loss is applied to supervise the pose predictions ${\theta}_x$ and ${\theta}_y$ for points on sloped terrain only. 
We denote $\boldsymbol{b}_{3\text{D}}^g=(x_c^g,y_c^g,z_c^g,l^g,w^g,h^g,\theta_x ^g,\theta_y^g,\theta_z^g)$ as a ground-truth box. 
The ground segmentation label $c_g$ and the pose regression targets $\theta_x^t$ and $\theta_y^t$ of each coarse center can be generated naturally from ground-truth bounding boxes by
\begin{align}
\label{target_calculation}
\begin{split}
&c_g = \mathbbm{1}( \ \theta_x^g \ge t_{\theta_x}  \text{or} \ \theta_y^g\ge t_{\theta_x})\\
&\theta_x^t =  \cfrac{\theta_x^g-t_{\theta_x}}{\pi/2} ,\
\theta_y^t =  \cfrac{\theta_y^g-t_{\theta_y}}{\pi/2}, \\
\end{split}
\end{align}
where $\mathbbm{1}$ is the indicator function, and $t_{\theta_x}$ and $t_{\theta_y}$ are the thresholds used to determine the terrain category. 
The orientation branch then utilizes the segmentation results to assist the network in recognizing the latent relationship between pose and terrain, as detailed in Fig.  \ref{fig:branch_details}. 
Specifically, we obtain the final prediction ${\theta}_x$ and ${\theta}_y$ by associating with $s_g$, as 
\begin{equation}
\label{pitch_and_roll}
{{\theta}_a} = \begin{cases}
\theta^p_a,&{\text{if}}\ {s_g}>0.5 \\ 
{0,}&{\text{otherwise.}} 
\end{cases}, \ \text{for} \ a \in \{x,y\},
\end{equation}
where $\theta^p_a$ is decoded from branch predicted output $\hat{\theta}_a$ via \eqref{target_calculation}. 
\input{_data__branch_details}
For heading $\theta_z$, we divide it into $N_{{\theta}_z}$ discrete bins. Then classify which bin it belongs to with a cross-entropy loss and regress the residual within this bin with a smooth-L1 loss. 
The bin label $c_{\theta_z}$ and the residual in a bin $\Delta\theta_z^t$ for heading prediction are calculated as in previous works \cite{shi2019pointrcnn} by
\begin{align}
\label{target_calculation_yaw}
\begin{split}
&c_{\theta_z} =  \lfloor \cfrac{\theta_z^g}{\delta_{\theta_z}} \rfloor \\
&\Delta\theta_z^t =  \frac{\theta_z-c_{\theta_z}\delta_{\theta_z}+\frac{\delta_{\theta_z}}{2}}{\delta_{\theta_z}},
\end{split}
\end{align}
where $\delta_{\theta_{z}}=2\pi/N_{\theta_z}$ represents the size of each bin in radian and $\lfloor \cdot \rfloor$ means floor operation to get an integer value.

{\bf Position Head.} 
For more accurate object position prediction, we regress the offset $\Delta{\boldsymbol{p_{cc}}}$ of each coarse center in $\mathcal{P}_{cc}$ to its actual center with smooth-L1  loss.

{\bf Dimension Head.} 
We also regress the dimension of objects.
The dimensions are log-mapped, and the results are used as the regression target $u^t=\log u$, where $u\in\{l,w,h\}$.

{\bf Classification Head.} 
Benefiting from the downsampling strategy \cite{chen2022sasa}, the categories distribution of coarse centers is more balanced. 
Thus, we directly apply cross-entropy to calculate the classification loss from prediction $\hat{c}$ and label $c$. 

{\bf Loss.} 
The overall training loss of bounding boxes $\left\{\boldsymbol{b}_{3\text{D}}\right\}$ consists of different branch loss terms, as
\begin{align}
\label{loss_function}
\begin{split}
&L_{\theta_{x,y}} = \cfrac{1}{N_p}\sum\mathbbm{1}(c>0)\mathcal{L}_{seg}(s_g,c_g)+ \\&\ \ \ \ \ \ \ + \cfrac{1}{N_{s}}\sum_{a\in \{x,y\}}\mathbbm{1}(c_g>0)\mathcal{L}_{reg}(\hat{\theta}_a,\theta_a^t)\\
&L_{\theta_{z}} = \sum \cfrac{\mathbbm{1}(c>0)}{N_{p}}[\mathcal{L}_{cls}(\hat{c}_{\theta_z},c_{\theta_z}) + \mathcal{L}_{reg}(\Delta\hat{\theta}_z,\Delta\theta_z^t)]\\
&L_{box} = L_{cls} + L_{dim}+L_{posi} + L_{\theta_{x,y}} + L_{\theta_{z}},
\end{split}
\end{align}
where the normalization factor $N_{p}$ is the number of foreground points in $\mathcal{P}_{cc}$, and $N_s$ is the number of foreground points in $P_{cc}$ that lie on sloped terrain. 
$\mathbbm{1}(c_g>0)$ indicates that only foreground points on sloped terrain contribute to the loss item. 
The smooth-L1 is utilized for regression $L_{reg}$. $\mathcal{L}_{seg}$ is focal loss for terrain segmentation and $\mathcal{L}_{cls}$ is cross-entropy. 

\subsection{Slope Augmentation to Help Full Pose Learning}
\input{_data__slope_aug}
The existing open datasets \cite{kitti,nuscenes,waymo} contain only flat scenes and lack the required $(\theta_x,\theta_y)$ annotations. 
Moreover, it is laborious to collect data from non-flat scenes and manually annotate 3D bounding boxes with full pose in 3D space. 
Therefore, Slope-Aug is proposed to address this problem by synthesizing random slope scenes in the input point cloud and generating full pose annotations during the training phase. 

Specifically, we perform the steps shown in Fig.  \ref{fig_slope_aug} with probability $p_s$ for each input. 
We first select a anchor point $\boldsymbol{\tau}=(r,\alpha,0)$ in cylinder coordinates. 
The tangent direction of this point is the desired rotation axis $\boldsymbol{v}$. 
The input point cloud $\mathcal{P}$ is divided into two parts by the anchor $\boldsymbol{\tau}$ and the axis $\boldsymbol{v}$, where the part containing the origin is denoted as $\mathcal{P}_1$, and the other part $\mathcal{P}_2$ can be formulated as
\begin{equation}
\label{slope-aug-divide}
\mathcal{P}_2=\left\{
\begin{array}{l|l}
&\\
\boldsymbol{p}_i \\
&
\end{array} 
\begin{array}{l}
\boldsymbol{\tau}^{\text{T}}(\boldsymbol{\tau}-\boldsymbol{p}_i)<0, \\
\ \ \ \ \  \forall \boldsymbol{p}_{i} \in \mathcal{P}
\end{array}\right\}.
\end{equation}
To simulate the points on slope $\tilde{\mathcal{P}}_2$, we rotate $\mathcal{P}_2$ by the axis angle $(\boldsymbol{v},\gamma)$ at $\boldsymbol{\tau}$ and then obtain pseudo-slope points $\mathcal{P}_s=\mathcal{P}_1\cap\tilde{\mathcal{P}}_2$\looseness=-1. 
Full-pose bounding box annotations $\mathcal{B}_{3\text{D}}$ are generated simultaneously from $\left\{\boldsymbol{b}_{2.5\text{D}}\right\}$ as
\begin{equation}
\label{slope-aug-boxes}
\mathcal{B}_{3\text{D}}=
\left\{
	\begin{array}{l|l}
    	&\\
        \Big [\boldsymbol{b}_{i,:7},\boldsymbol{r},\boldsymbol{b}_{i,7}\Big ]\\
        &
    \end{array}
    \begin{array}{l}
        \boldsymbol{b}_i \in \left\{\boldsymbol{b}_{2.5\text{D}}\right\},\\
        c_i=\boldsymbol{\tau}^{\text{T}}(\boldsymbol{\tau}-\boldsymbol{b}_{i,:3})<0,\\
        \boldsymbol{r}= 
        c_i\textrm{ToEular}_{xy}(\boldsymbol{v},\gamma)\\ 
    \end{array}
\right\},
\end{equation}
where the second subscript of $\boldsymbol{b}_{i,[\cdot]}$ means the slice operation in vector $\boldsymbol{b}_{i}$, i.e., to obtain elements of a vector.

In this way, we can leverage $\mathcal{P}$ and $\{\boldsymbol{b}_{2.5\text{D}}\}$ from existing datasets to synthesize slope scene $\mathcal{P}_s$ and generate full-pose box annotations $\mathcal{B}_{3\text{D}}$. 
Despite the domain gaps between synthetic and natural data,
the following experiments show that the models trained with these pseudo-slope scenes can be generalized to actual scenes. 
Moreover, the highly unbalanced distribution of yaw and roll is alleviated by this augmentation method.

%% file: _data__pipeline.tex
\begin{figure*}[!t]
\centering
\includegraphics[width=0.975\textwidth]{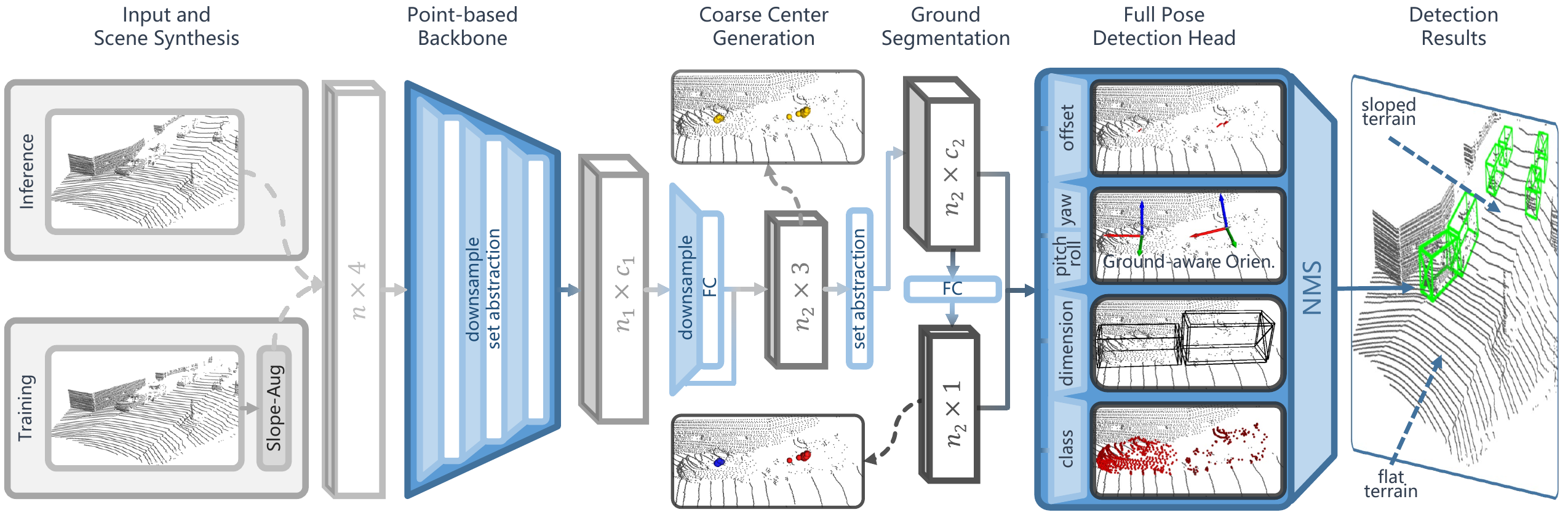}
\caption{
The overall framework of Det6D. 
In the training phase, the raw point cloud is randomly synthesized as a sloped scene via Slope-Aug, while it is fed directly into the network in the test phase. 
The point cloud backbone downsamples the input and extracts multi-scale semantic features. 
These semantic features are used to generate a little set of coarse center points of objects. 
And then, the center points and their features are segmented into different classes by a ground segmentation module. 
Finally, the detection head combines the ground segmentation results and the features of center points to predict boxes with full poses. 
Best view in colors.}
\label{fig:pipeline}
\end{figure*}

%% file: _data__distribution.tex
\begin{figure}[!t]
	\centering
	\includegraphics[width=1.0\linewidth]{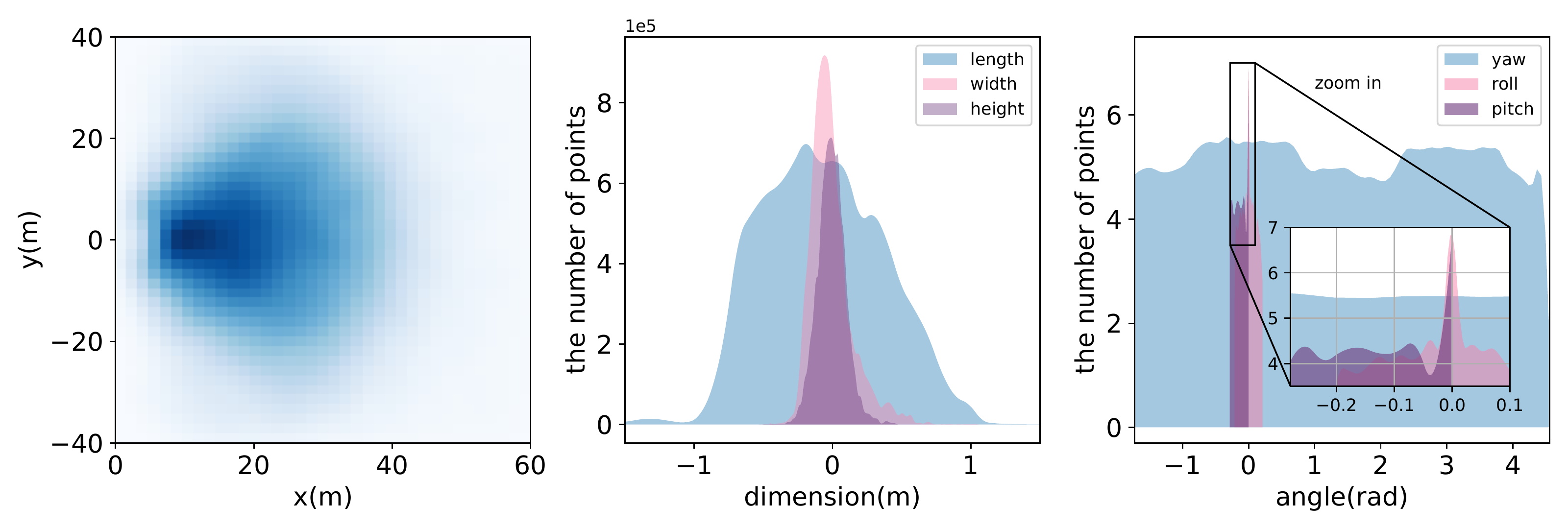}
\caption{ Distribution of object attributes in the SlopedKITTI dataset. 
(L) Distribution of object centers in BEV. 
(M) Distribution of object dimension. 
(R) Distribution of object poses (log-10 scale).}
\label{fig:distribution}
\end{figure}

%% file: _data__branch_details.tex
\begin{figure}[!t]
	\centering
	\includegraphics[width=1.0\linewidth]{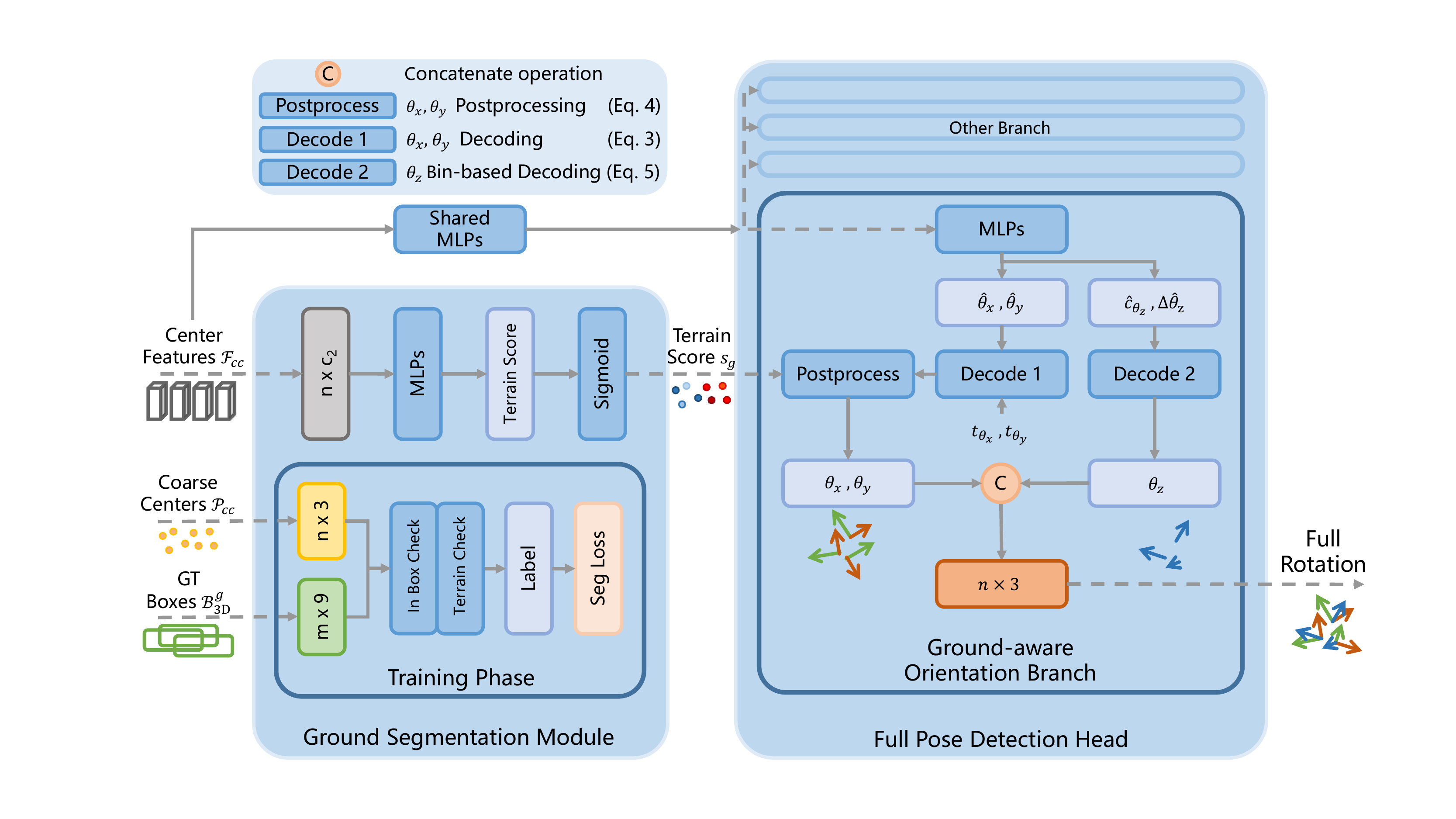}
    \caption{  
    Architectures of the ground segmentation module and ground-aware orientation branch. 
    The ground segmentation module predicts which classes of terrains the coarse centers are located via ground-truth boxes supervising.
    The features of coarse centers are fed into shared MLPs followed by orientation branches.
    The orientation branch predicts bin index $\hat{c}_{\theta_{z}}$, residual $\Delta\hat{\theta}_{z}$ for yaw, and $\hat{\theta}_{x}$, $\hat{\theta}_{y}$ for pitch and roll.
    Then we recover the normalized $\hat{\theta}_{x}$ and $\hat{\theta}_{y}$ to $\theta_{x}^p$ and $\theta_{y}^p$ by equation \eqref{target_calculation}, and $\hat{c}_{\theta_{z}}$ and  $\Delta\hat{\theta}_{z}$ to $\theta_{x}$ by bin-based methods \eqref{target_calculation_yaw}, respectively.
    Additionally, the ground segmentation scores are utilized to post-process $\theta_{x}^p$ and $\theta_{y}^p$ via equation \eqref{pitch_and_roll}.
    Finally, these three predicted orientations are concatenated and treated as output.
    }
    \label{fig:branch_details}
\end{figure}

%% file: _data__slope_aug.tex
\begin{figure}
    \centering
    \subfloat[]{%
        \includegraphics[width=0.23\linewidth]{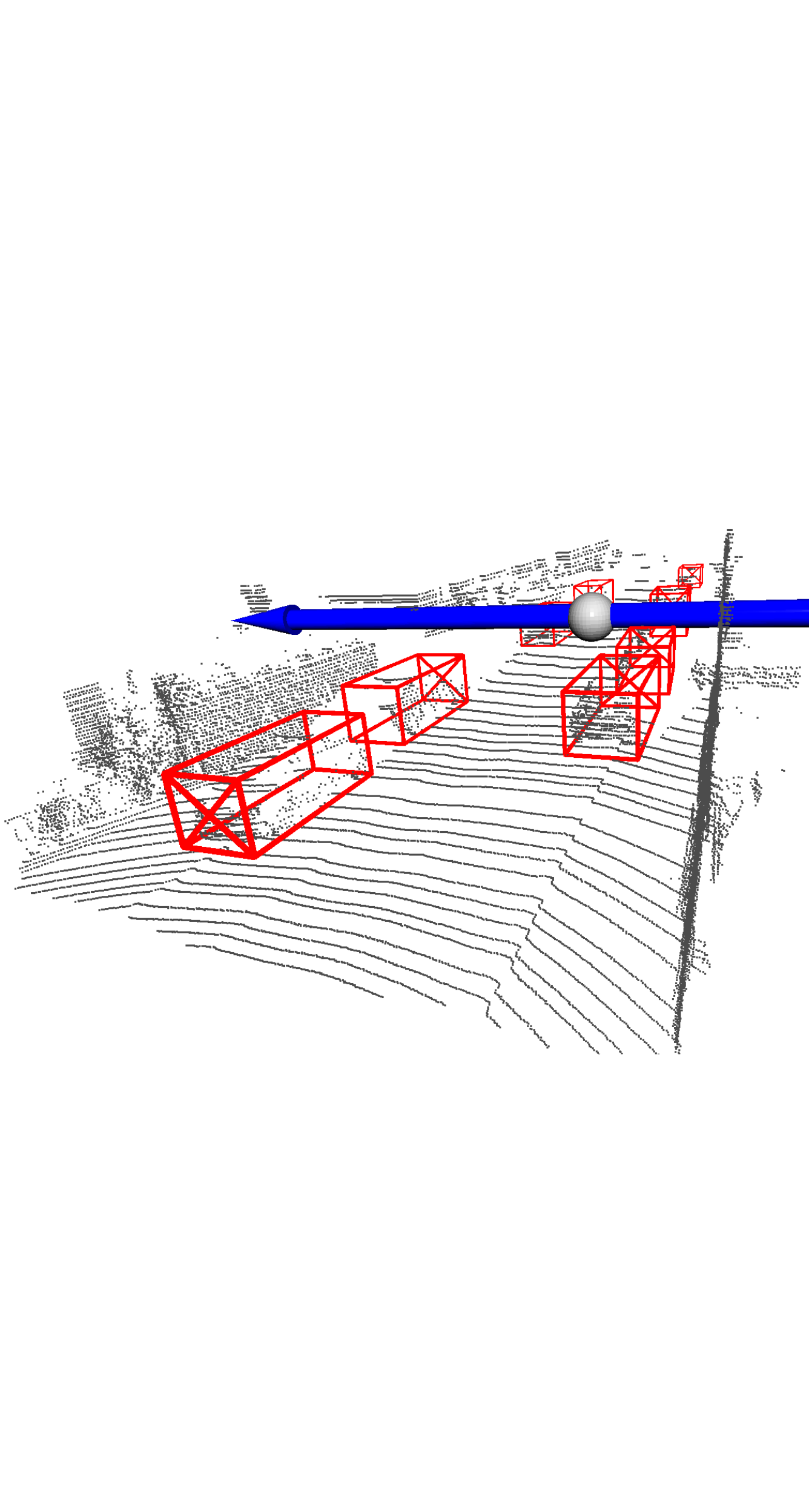}
    } 
    \subfloat[]{%
        \includegraphics[width=0.23\linewidth]{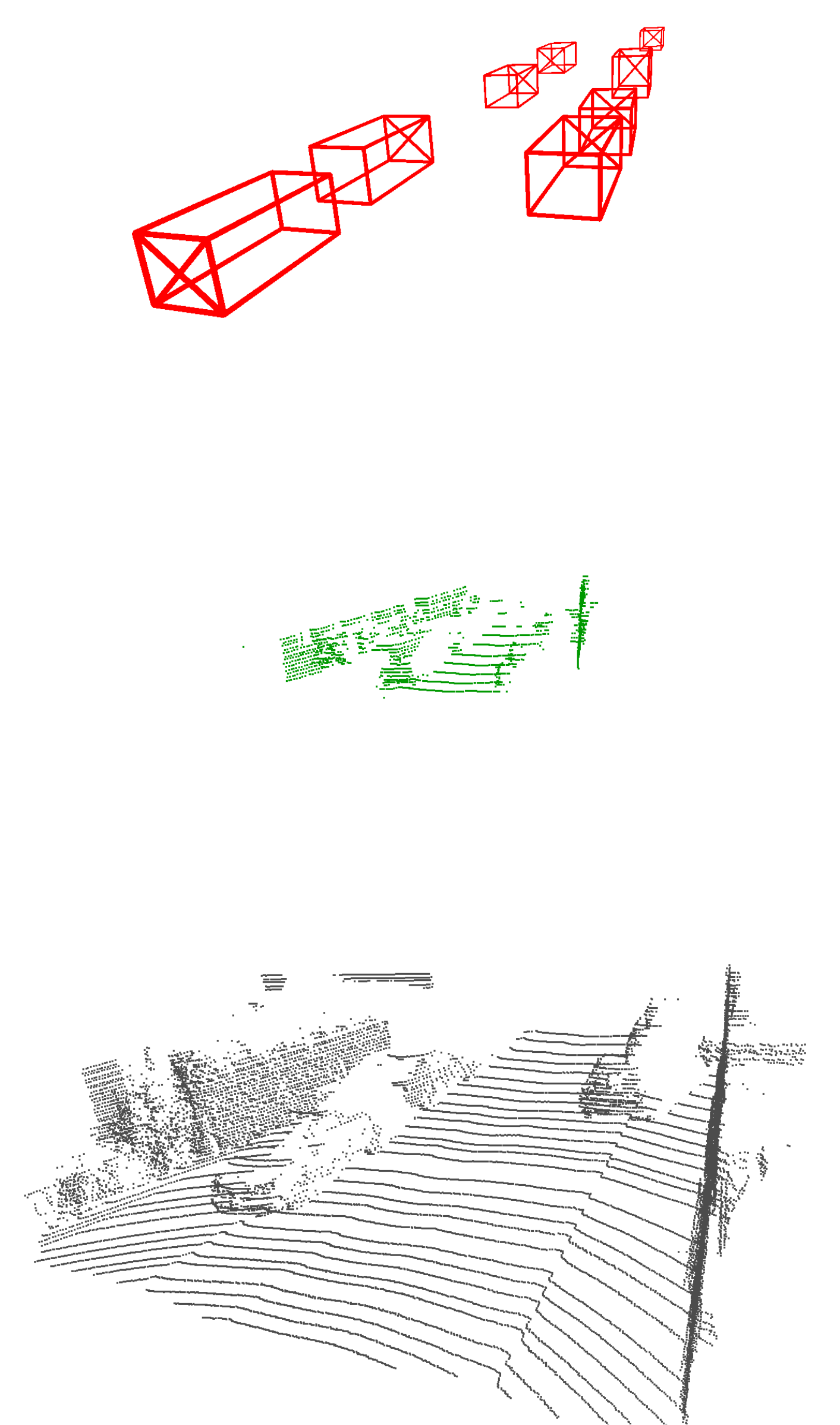} 
    } 
    \subfloat[]{%
        \includegraphics[width=0.23\linewidth]{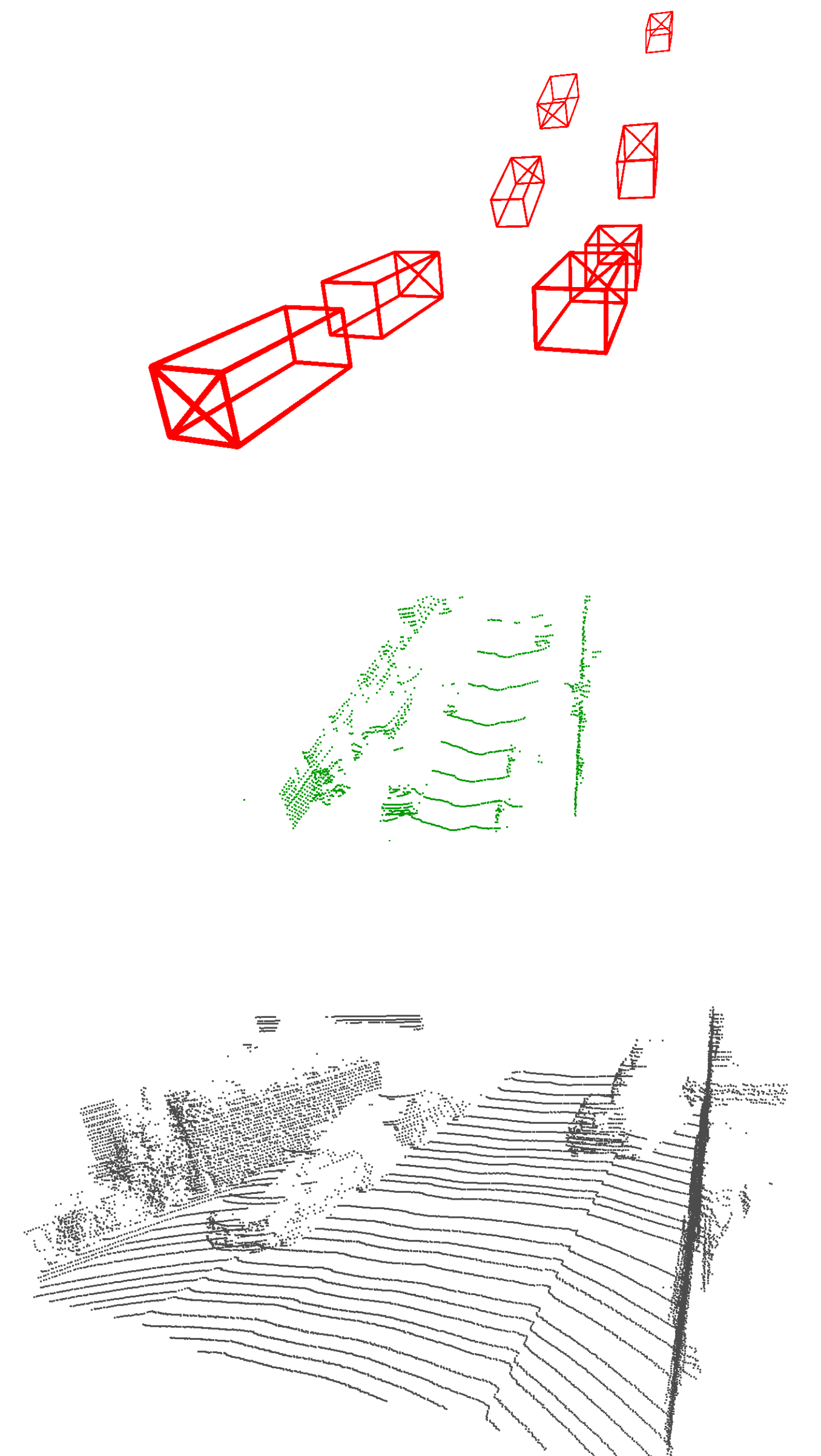} 
    } 
    \subfloat[]{%
        \includegraphics[width=0.23\linewidth]{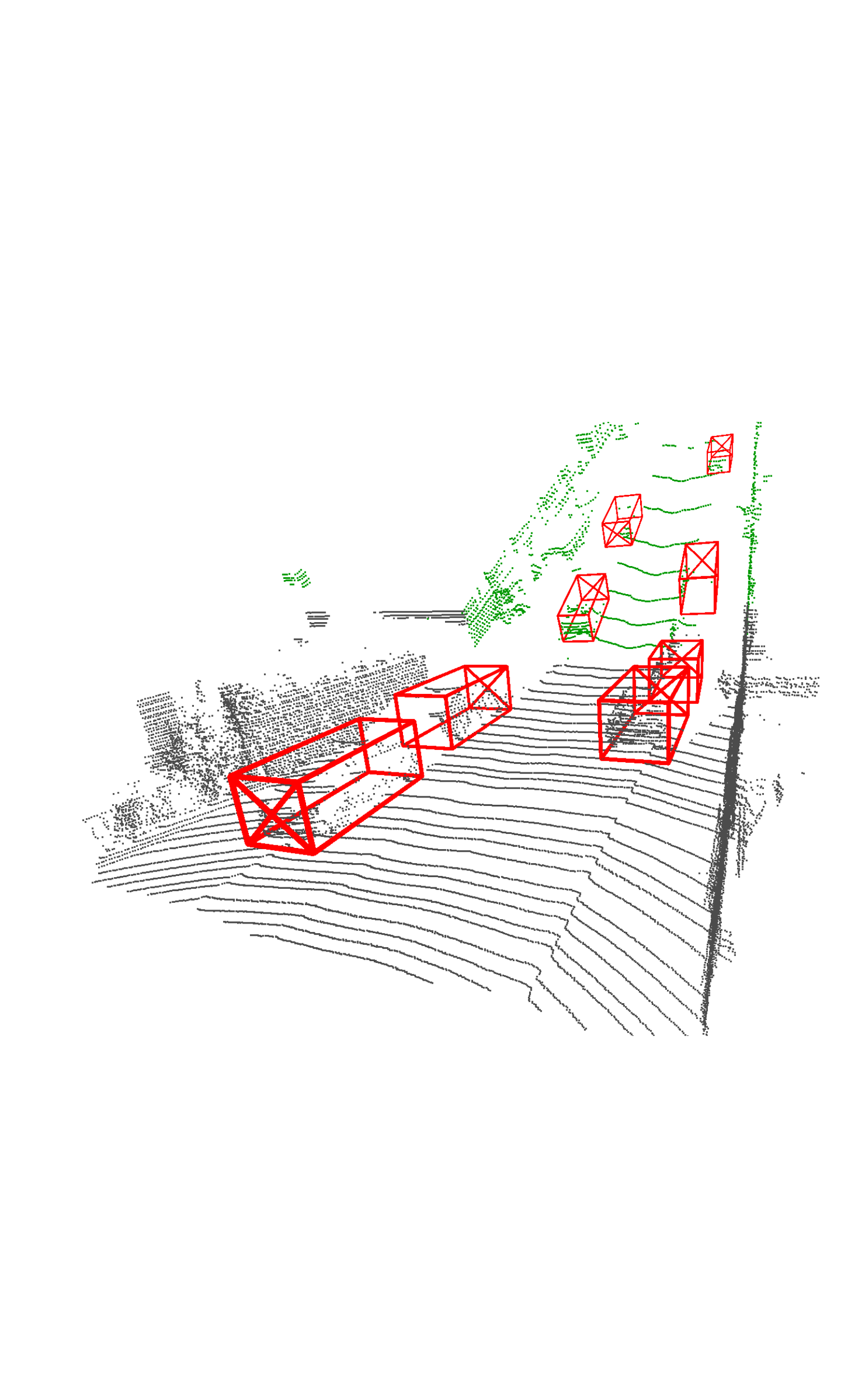} 
    }

\caption{
Overview of the Slope-Aug procedure. 
(a) Pick a point $\boldsymbol{\tau}$ (grey ball) and rotation axis $\boldsymbol{v}$ (blue arrow) randomly.
(b) Divide the input points $\mathcal{P}$ into $\mathcal{P}_1$ and $\mathcal{P}_2$ (green points). 
(c) Rotate $\mathcal{P}_2$ and ground-truth boxes. 
(d) Merge the two parts of points and generate box annotations with full poses.
}
\label{fig_slope_aug}
\end{figure} 

%% file: experiments.tex
\section{EXPERIMENTS}
\input{_exp__fig_slopedkitti_small}

\subsection{Dataset}
\subsubsection{KITTI}
The competitive 3D object detection dataset KITTI \cite{kitti} is used to evaluate the performance of Det6D in {\bf flat scenes}. 
It contains 7481 training and 7518 test samples collected with $\left\{\boldsymbol{b}_{2.5\text{D}}\right\}$ annotated. 
We divide the training set into \textit{train} split and \textit{val} split with 3712 and 3769 samples, respectively. 
For submission to the KITTI test server, we refer to strategy \cite{shi2020pv} to generate the \textit{train} split.
\subsubsection{SlopedKITTI}
Since existing open datasets do not contain non-flat scenes, we build SlopedKITTI based on KITTI \textit{val} split by synthesizing pseudo-slope. It contains slope and full pose annotations and is used to validate the performance of our method in {\bf non-flat scenes}. 
\subsubsection{GTA-V and GAZEBO}
Although SlopedKITTI contains non-flat urban scenes, it is similar to the original KITTI with Slope-Aug applied.
For fair comparison and to demonstrate the generalization performance, we also test Det6D on two different environments. 
GTA-V is a commercial video game with realistic urban scenes in which we collect real non-flat scenes.
And GAZEBO is a common robot simulation software used for additional experiments.  
\input{_exp__table_slopedkitti}
\input{_exp__table_kitti}

\subsection{Metrics}
The same as official KITTI, all results are evaluated by average precision (AP), which is calculated using 11 recall positions on \textit{val} split and 40 positions on \textit{test} set. 
For vehicle detection, we use an intersection over union (IoU) threshold of 0.7 to determine the true positive (TP). 
Rotated BEV IoU and rotated BEV 3D IoU are used for BEV and 3D detection. 

Since the IoU used above is calculated based on BEV, the relationship between the two full-pose bounding boxes in 3D space cannot be correctly described. 
Inspired by NDS\cite{nuscenes}, we provide an additional rotated 3D metric (R3DM) to reflect the performance in 3D space accurately. 
It replaces the IoU threshold of 0.7 with the center distance of 1.0m, and the corresponding AP is denoted as $\text{AP}_{cd}$. 
For TPs, we additionally compute the average translation score (ATS), average scale score (ASS), and average full orientation score (AOS) from their corresponding error item. 
The composite rotated object detection score (RODS) is computed by  
\begin{equation}
\label{OSD}
\text{RODS}=\frac{1}{6} (3\text{AP}_{cd}+\text{ATS}+\text{ASS}+\text{AOS}).
\end{equation}
\subsection{Implementation Details}
Our Det6D is implemented based on OpenPCDet\cite{openpcdet2020} toolbox and runs on single GTX2080Ti GPU. 
\subsubsection{Network Architecture}
We first randomly sample 16384 points for each raw point cloud and shuffle them as input. 
In the backbone, we gradually sample (4096, 1024, 512) points using three downsampling layers. 
The output feature channels of each SA layer are 64, 128, and 256, respectively. 
Then we select $n_2=256$ points from the $n_1=512$ representative points obtained by the last downsampling layer as the initial center points and feed their features into $\text{MLPs}$(256,128,3) to predict offsets. 
The detection head contains a shared $\text{MLPs}$(512,256). 
The parameter $N_{\theta_z}$ is 12 and the threshold $t_{\theta_x}$ and $t_{\theta_y}$ are $10^{\circ}$. 
During the testing phase, we use the IoU threshold of 0.1 for non-maximum suppression (NMS).

\subsubsection{Training}
The model was trained on KITTI \textit{train} split for 7 hours using a batch size of 4 and a one-cycle turning strategy with a 0.1 learning rate.
We first use GT-Aug\cite{yan2018second} to randomly copy some foreground objects from other scenes into the current scene and then use the proposed Slope-Aug with probability $p_s=0.1$.
Other strategies have also been applied, such as global flipping, scaling, and rotation.
Except for testing on the \textit{test} set, models for all experiments were trained on \textit{train} split and then validated on \textit{val} split. 

\input{_exp__table_ablation}
\subsection{Main Results} 
To evaluate the performance in non-flat scenes, we quantitatively compare the proposed method with state-of-the-art. 
All methods are trained on the same KITTI \textit{train} split and validated on the SlopedKITTI \textit{val} split. 
Some results are visualized in Fig. \ref{fig:slopekitti_viz}.
As shown in Table \ref{tab:table1}, the proposed method outperforms all previous methods by a large margin on all metrics in sloped scenes.  
For 3D object detection, Det6D dramatically improves the easy, moderate, and hard APs by 25.76\%, 34.0\%, and 33.26\%, respectively.  
For the most concerned rotated 3D object detection, we lead by 12.76\%, 1.17\%, 0.04\%, 1.36\%, and 11.54\% in $\text{AP}_{cd}$, each TP score and RODS, respectively. 

It can be found in Table \ref{tab:table1} that point-based methods outperform voxel-based methods in BEV and $\text{AP}_{cd}$ because of the capability of detecting objects anywhere the point exists, i.e., no detection range limitation. 
While in terms of TP scores, previous point-based methods perform worse than voxel-based methods. 
It is mainly because all TPs on sloped terrains are detected with incorrect poses by previous point-based methods. 
Thus, except for ours, point-based methods fall behind voxel-based methods in each TP score. 
Nevertheless, all these methods perform poorly in 3D AP since full pose prediction is not considered, resulting in low IoU. 
In addition, it is worth noting that the previous point-based methods detected lots of false positives on the slope because the scan lines from slopes are similar to those on the side of vehicles. 
While the proposed method not only correctly detects objects on slopes and some unlabeled vehicles but also accurately predicts all poses.

\input{_exp__fig_performance_drop}
To evaluate performance in flat scenes, we also report results on KITTI as shown in Table \ref{tab:table2}. 
The results show that our method performs similarly to other point-based methods in flat scenes.  
Comparing the results in Table \ref{tab:table1} and Table \ref{tab:table2} shows that existing methods have a significant performance drop in non-flat scenes.
We further report the results on SlopedKITTI with different slopes to explore the effect of terrain on detector performance. 
As reflected in Fig. \ref{fig:performance_drop}, existing methods drop up to 45\% performance in scenes with a $20^\circ$ slope, while our method has a relatively slight drop of 15\%.
We correctly detect objects with full pose in non-flat scenes, improving perception robustness in autonomous driving. 
 
\subsection{Effectiveness}
\input{_exp__fig_gtav}
To verify the effectiveness of the proposed method, we test these methods on GTA-V and GAZEBO, which contain actual sloped terrain, for qualitative comparison. 
All models are trained on KITTI, and some results are visualized in Fig. \ref{fig:gtav_viz}. 
Consistent with previous conclusions, voxel-based methods cannot detect objects on slopes, and previous point-based methods cannot accurately detect objects on slopes. 
In contrast, the proposed method can detect objects correctly and predict full poses accurately.
It proves that the model trained by Slope-Aug can effectively handle real slopes. 
\input{_exp__fig_gazebo2}
{\bf What did the network learn?} 
We designed an extended experiment in GAZEBO to verify our conjecture that the predictions of pitch and roll depend on the ground while yaw depends on the object itself. 
When the vehicle is on a slope (Fig. \subref*{fig:exp_gazebo_2:1}), regardless of the actual poses, the predicted pose is consistent with the local ground normal. 
When the vehicle is on flat ground (Fig. \subref*{fig:exp_gazebo_2:2}), the predicted poses variate with the local terrain. 
The above experimental results demonstrate that the proposed ground-aware orientation branch accurately predicts pitch and roll with an awareness of the ground, which is inherently different from yaw prediction.

\subsection{Ablation Studies}
The result in Table \ref{tab:ablation_study} shows how each module affects the performance.  
The first two rows show that Sloed-Aug improves detection performance in non-flat scenes significantly. 
D.R. and G.O.B are different methods for predicting pitch and roll. 
The $\text{2}^{\text{nd}}$ row and $\text{3}^{\text{rd}}$ row show that direct regression has some effects.  
Pitch and roll errors decrease while yaw errors increase. 
Since the vehicle is usually a cuboid, yaw has a much more impact on IoU than pitch and roll, which is why the $\text{3}^{\text{rd}}$ row has 10.70\% less 3D AP than the $\text{2}^{\text{nd}}$ row. 
The last two rows indicate that the proposed ground-aware pose head can better predict the full poses than direct regression. 
In all configurations, the last row achieves the best performance, reflecting the effectiveness of the proposed framework.


%% file: _exp__fig_slopedkitti_small.tex
\begin{figure}
    \centering\hspace{-10pt}
    \subfloat{%
        \setcounter{subfigure}{0}
        \begin{minipage}[]{0.01\linewidth}
            \rotatebox{90}{\tiny SlopedKITTI}
        \end{minipage}
    }
    \subfloat[]{%
        \begin{minipage}[]{0.18\linewidth}
        \includegraphics[width=1\linewidth]{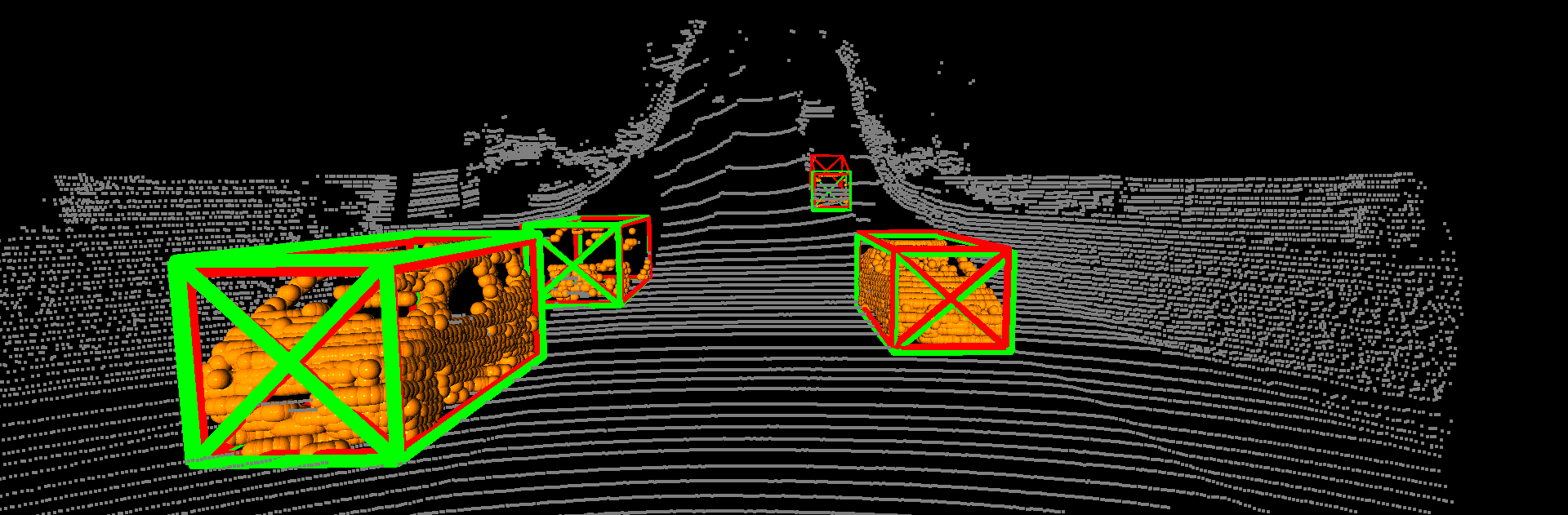}\vspace{-4.5pt}
        \includegraphics[width=1\linewidth]{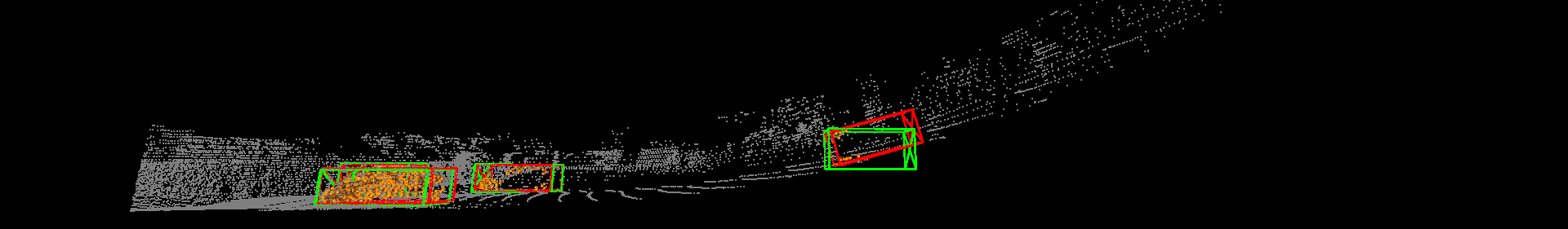}\vspace{2pt}
        \includegraphics[width=1\linewidth]{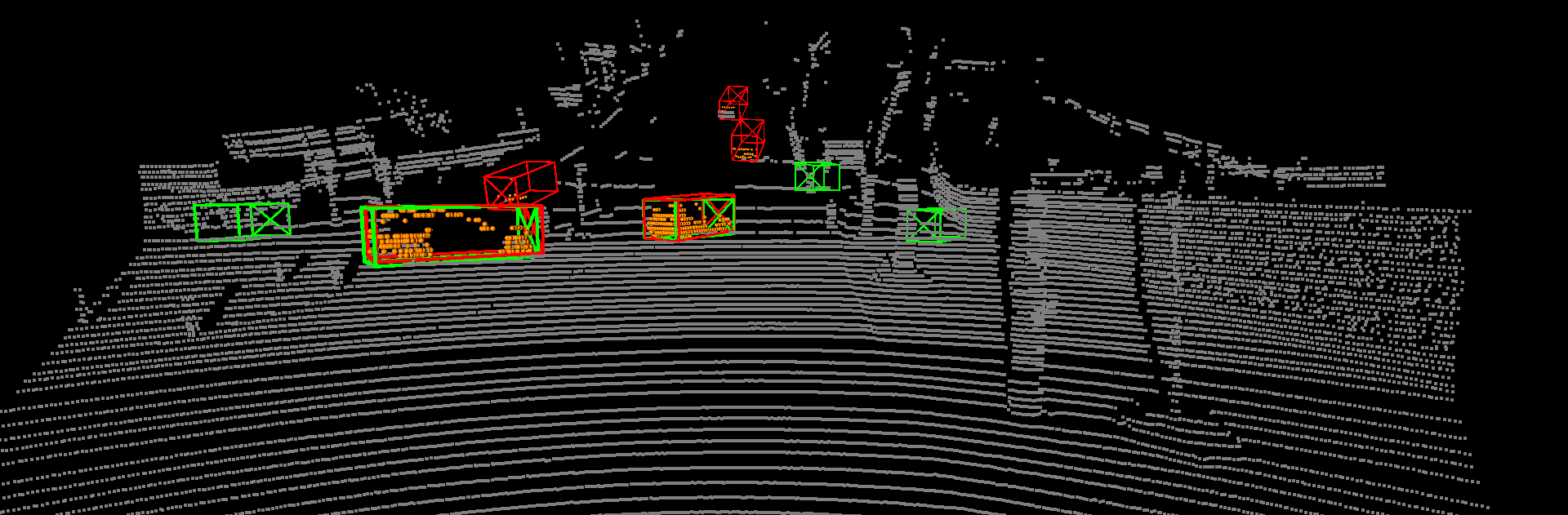}\vspace{-4.5pt}
        \includegraphics[width=1\linewidth]{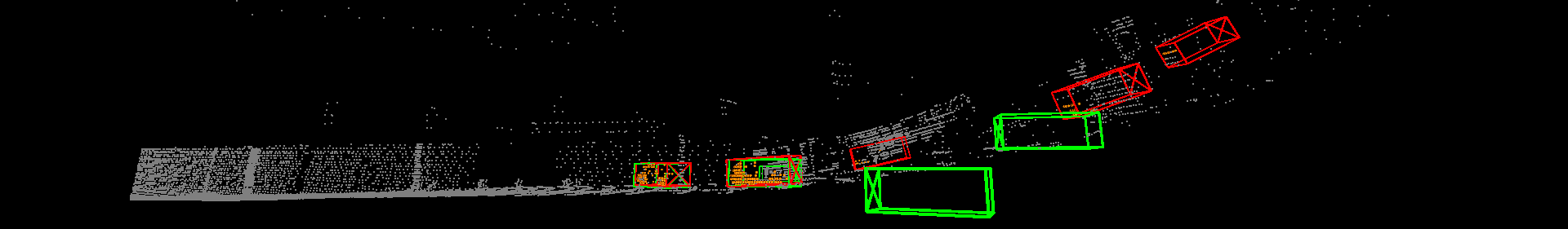}\vspace{2pt}
        \includegraphics[width=1\linewidth]{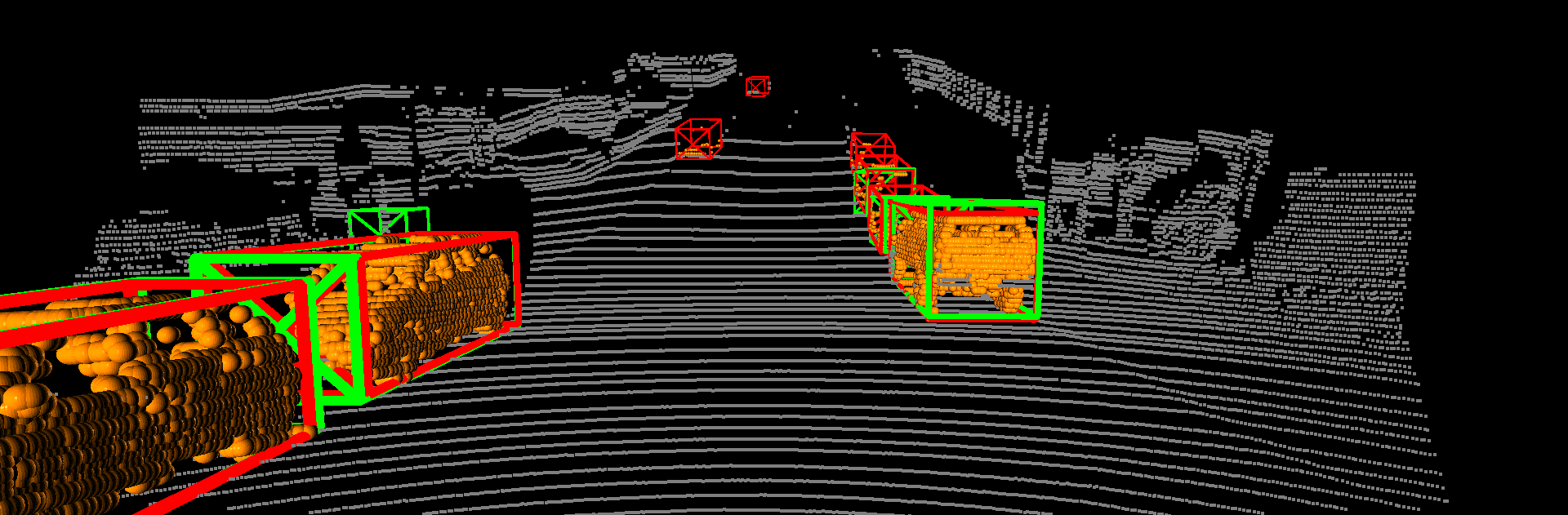}\vspace{-4.5pt}
        \includegraphics[width=1\linewidth]{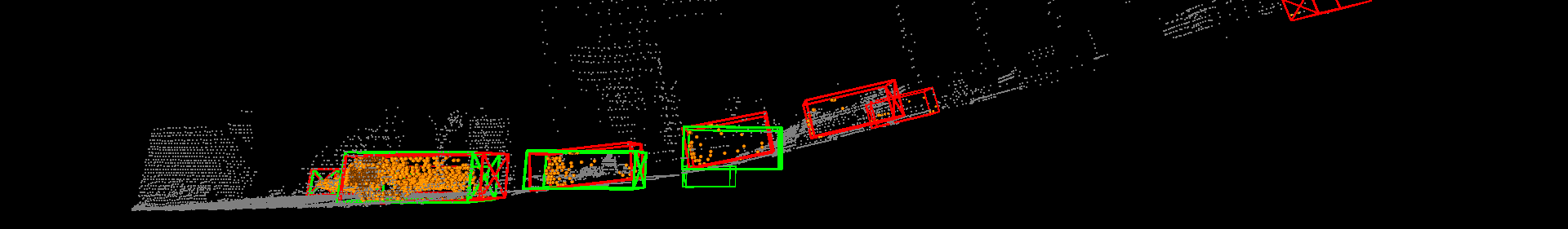}\vspace{2pt}
        \includegraphics[width=1\linewidth]{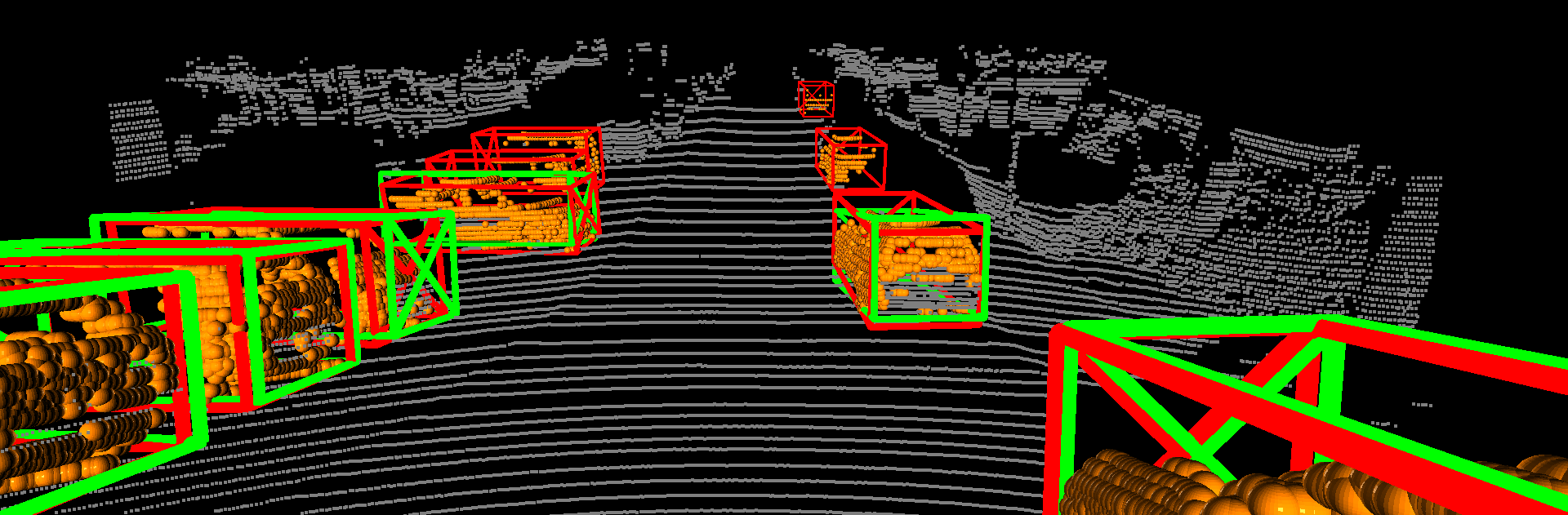}\vspace{-4.5pt}
        \includegraphics[width=1\linewidth]{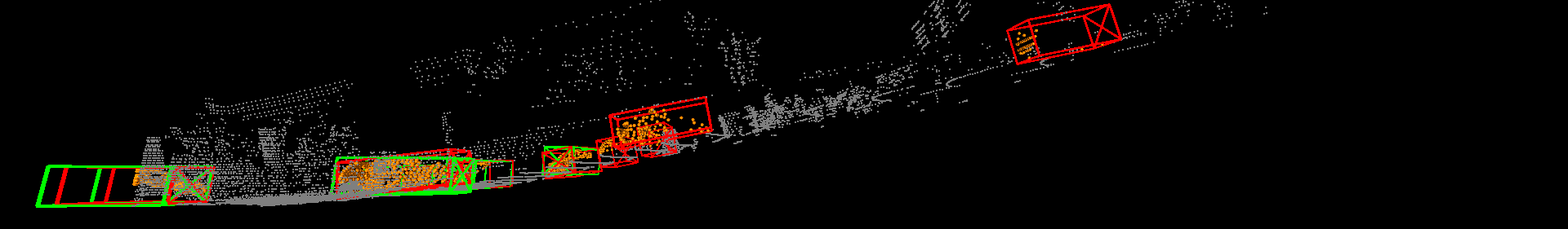}
        \end{minipage}
    } 
    \subfloat[]{%
        \begin{minipage}[]{0.18\linewidth}
        \includegraphics[width=1\linewidth]{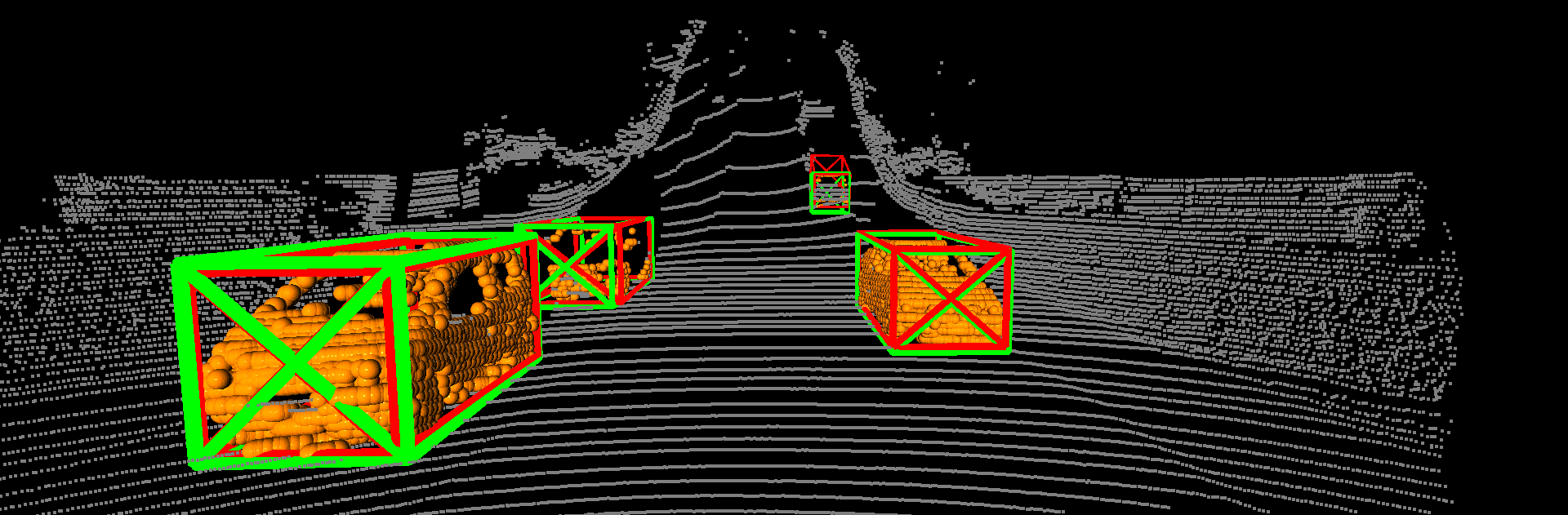}\vspace{-4.5pt}
        \includegraphics[width=1\linewidth]{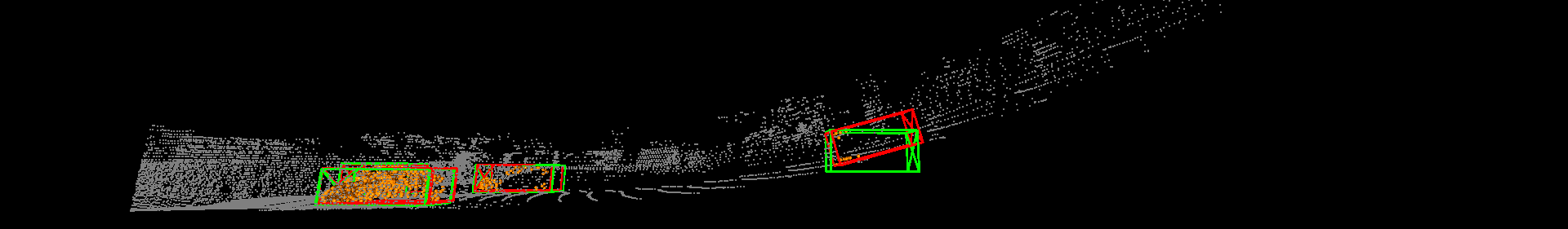}\vspace{2pt}
        \includegraphics[width=1\linewidth]{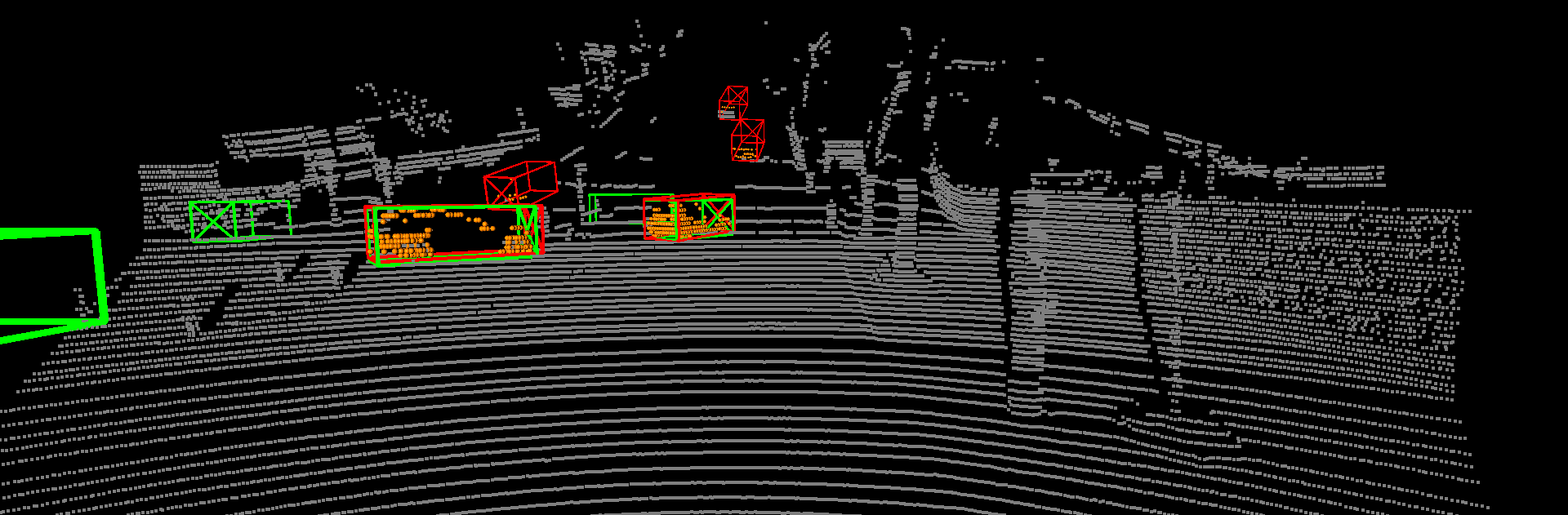}\vspace{-4.5pt}
        \includegraphics[width=1\linewidth]{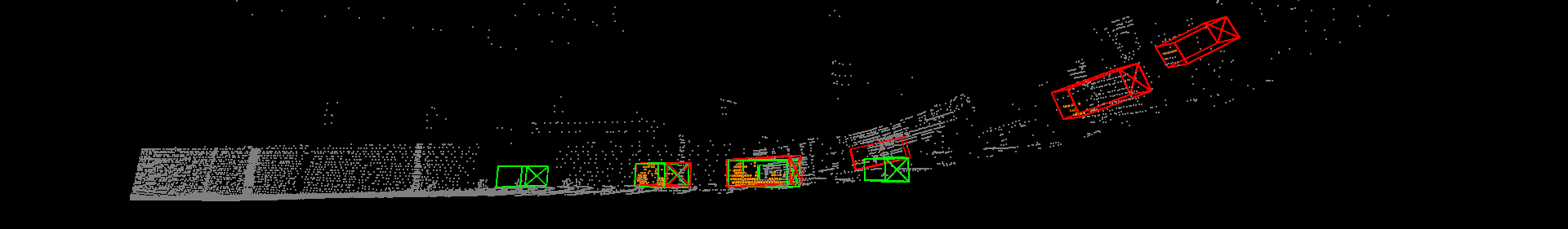}\vspace{2pt}
        \includegraphics[width=1\linewidth]{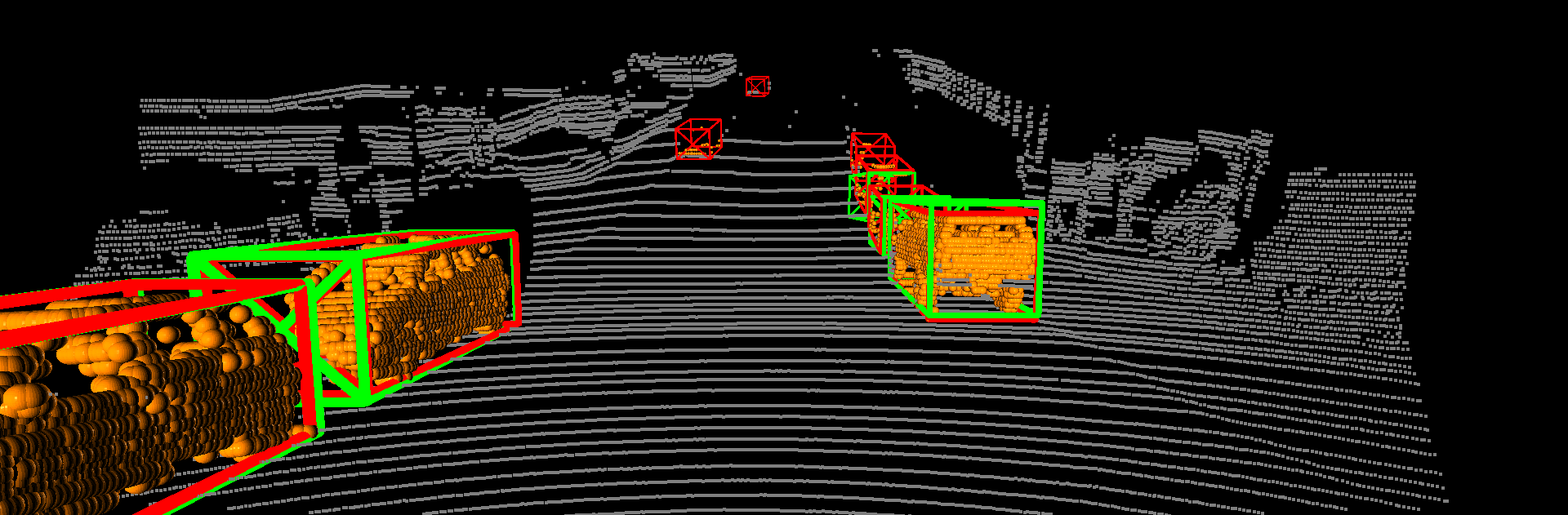}\vspace{-4.5pt}
        \includegraphics[width=1\linewidth]{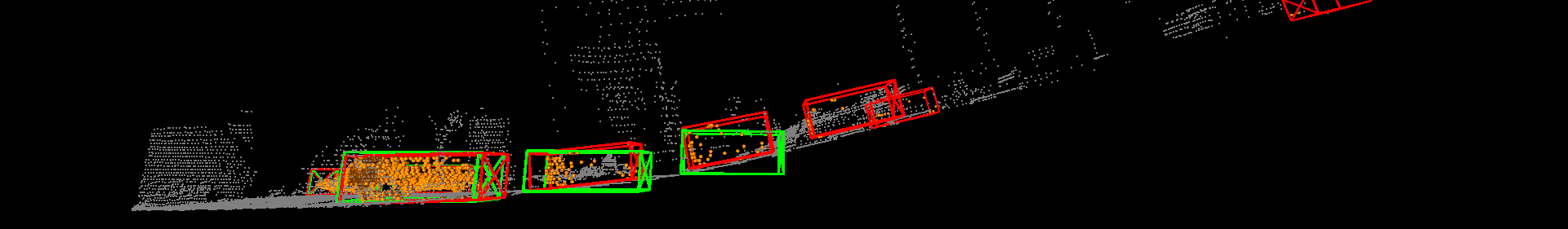}\vspace{2pt}
        \includegraphics[width=1\linewidth]{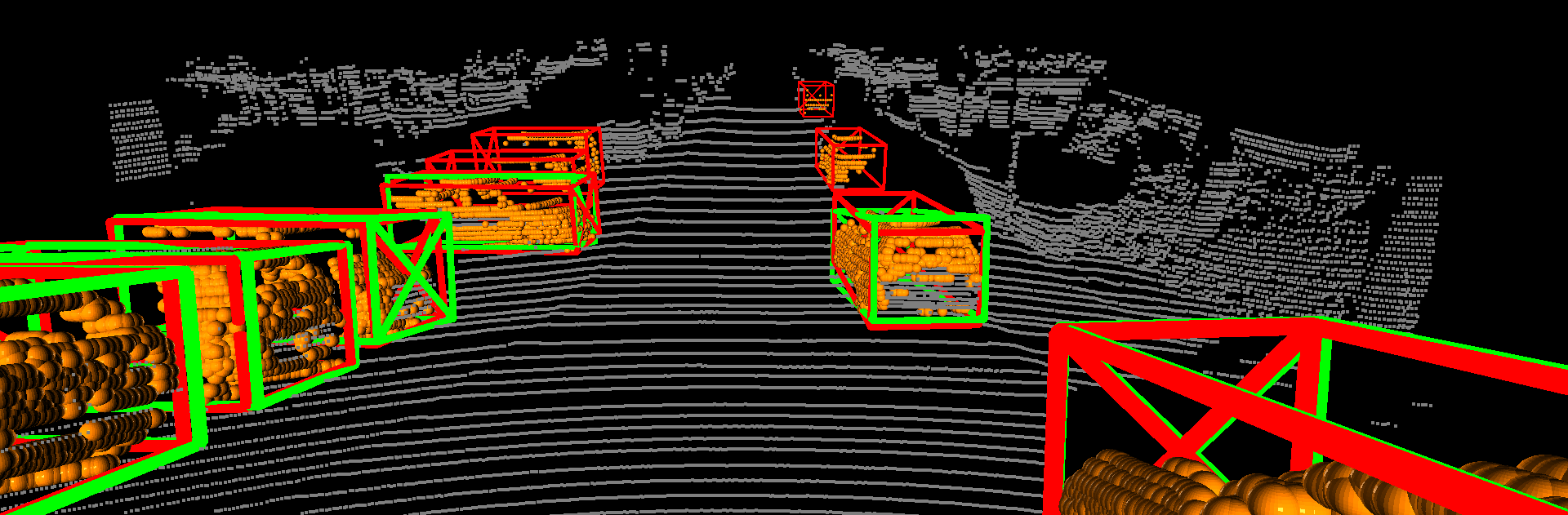}\vspace{-4.5pt}
        \includegraphics[width=1\linewidth]{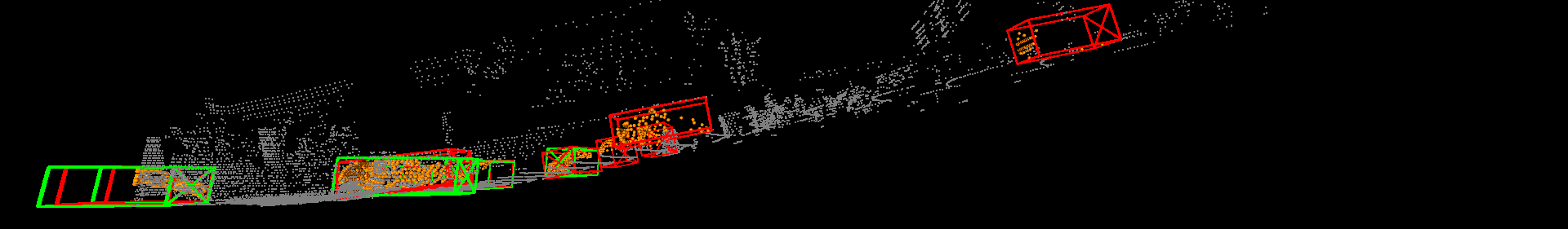}
        \end{minipage}
    } 
    \subfloat[]{%
        \begin{minipage}[]{0.18\linewidth}
        \includegraphics[width=1\linewidth]{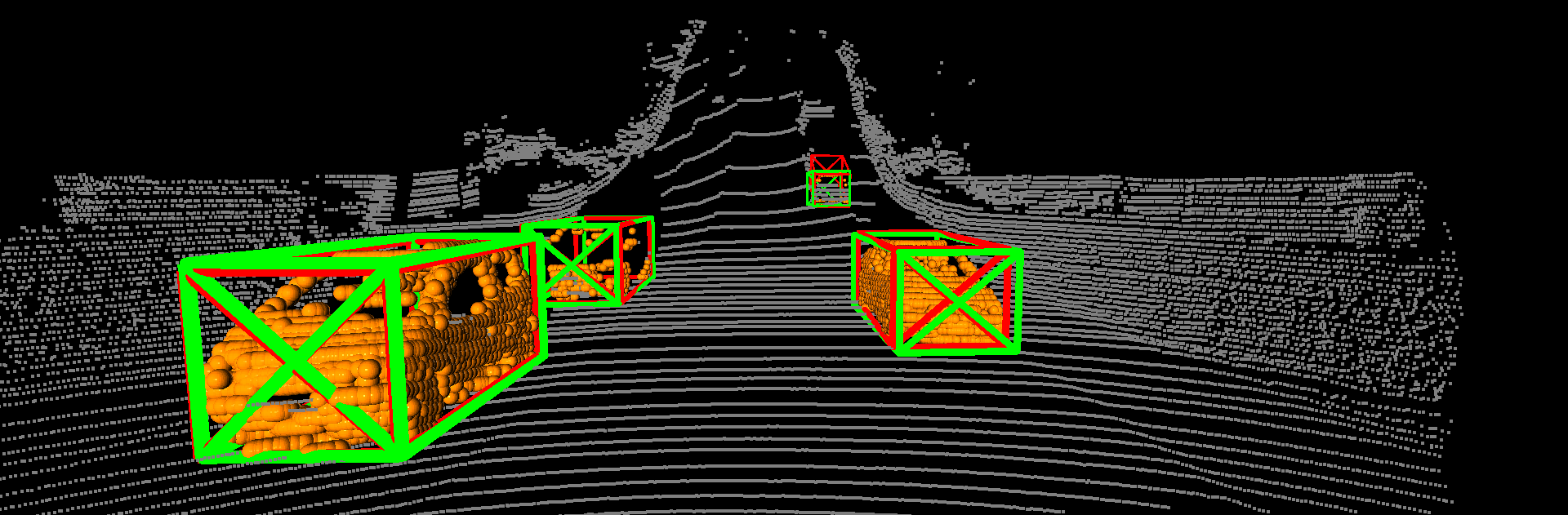}\vspace{-4.5pt}
        \includegraphics[width=1\linewidth]{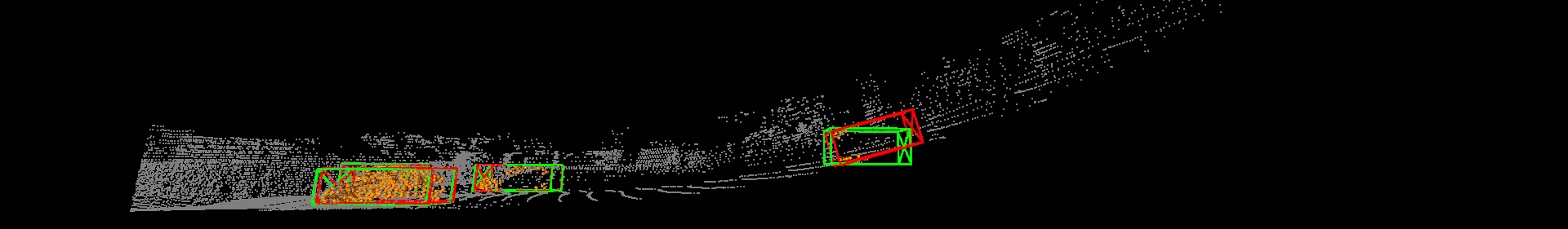}\vspace{2pt}
        \includegraphics[width=1\linewidth]{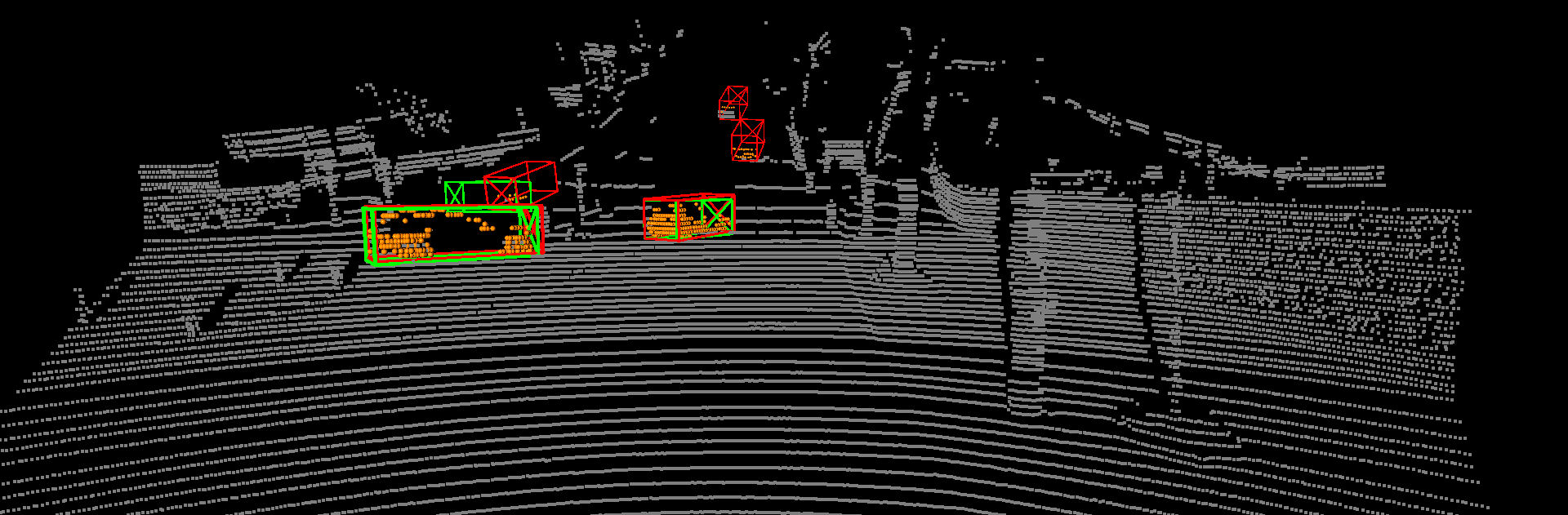}\vspace{-4.5pt}
        \includegraphics[width=1\linewidth]{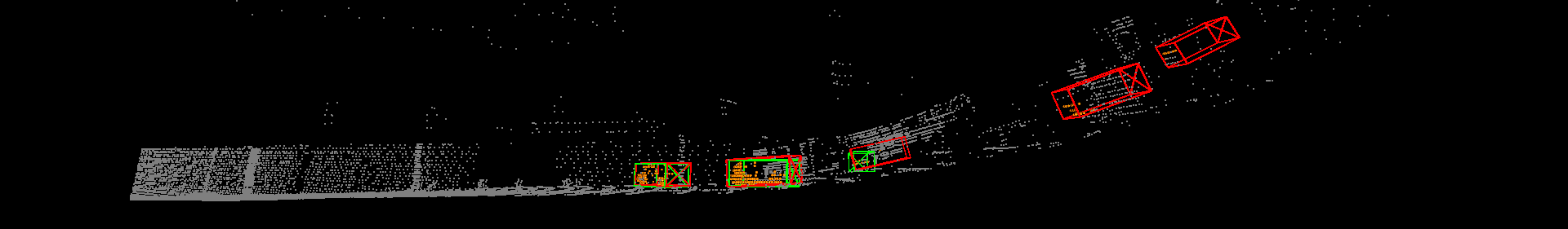}\vspace{2pt}
        \includegraphics[width=1\linewidth]{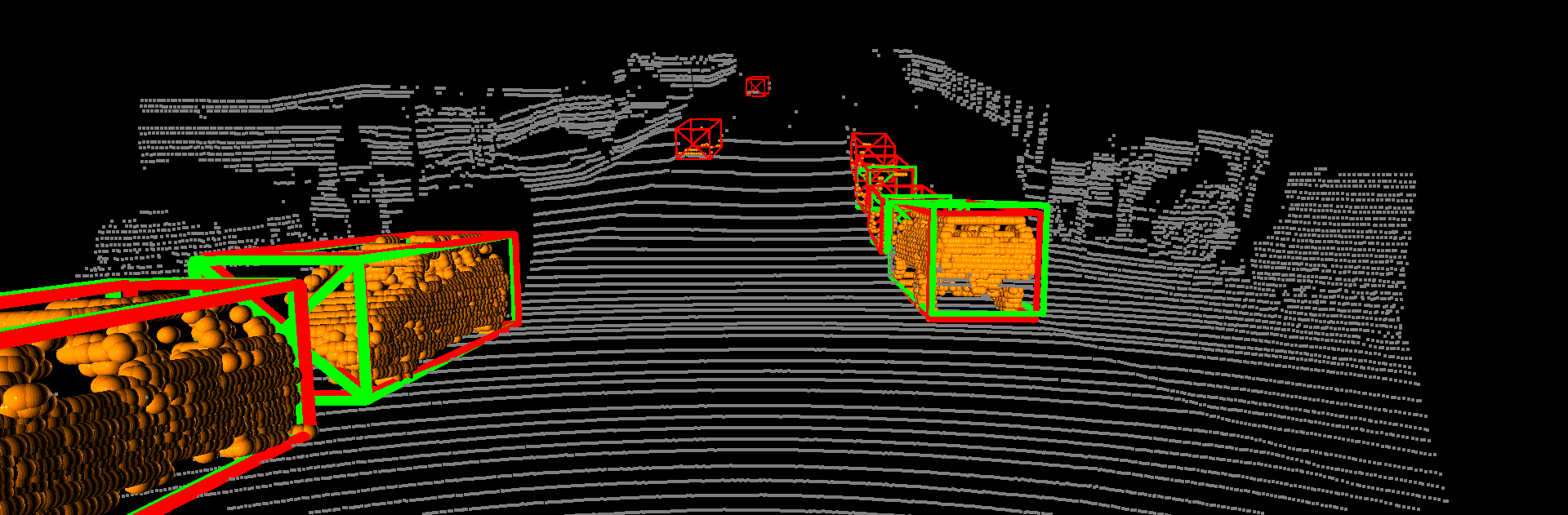}\vspace{-4.5pt}
        \includegraphics[width=1\linewidth]{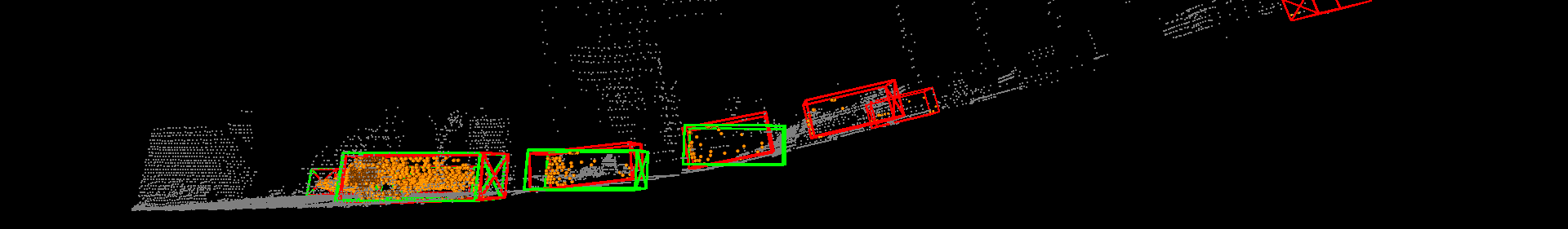}\vspace{2pt}
        \includegraphics[width=1\linewidth]{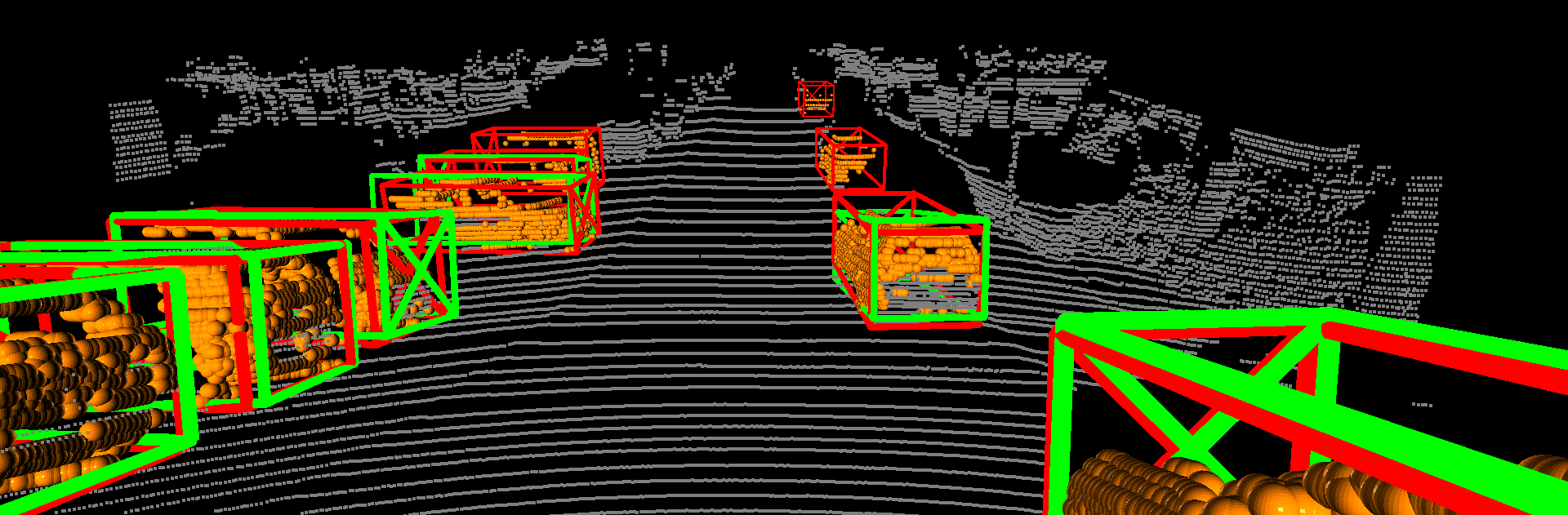}\vspace{-4.5pt}
        \includegraphics[width=1\linewidth]{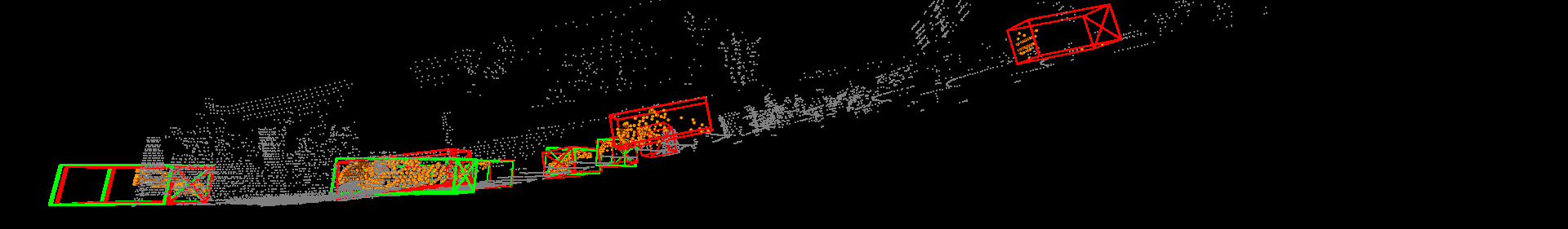}
        \end{minipage}
    } 
    \subfloat[]{%
        \begin{minipage}[]{0.18\linewidth}
        \includegraphics[width=1\linewidth]{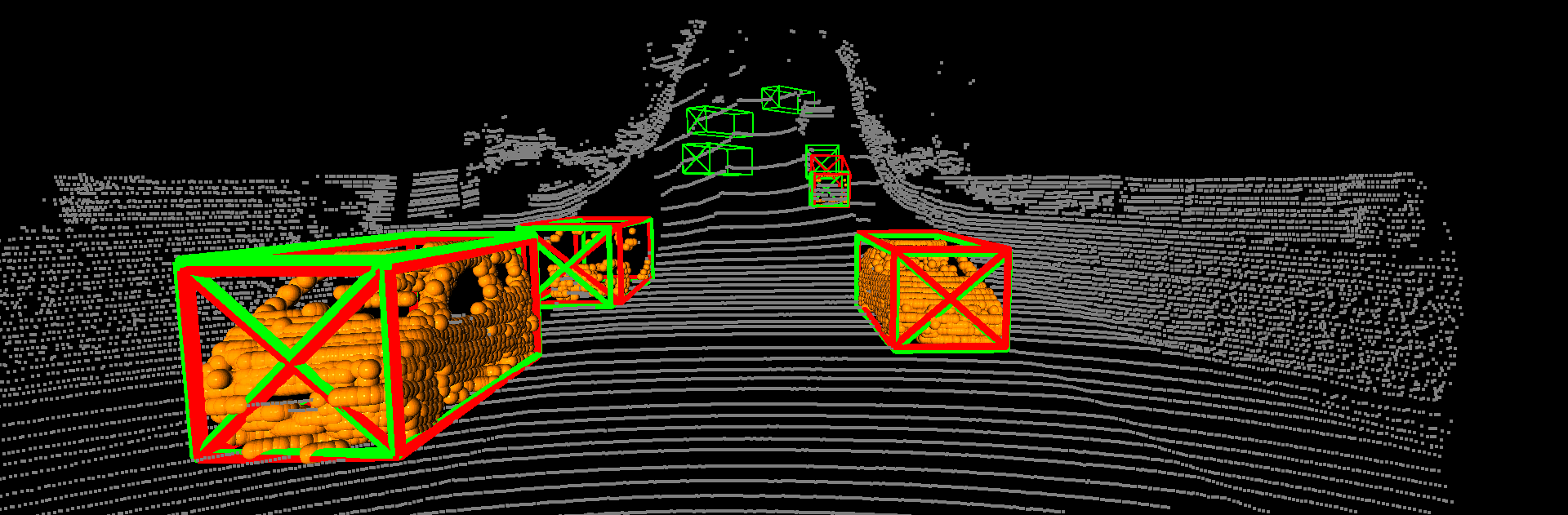}\vspace{-4.5pt}
        \includegraphics[width=1\linewidth]{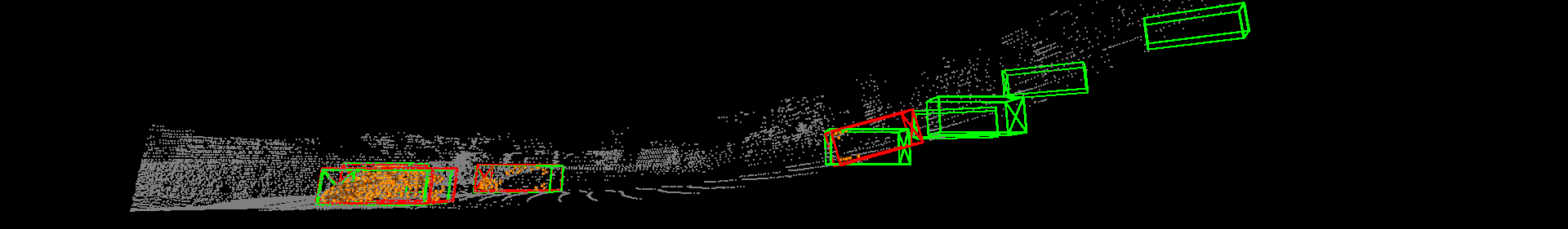}\vspace{2pt}
        \includegraphics[width=1\linewidth]{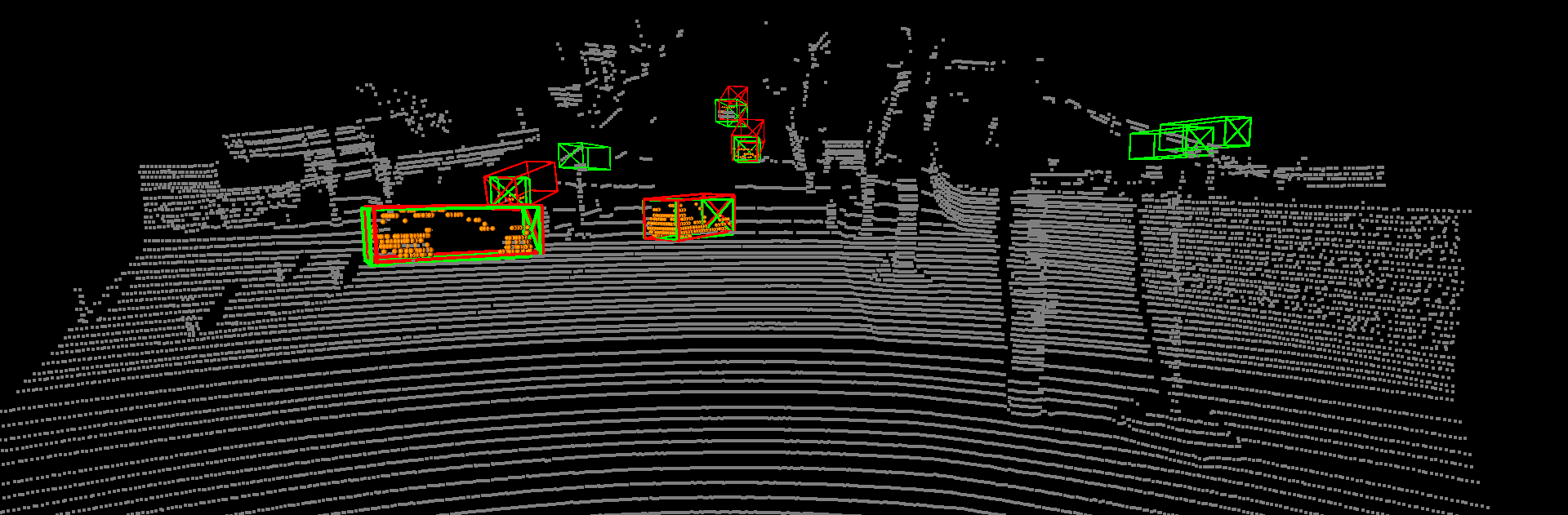}\vspace{-4.5pt}
        \includegraphics[width=1\linewidth]{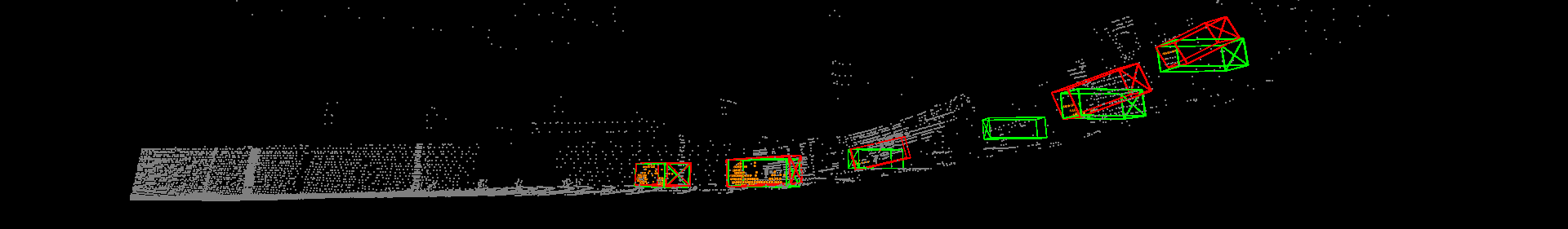}\vspace{2pt}
        \includegraphics[width=1\linewidth]{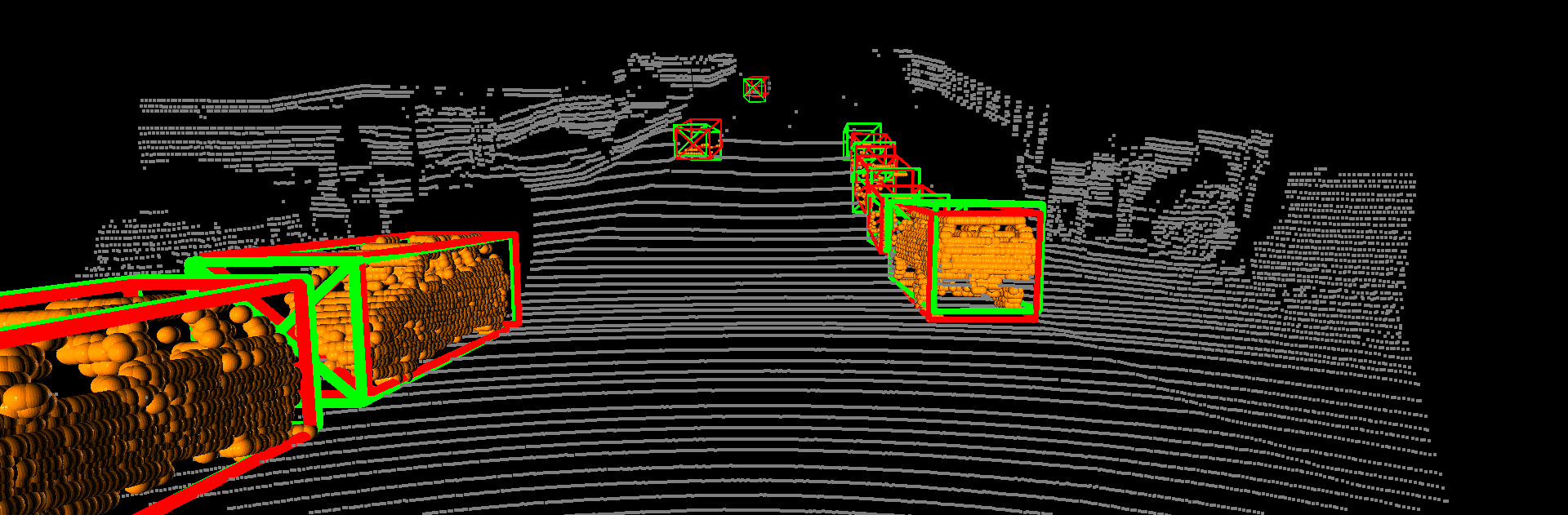}\vspace{-4.5pt}
        \includegraphics[width=1\linewidth]{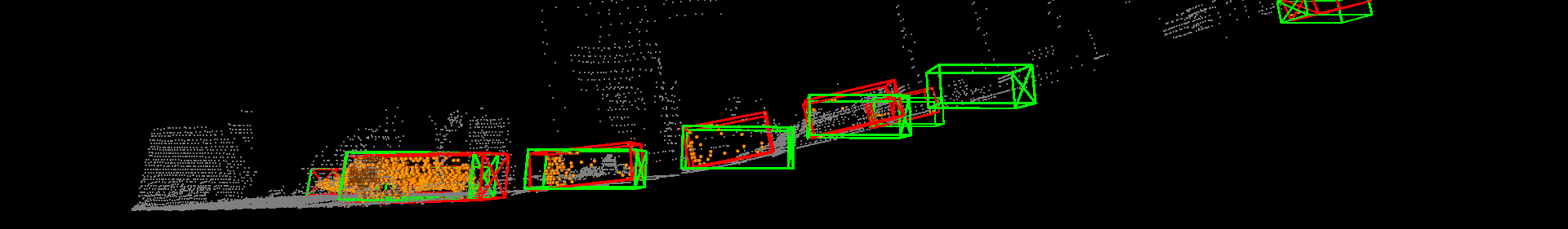}\vspace{2pt}
        \includegraphics[width=1\linewidth]{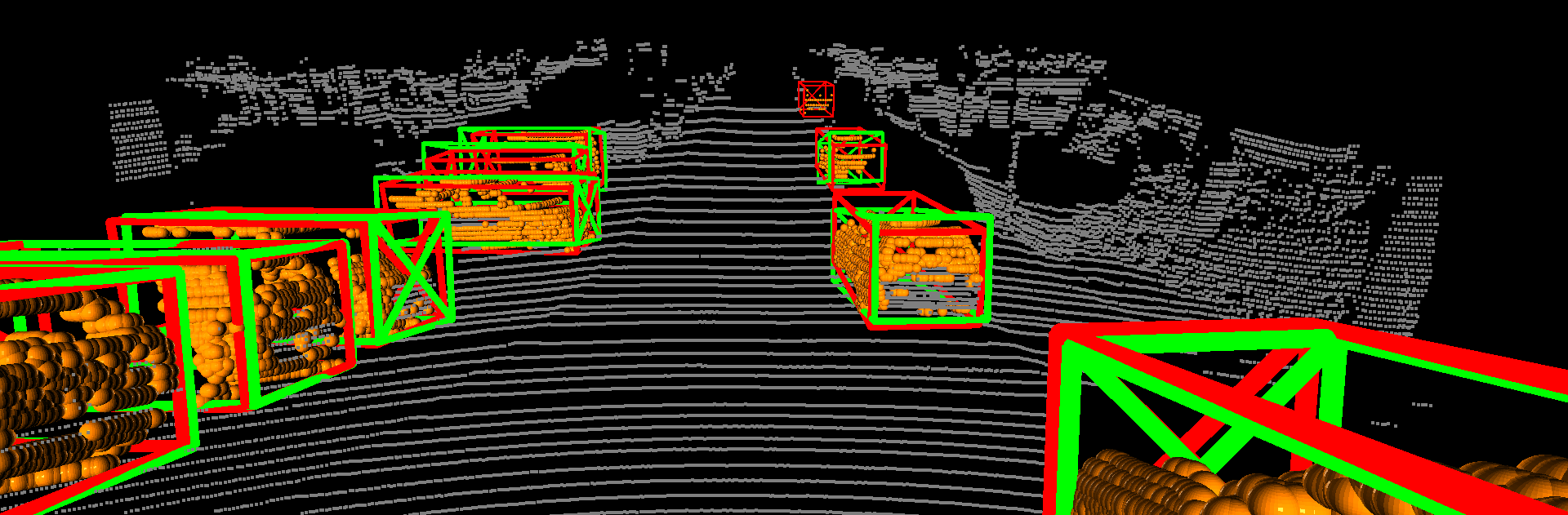}\vspace{-4.5pt}
        \includegraphics[width=1\linewidth]{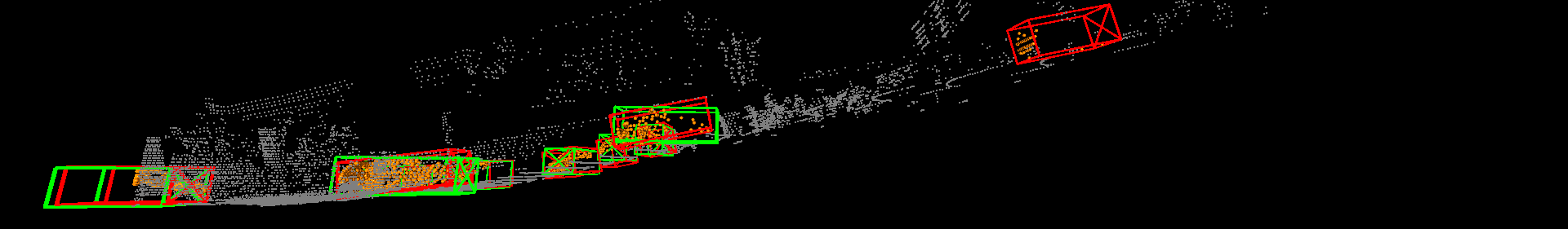}
        \end{minipage}
    } 
    \subfloat[]{%
        \begin{minipage}[]{0.18\linewidth}
        \includegraphics[width=1\linewidth]{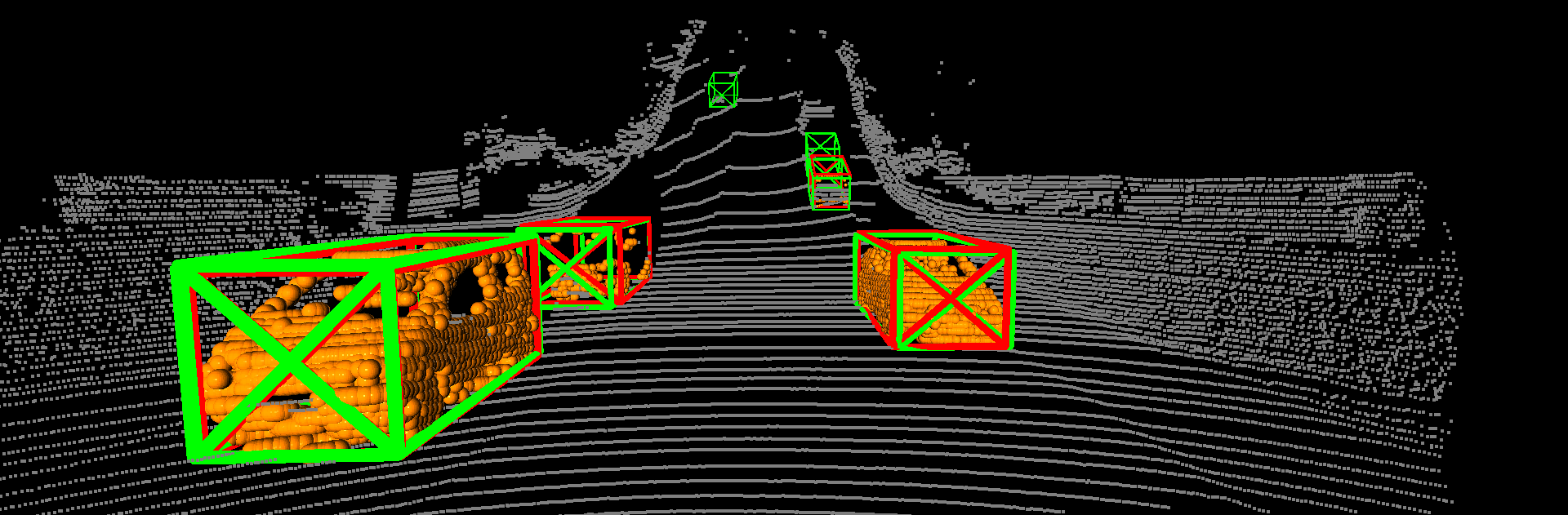}\vspace{-4.5pt}
        \includegraphics[width=1\linewidth]{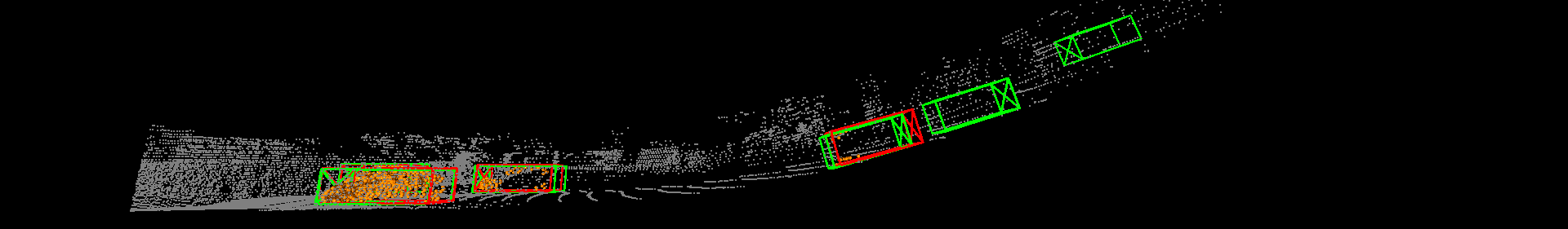}\vspace{2pt}
        \includegraphics[width=1\linewidth]{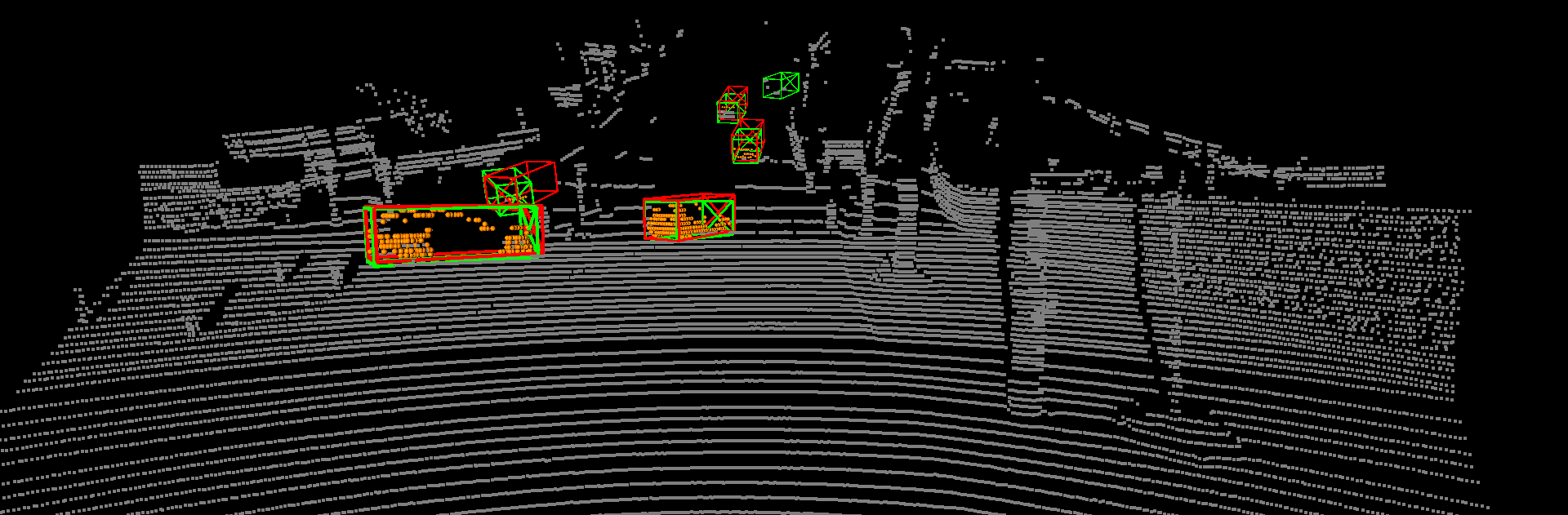}\vspace{-4.5pt}
        \includegraphics[width=1\linewidth]{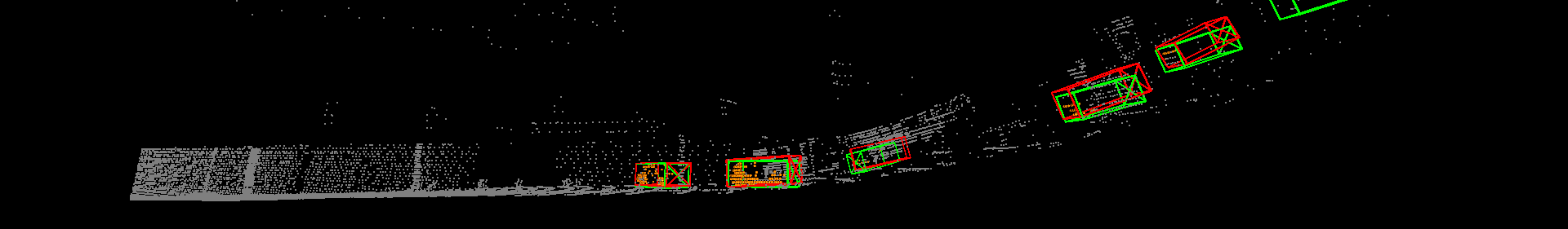}\vspace{2pt}
        \includegraphics[width=1\linewidth]{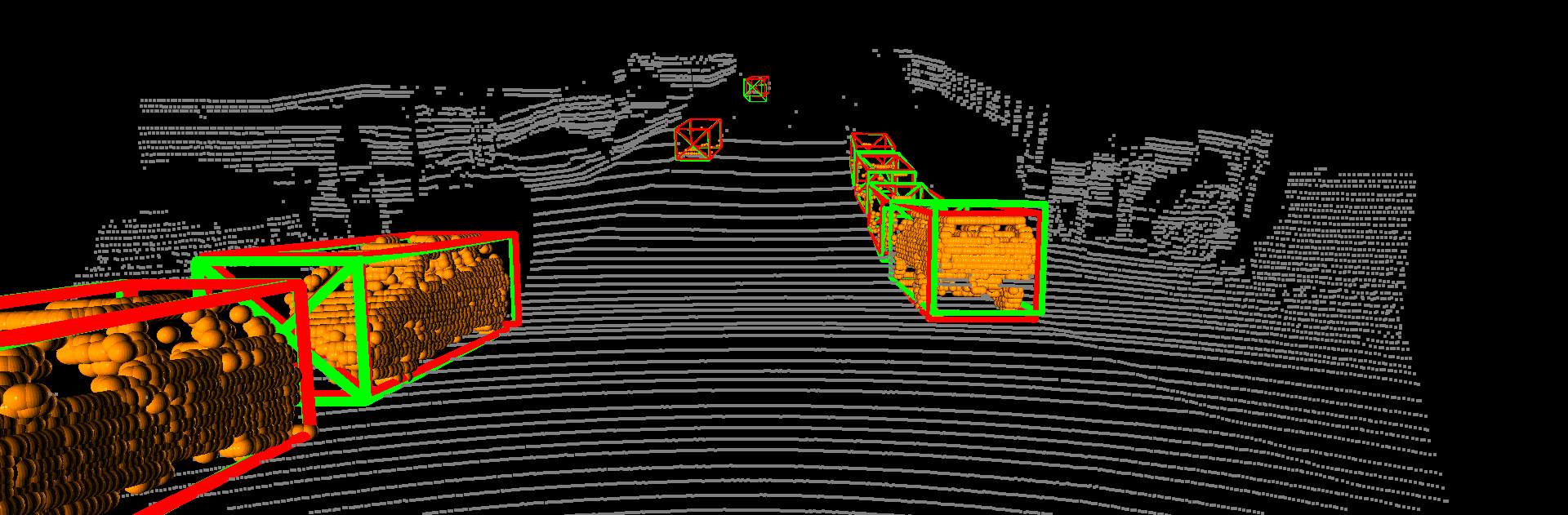}\vspace{-4.5pt}
        \includegraphics[width=1\linewidth]{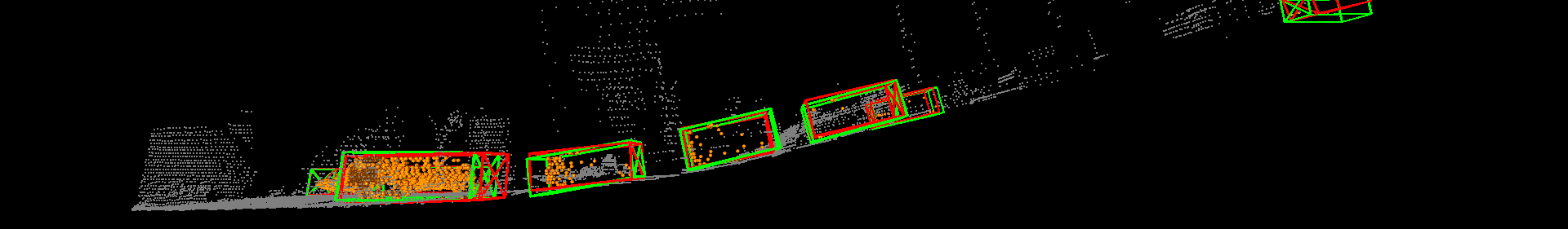}\vspace{2pt}
        \includegraphics[width=1\linewidth]{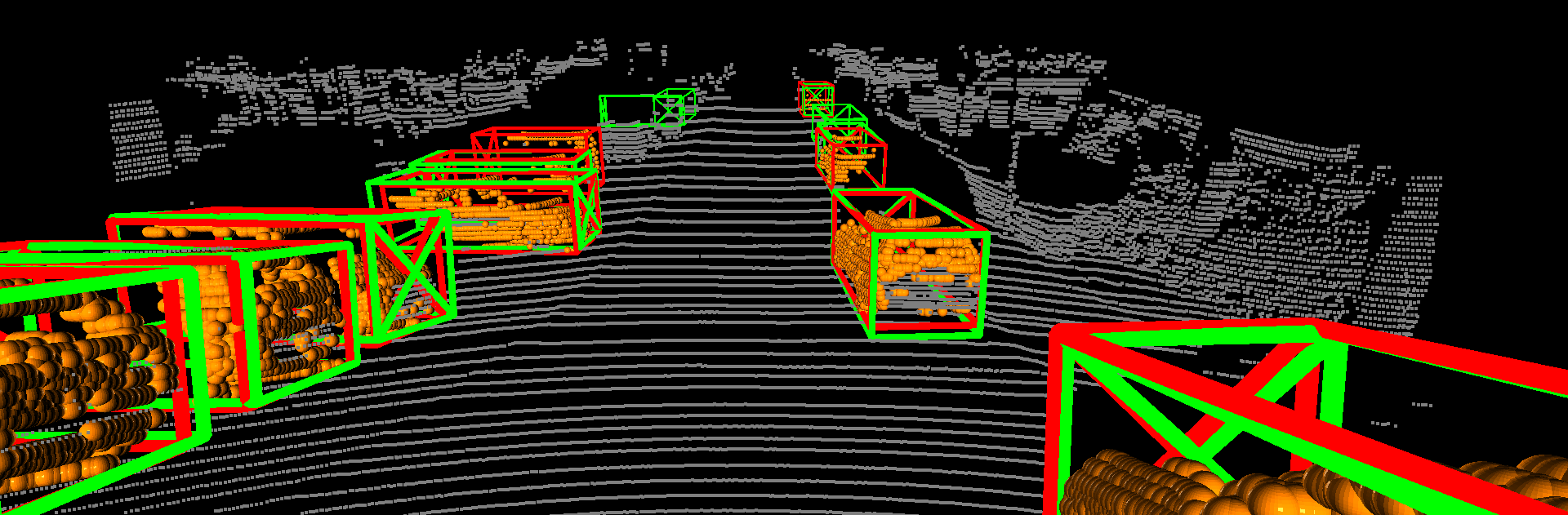}\vspace{-4.5pt}
        \includegraphics[width=1\linewidth]{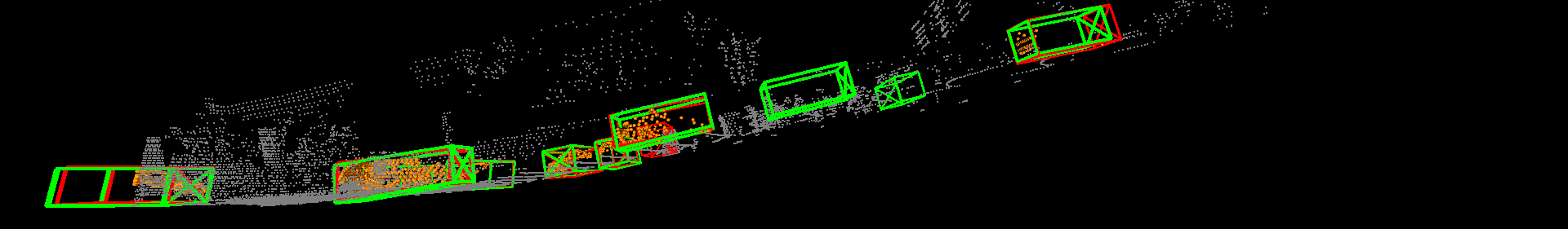}
        \end{minipage}
    }\hspace{-5pt}
    \subfloat{%
        \begin{minipage}[]{0\linewidth}
            \rotatebox{90}{\tiny Scene 4 \hspace{4pt} Scene 3 \hspace{4pt} Scene 2 \hspace{4pt} Scene 1}
        \end{minipage}
    }
\caption{
Visualization results on SlopedKITTI. (a) CenterPoint\cite{yin2021center}. (b) PV-RCNN\cite{shi2020pv}. (c) Voxel R-CNN\cite{deng2020voxel}. (d) PointRCNN\cite{shi2019pointrcnn}. (e) Det6D(Ours). 
Each row represents a different scene. 
In each result, the top image is the front view, and the bottom is the side view showing the shape of terrain. 
Red and green boxes are the ground truths and predicted results, respectively. 
Please zoom in for details.}
\label{fig:slopekitti_viz}
\end{figure} 

%% file: _exp__table_slopedkitti.tex
\begin{table*}[t]
\caption{
Performance Comparison of the State-of-the-art Methods for Car Detection on the SlopedKITTI Val Set.
Average Precision (AP) is Calculated with an IoU Threshold of 0.7 for 3D and BEV Detection, and with a Center Distance Threshold of 1.0m for Rotated Full 3D Detection.
The Subscript R40 Indicates That AP is Calculated with 40 Recall Positions.
\label{tab:table1}}
\centering
\resizebox{\textwidth}{!}{
\begin{tabular}{c|r|r||ccc|ccc||ccccc}
\hline 
&
\multirow{2}{*}{ Method } &
\multirow{2}{*}{ Reference }&
\multicolumn{3}{c|}{$\text{Car 3D AP}_{\text{R}40}@0.7$} & 
\multicolumn{3}{c||}{$\text{Car BEV AP}_{\text{R}40}@0.7$}& 
\multicolumn{5}{c}{$\text{Car Rotated 3D Metrics}_{\text{R}40}@1.0m$}
\\
&&& 
Easy & Mod. & Hard & 
Easy & Mod. & Hard &
$\text{AP}_{cd}$ & ATS$\uparrow$ & ASS $\uparrow$& AOS $\uparrow$& RODS$\uparrow$
\\
\hline
\hline
\multirow{5}{*}{\rotatebox{90}{Voxel-based}}
&SECOND\cite{yan2018second}&Sensors 2018&
56.99&37.23&35.34&
67.25&44.19&44.26&
49.50&77.22&86.49&76.33&64.76\\ 

&PointPillars\cite{pointpillars}&CVPR 2019&
53.86&34.10&33.41&
70.58&46.40&46.05&
47.37&76.95&86.22&77.94&63.87\\ 

&Part-$A^2$\cite{shi2019part}&TPAMI 2020& 
56.43 & 36.92 & 35.09 &
70.52 & 48.32 & 47.21 &
50.15 & 77.37 & 86.46 & 80.34 & 65.77\\

&PV-RCNN\cite{shi2020pv}&CVPR 2020& 
56.99&37.25&35.34&
67.25&44.25&44.19&
46.94&79.80&86.81&83.00&65.07\\ 

&CenterPoint\cite{yin2021center}&CVPR 2021&
56.30&36.50&38.59&
74.67&50.00&49.83&
51.06&78.17&86.64&77.73&65.95\\ 

&Voxel R-CNN\cite{deng2020voxel}&AAAI 2021&
57.16&37.50&36.93&
73.41&49.92&49.89&
50.99&78.59&86.85&78.60&66.17\\ 
\hline
\hline
\multirow{5}{*}{\rotatebox{90}{Point-based}}
&PointRCNN\cite{shi2019pointrcnn}&CVPR 2019&
57.88&39.11&37.80&
79.13&68.63&67.66&
74.12&68.47&83.99&64.38&72.20\\ 

&3DSSD\cite{yang20203dssd}&CVPR 2020&
55.13&37.01&34.56&
88.05&69.72&65.38&
69.36&69.04&82.43&70.98&71.75\\ 

&3DSSD-SASA\cite{chen2022sasa}&AAAI 2022&
55.65&37.28&34.57&
88.01&74.03&67.93&
72.01&69.23&83.33&69.12&72.94\\ 

&IA-SSD\cite{zhang2022not}&CVPR 2022&
58.27 & 39.55 & 37.27 &
83.10 & 65.71 & 63.21 &
67.83 & 70.22 & 83.87 & 63.19 & 70.13 \\

&\cellcolor{cyan!20}Det6D(Ours)&\multicolumn{1}{c||}{\cellcolor{cyan!20}-}&
\cellcolor{cyan!20}\textbf{84.03}&\cellcolor{cyan!20}\textbf{73.55}&\cellcolor{cyan!20}\textbf{71.06}&
\cellcolor{cyan!20}\textbf{92.89}&\cellcolor{cyan!20}\textbf{86.00}&\cellcolor{cyan!20}\textbf{85.22}&
\cellcolor{cyan!20}\textbf{86.88}&\cellcolor{cyan!20}\textbf{80.97}&\cellcolor{cyan!20}\textbf{86.89}&\cellcolor{cyan!20}\textbf{84.36}&\cellcolor{cyan!20}\textbf{85.48}\\ 
\hline
\end{tabular}
}
\end{table*}

%% file: _exp__table_kitti.tex
\begin{table}[t]
\caption{performance Comparison the State-of-the-art Methods for Car Detection on the KITTI Test Set and Val Split.
\label{tab:table2}}
\centering
\resizebox{\linewidth}{!}{
\begin{tabular}{c|r||ccc|ccc}
\hline 
&
\multirow{2}{*}{ Method } &
\multicolumn{3}{c|}{ Car 3D $\text{AP}_{\text{R}40}@0.7(test)$ }&
\multicolumn{3}{c}{ Car 3D $\text{AP}_{\text{R}11}@0.7(val)$}
\\
&
& Easy & Mod. & Hard & Easy & Mod. & Hard 
\\
\hline
\hline
\multirow{5}{*}{\rotatebox{90}{Voxel-based}}
&SECOND\cite{yan2018second}&
83.13 & 73.66 & 66.20 &
87.43 & 76.48 & 69.10\\ 

&PointPillars\cite{pointpillars}&
82.58 & 74.31 & 68.99 &
- & 77.98 & -  \\ 

&Part-$A^2$\cite{shi2019part}&
87.81 & 78.49 & 73.51 &
89.47 & 79.47 & 78.54  \\ 

&PV-RCNN\cite{shi2020pv}&
90.25 & 81.43 & 76.82 &
\itshape{89.35} & \itshape{83.69} & \itshape{78.70}  \\ 

&CenterPoint\cite{yin2021center}&
 - & - & - & 
\itshape{87.72} & \itshape{79.48} & \itshape{77.17}    \\

&Voxel R-CNN\cite{deng2020voxel}&
\textbf{90.90} & 81.62 & 77.06 &
89.41 & 84.52 & 78.93  \\ 

\hline
\hline
\multirow{5}{*}{\rotatebox{90}{Point-based}}

&PointRCNN\cite{shi2019pointrcnn}&
86.96 &	75.64 &	70.70 &
88.88 & 78.63 & 77.38 \\ 

&3DSSD\cite{yang20203dssd}&
88.36 & 79.57 & 74.55 &  
\textbf{89.71} & 79.45 & 78.67 \\

&3DSSD-SASA\cite{chen2022sasa}&
88.76 & \textbf{82.16} & \textbf{77.16} &  
\itshape{89.38} & \textbf{\itshape{84.80}} & \textbf{\itshape{79.01}} \\

&IA-SSD\cite{zhang2022not}&
88.87 & 80.32 & 75.10 &  
 - & 79.57 &  - \\

&\cellcolor{cyan!20}Det6D(Ours)&
\cellcolor{cyan!20}87.81 &\cellcolor{cyan!20} 79.62 &\cellcolor{cyan!20} 74.55 &  
\cellcolor{cyan!20}89.17 & \cellcolor{cyan!20} 84.41 &\cellcolor{cyan!20} 78.74\\ 
\hline
\end{tabular}
}
\end{table}

%% file: _exp__table_ablation.tex
\begin{table}[t]
    \caption{
    Ablation Study of Proposed Framework on SlopedKITTI Val Split. 
    "D.R.", "G.O.B." and "S.A." Mean Direct Regression, Ground-aware Orientation Branch and Slope-aug.
    \label{tab:ablation_study}}
    \centering
    \resizebox{0.5\textwidth}{!}{
    \begin{tabular}{cc|c||c|cc|cc}
    \hline 
    \multirow{2}{*}{G.O.B.} &
    \multirow{2}{*}{D.R.} &
    \multirow{2}{*}{S.A.}&
    3D $\text{AP}_{\text{R}40}@0.7$&
    \multicolumn{2}{c|}{$\text{R3DM}_{\text{R}40}@1.0m$}&
    \multicolumn{2}{c}{ Mean Error(rad)}\\
     & & & Moderate&AOS$\uparrow$ & OOSD$\uparrow$ & Yaw$\downarrow$ & Pitch\&Roll$\downarrow$ \\
    \hline 
    - & - & - &  36.34 & 69.01 & 72.68 & 0.15 & 0.33\\
    - & - & $\checkmark$&  70.27 & 65.56 & 81.89 & 0.15 & 0.34\\
    - & $\checkmark$  & $\checkmark$&  59.57 & 72.09 & 81.03 & 0.20 & 0.15\\
    $\checkmark$ & - & $\checkmark$ & 73.55 & 84.36 & 85.48 & 0.12 & 0.05 \\
    \hline 
    \end{tabular}
    }
\end{table}

%% file: _exp__fig_performance_drop.tex
\begin{figure}[!t]
\centering
\includegraphics[width=0.95\linewidth]{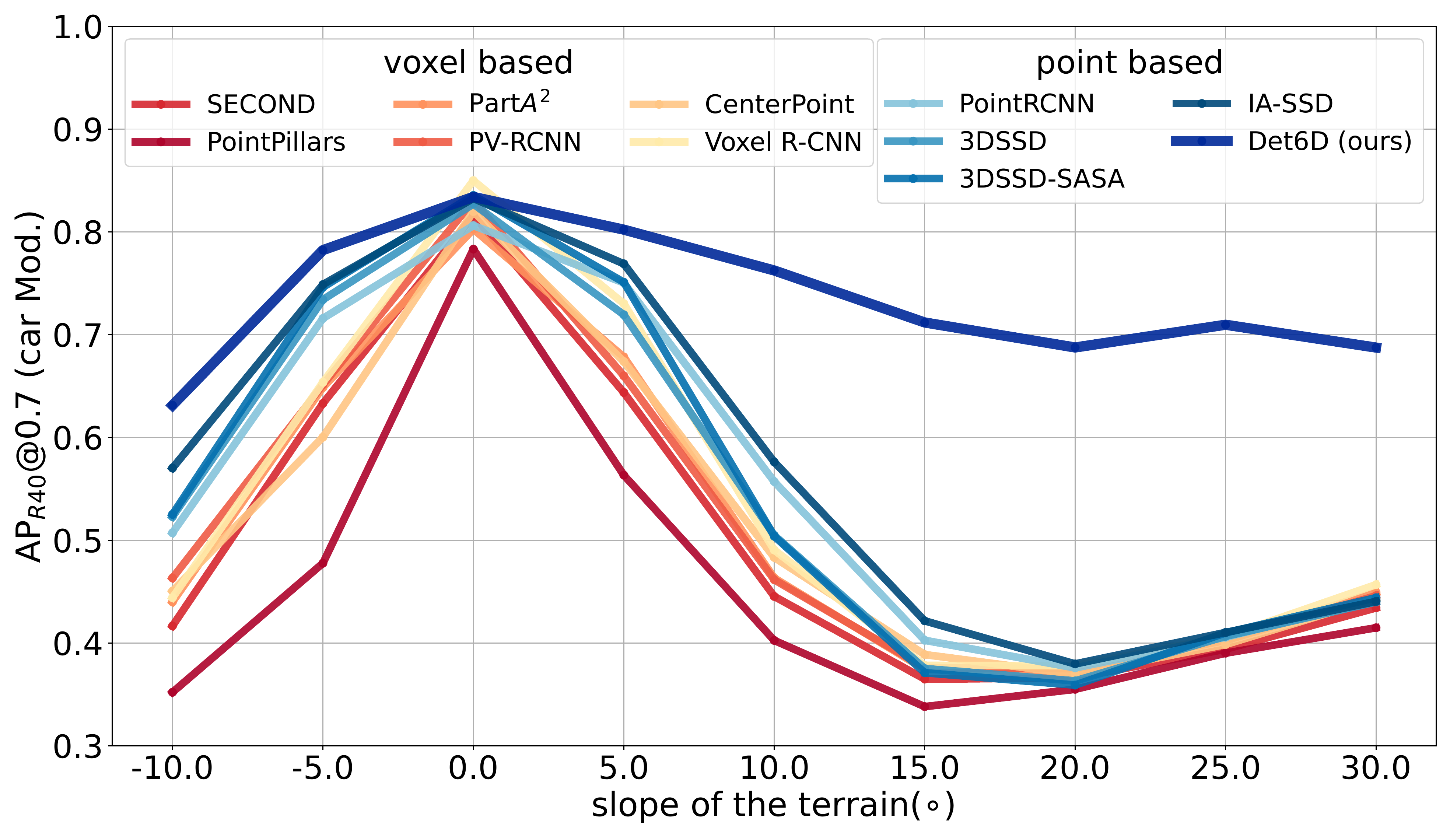}
\caption{Performance comparison in SlopedKITTI dataset with different slopes.}
\label{fig:performance_drop}
\end{figure}

%% file: _exp__fig_gtav.tex
\begin{figure}
    \centering\hspace{-5pt}
    \subfloat{%
        \setcounter{subfigure}{0}
        \begin{minipage}[]{0.01\linewidth}
            \rotatebox{90}{\tiny GTA-V}
        \end{minipage}
    }
    \subfloat{%
        \begin{minipage}[]{0.1255\linewidth}
            \includegraphics[width=1.0\linewidth]{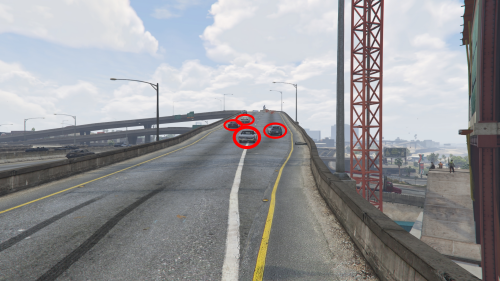}\vspace{2pt}
            \includegraphics[width=1.0\linewidth]{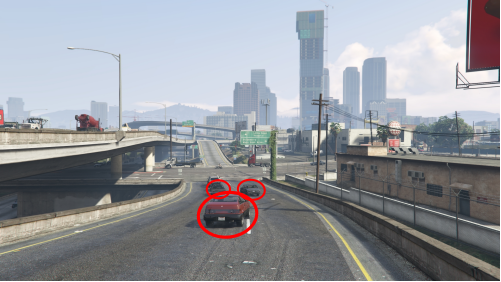}\vspace{2pt}
            \includegraphics[width=1.0\linewidth]{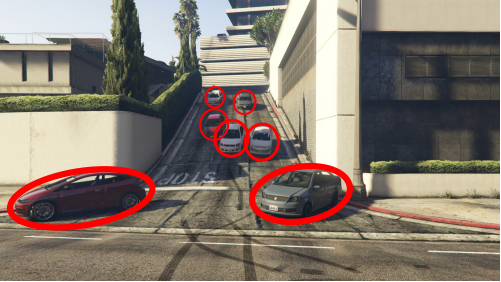}\vspace{2pt}
            \includegraphics[width=1.0\linewidth]{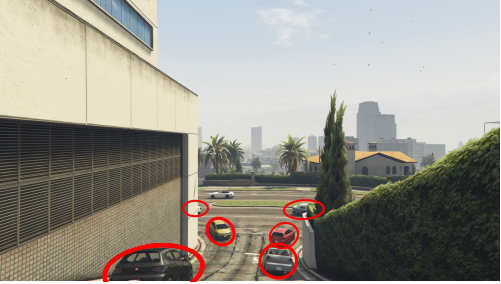}\vspace{2pt}
            \includegraphics[width=1.0\linewidth]{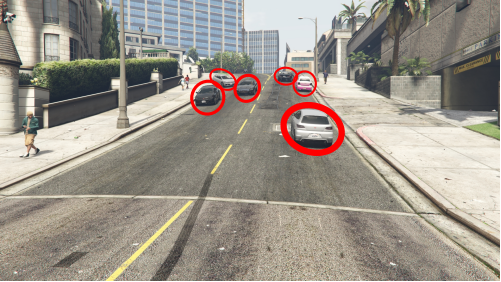}\vspace{2pt}
            \includegraphics[width=1.0\linewidth]{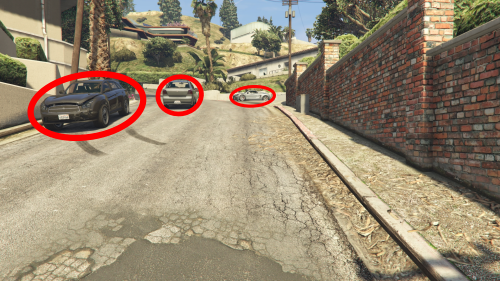}
        \end{minipage}
    }
    \subfloat{%
        \begin{minipage}[]{0.19\linewidth}
            \includegraphics[width=1.0\linewidth]{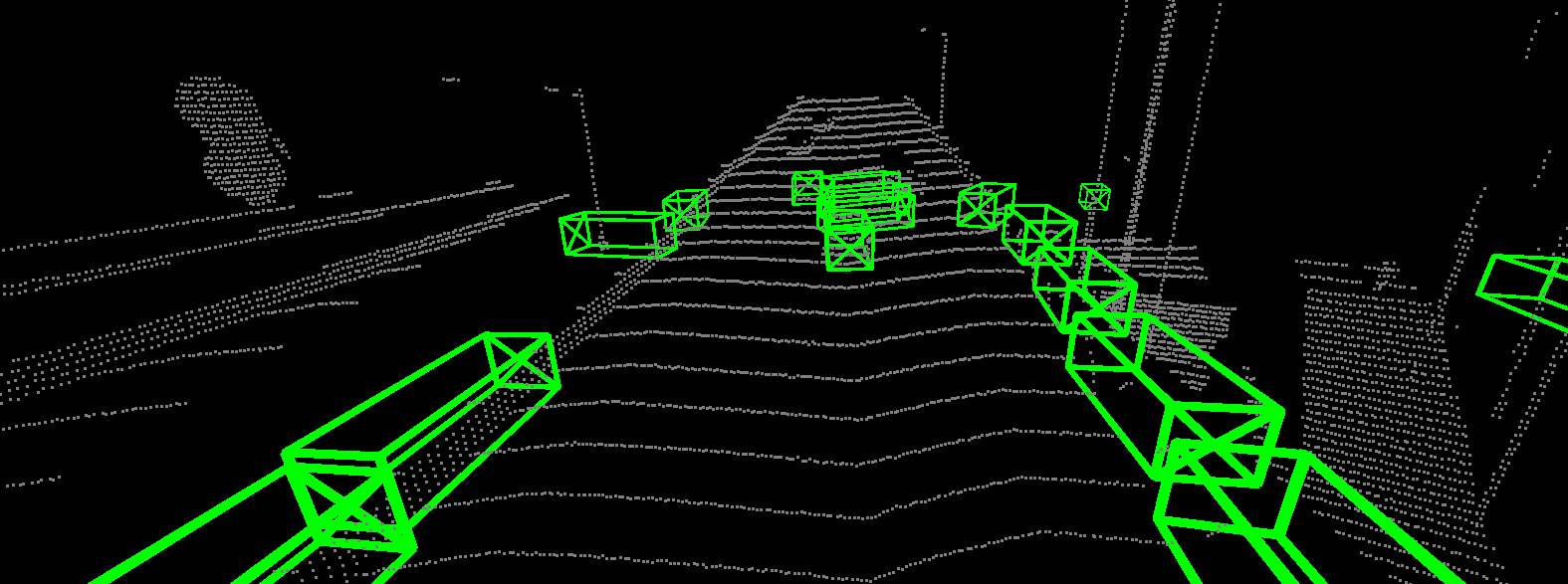}\vspace{2pt}
            \includegraphics[width=1.0\linewidth]{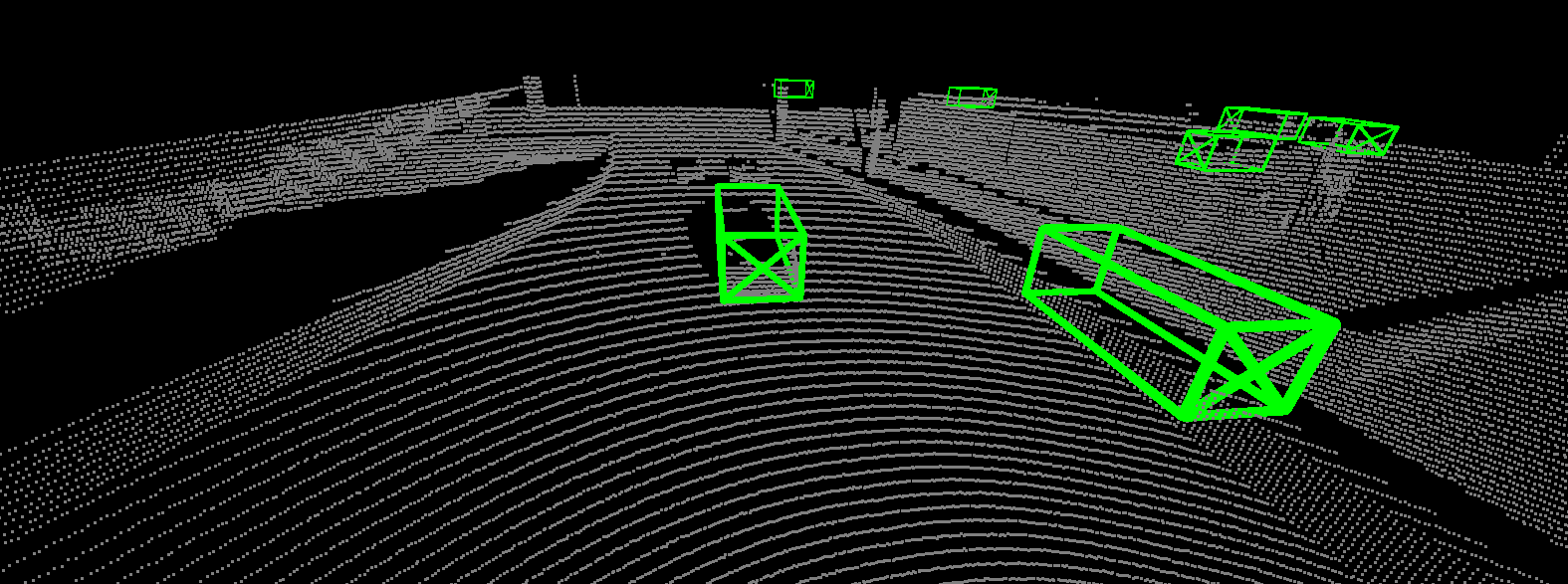}\vspace{2pt}
            \includegraphics[width=1.0\linewidth]{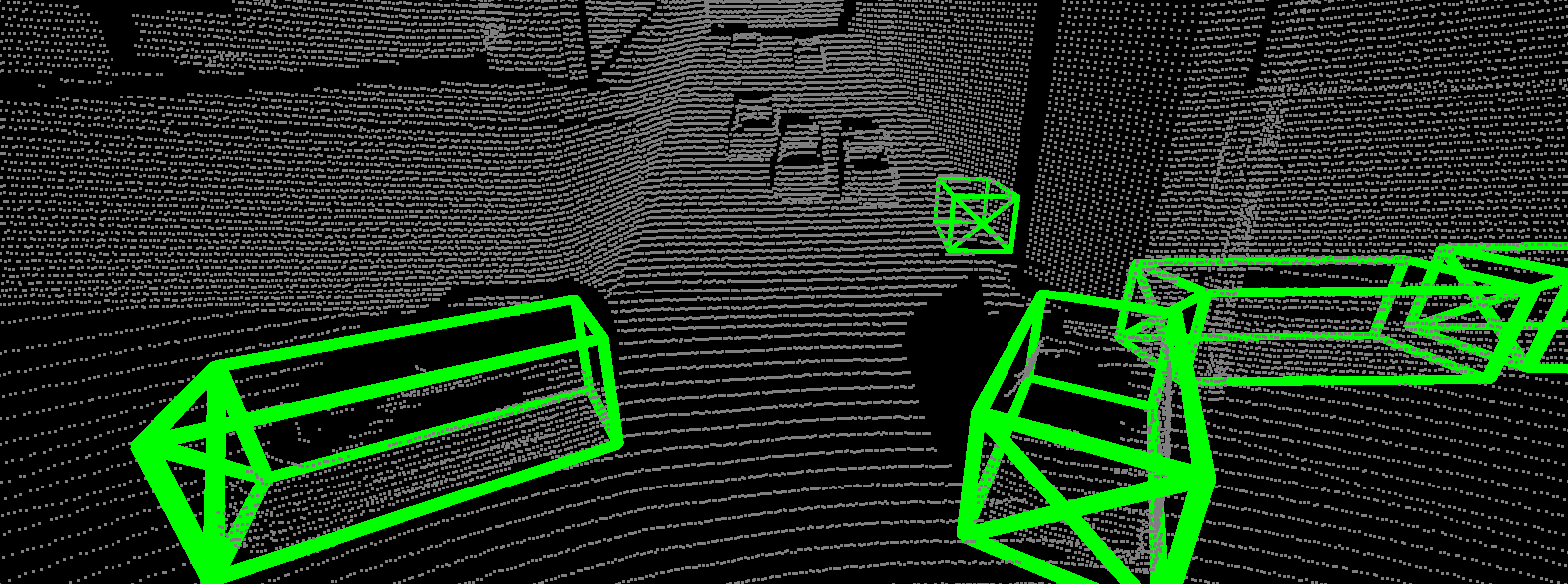}\vspace{2pt}
            \includegraphics[width=1.0\linewidth]{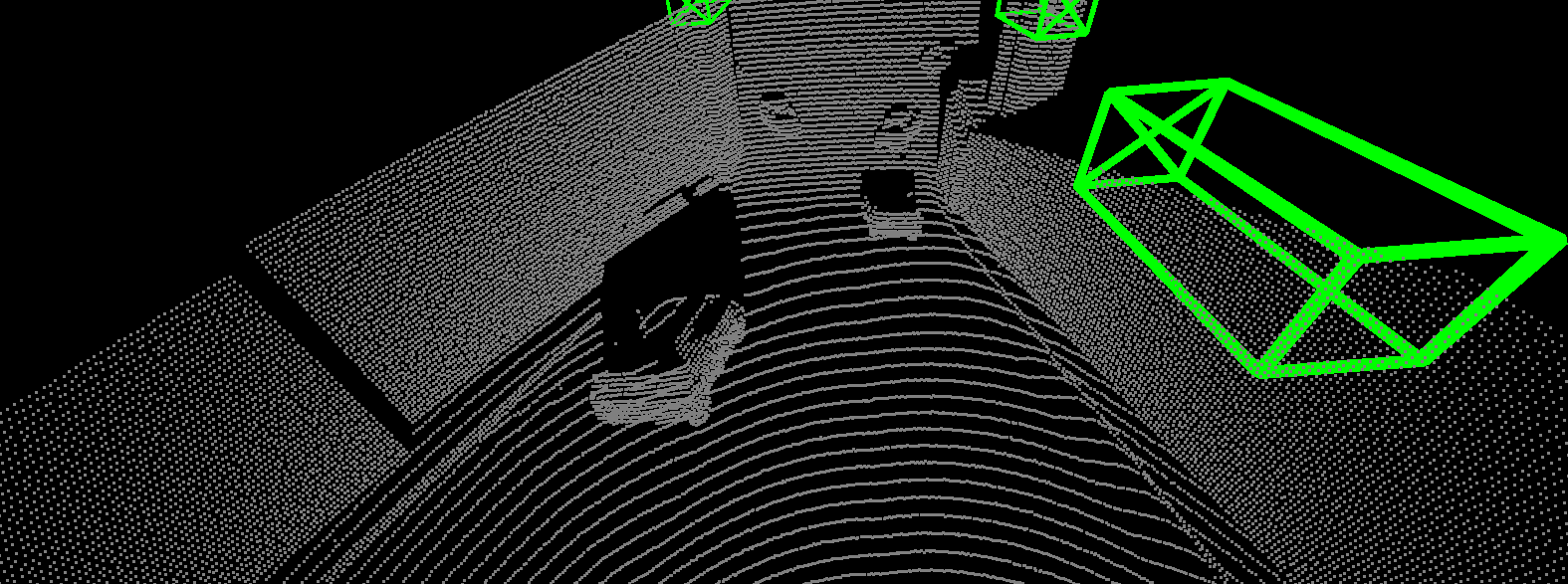}\vspace{2pt}
            \includegraphics[width=1.0\linewidth]{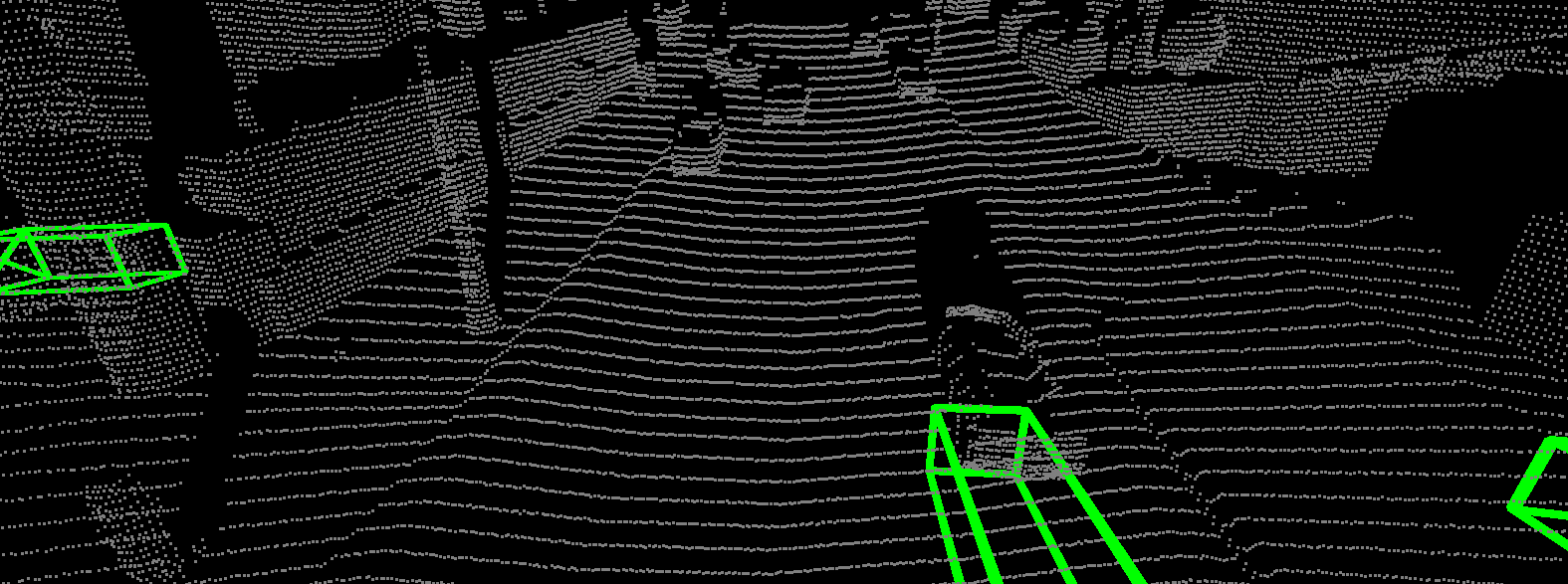}\vspace{2pt}
            \includegraphics[width=1.0\linewidth]{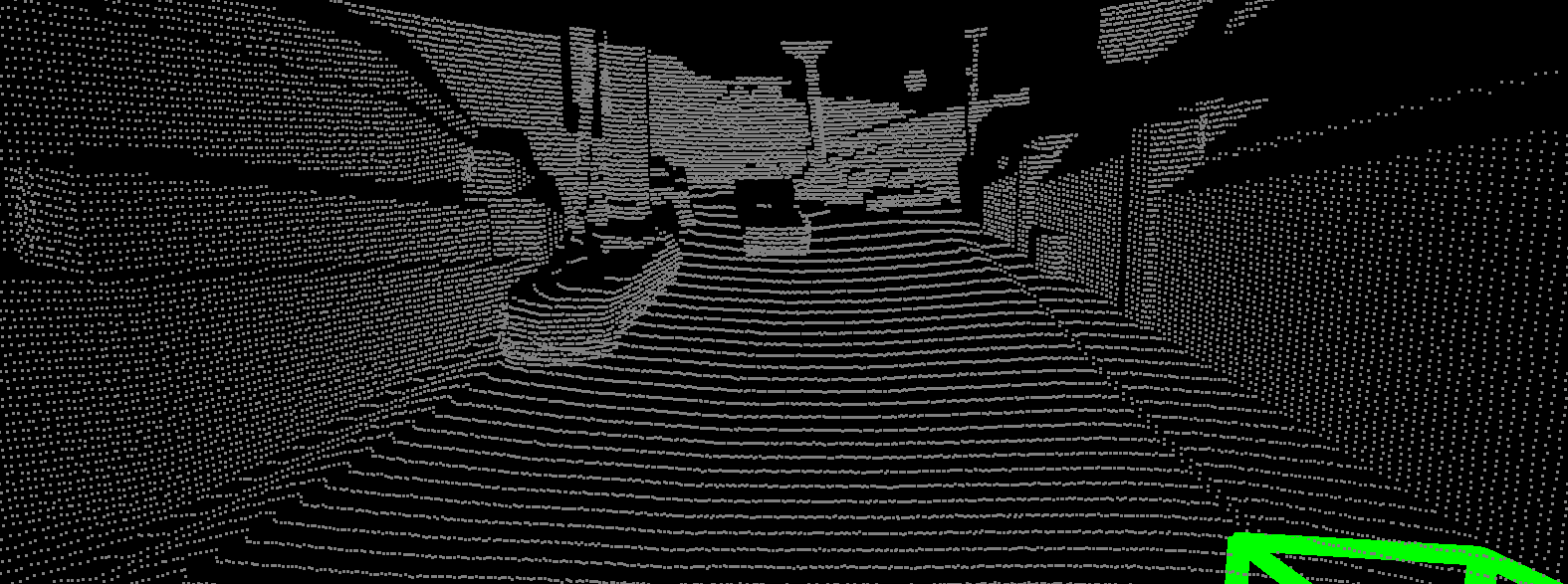}
        \end{minipage}
    }
    \subfloat{%
        \begin{minipage}[]{0.19\linewidth}
            \includegraphics[width=1.0\linewidth]{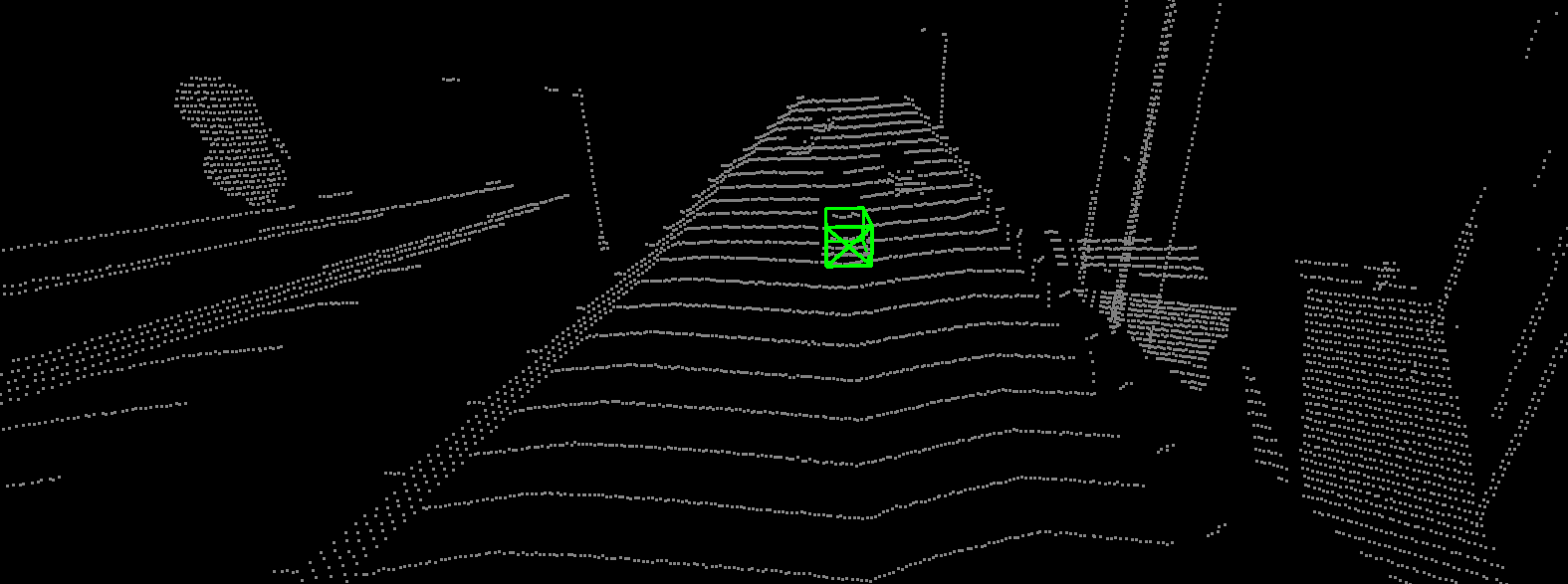}\vspace{2pt}
            \includegraphics[width=1.0\linewidth]{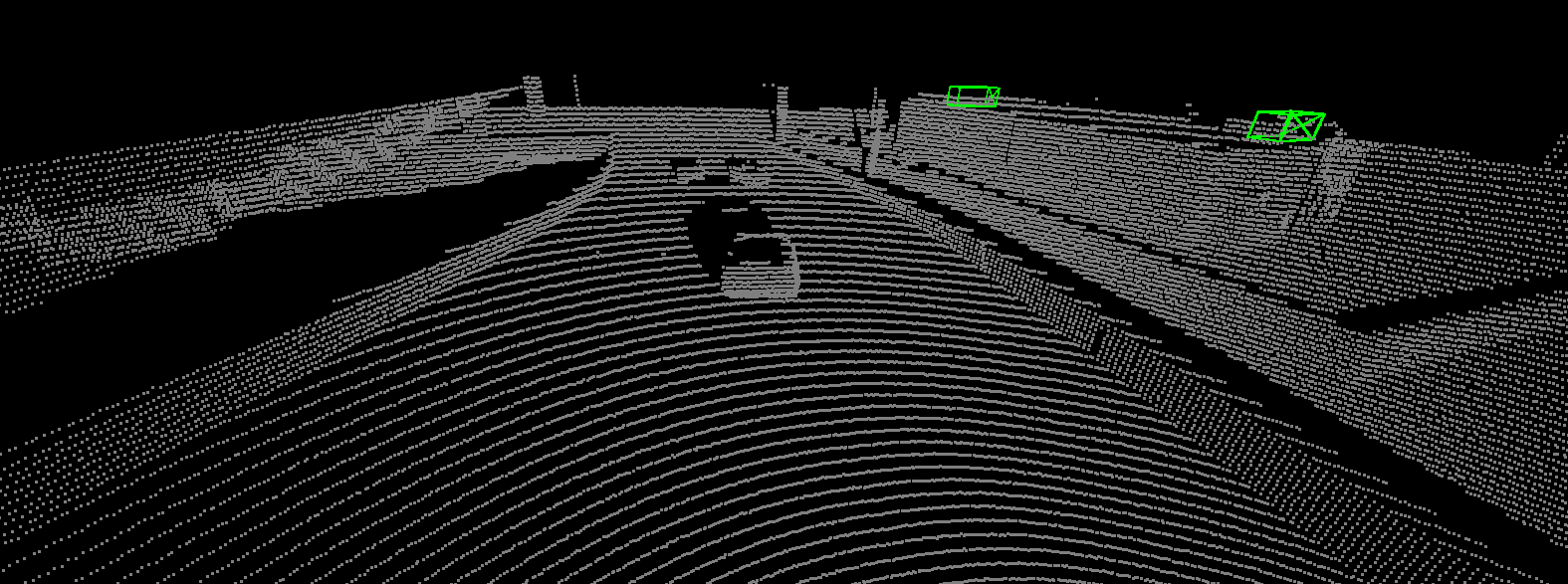}\vspace{2pt}
            \includegraphics[width=1.0\linewidth]{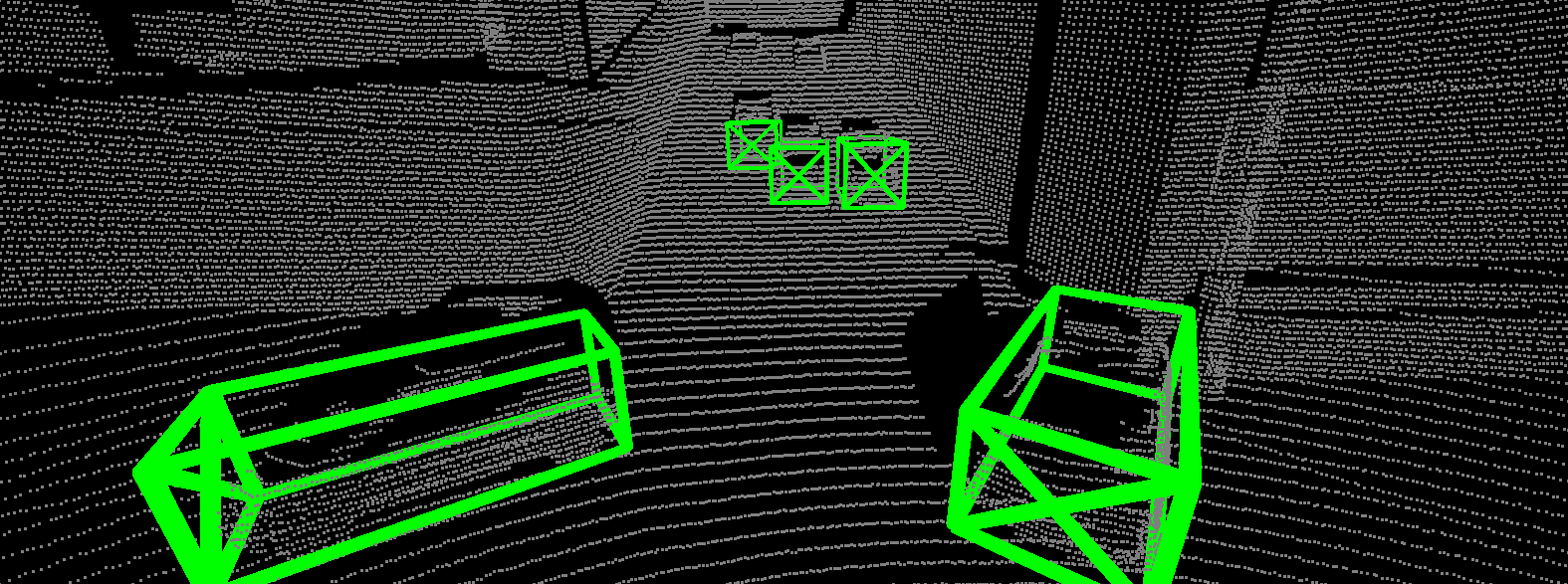}\vspace{2pt}
            \includegraphics[width=1.0\linewidth]{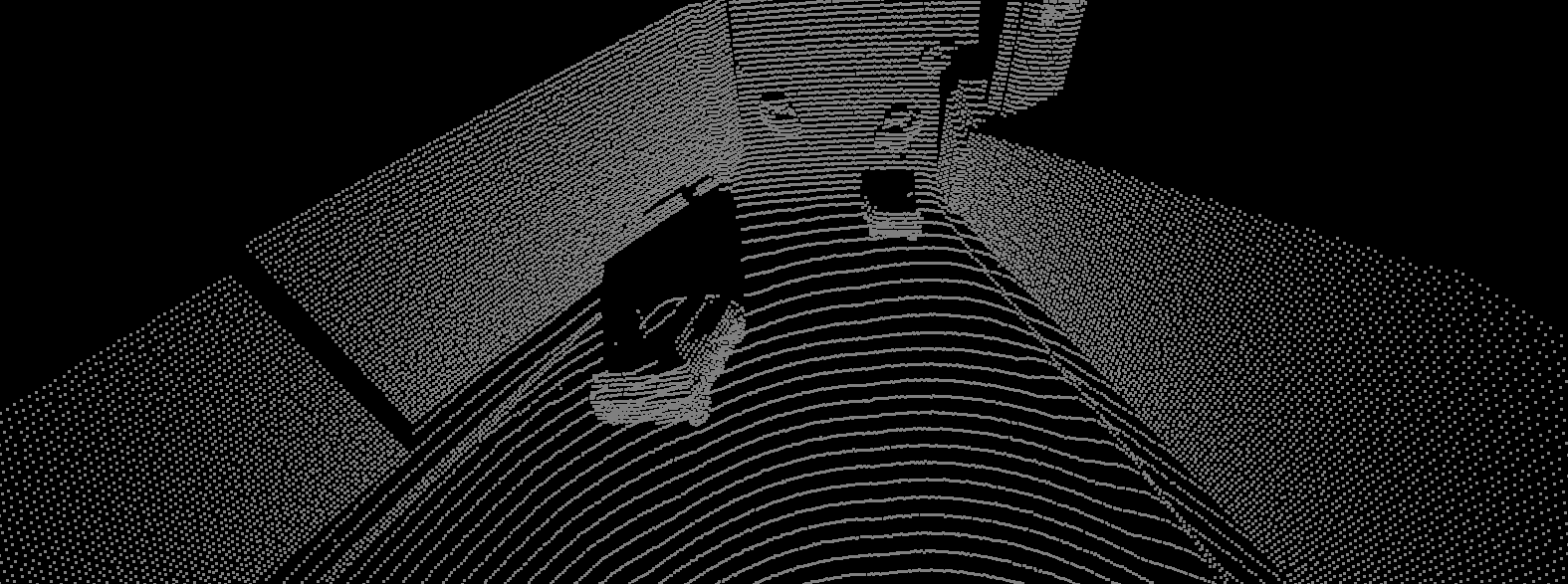}\vspace{2pt}
            \includegraphics[width=1.0\linewidth]{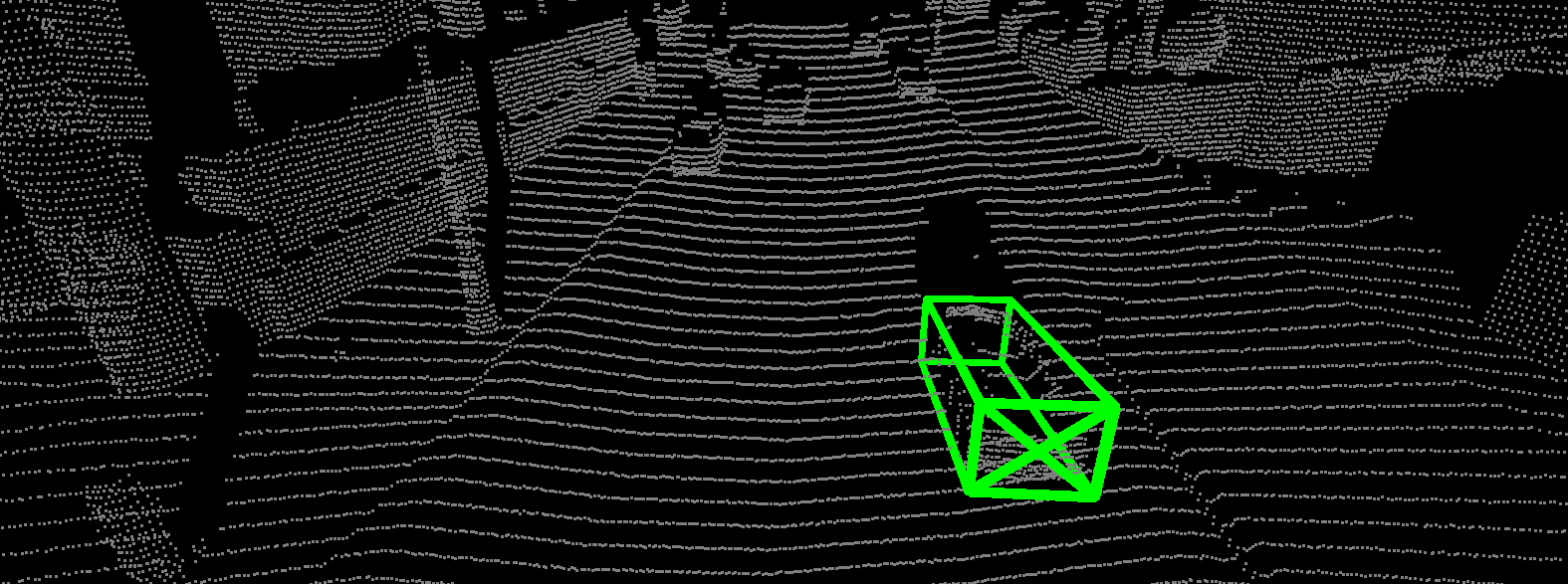}\vspace{2pt}
            \includegraphics[width=1.0\linewidth]{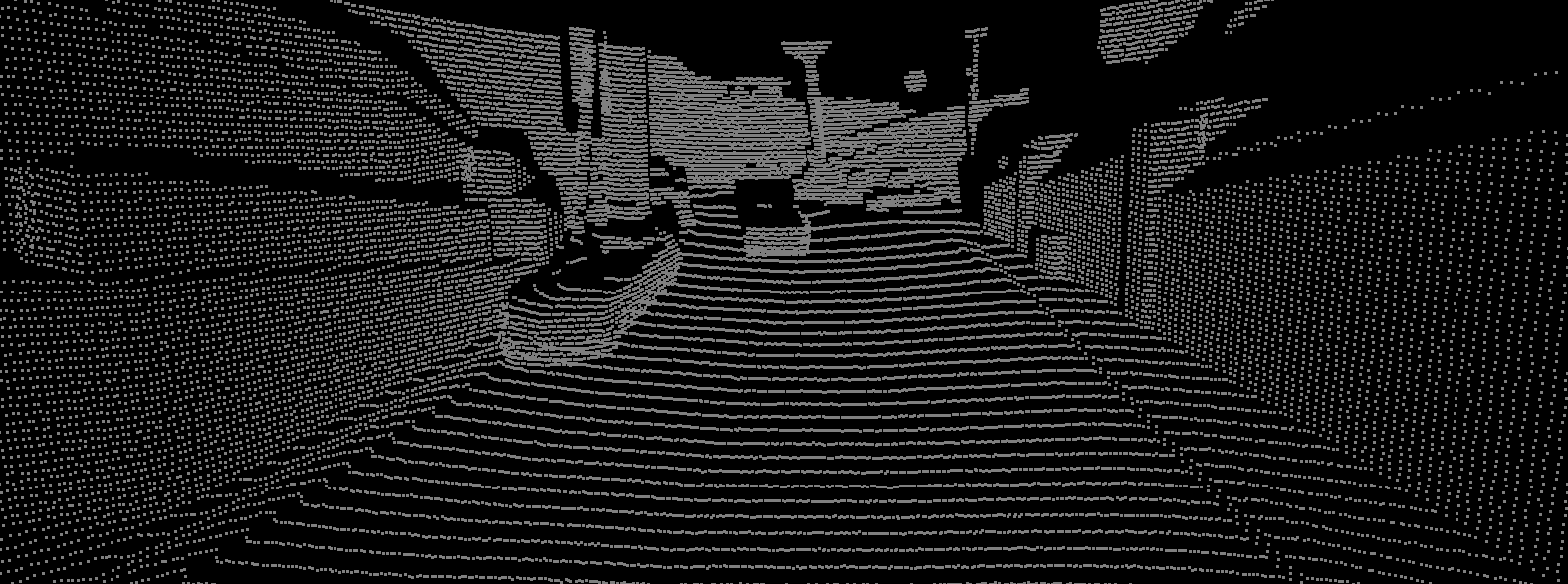}
        \end{minipage}
    }
    \subfloat{%
        \begin{minipage}[]{0.19\linewidth}
            \includegraphics[width=1.0\linewidth]{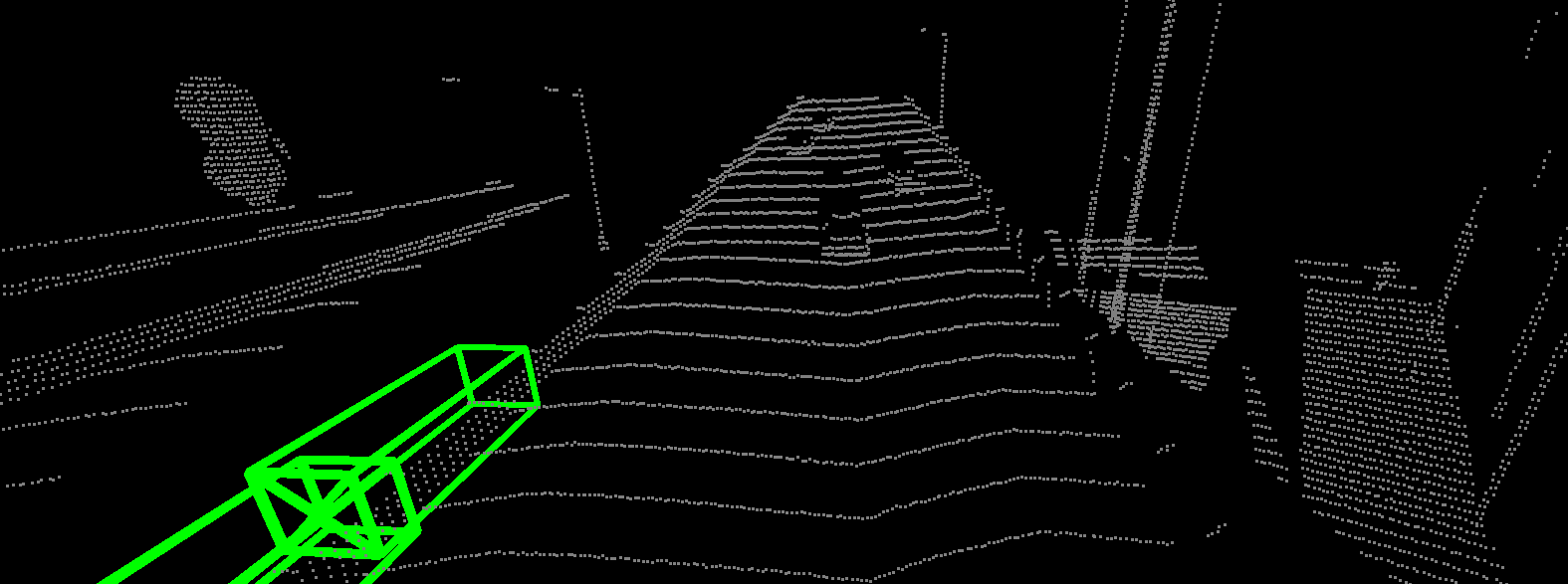}\vspace{2pt}
            \includegraphics[width=1.0\linewidth]{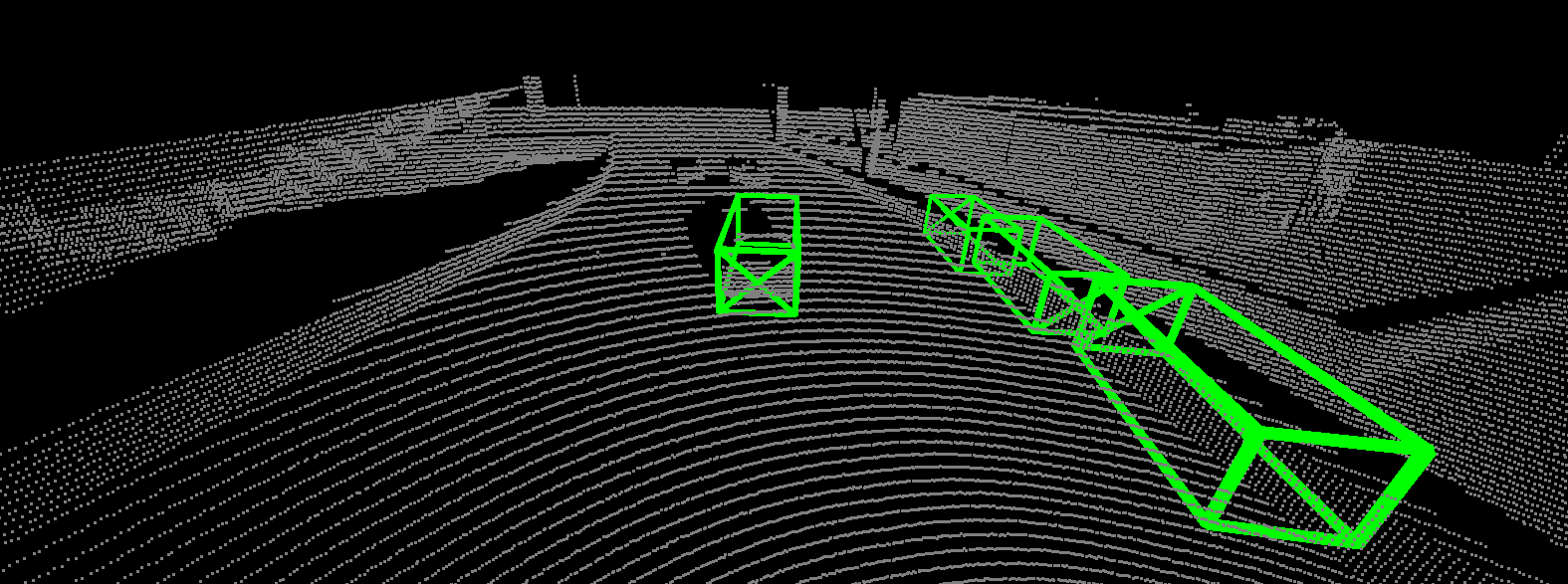}\vspace{2pt}
            \includegraphics[width=1.0\linewidth]{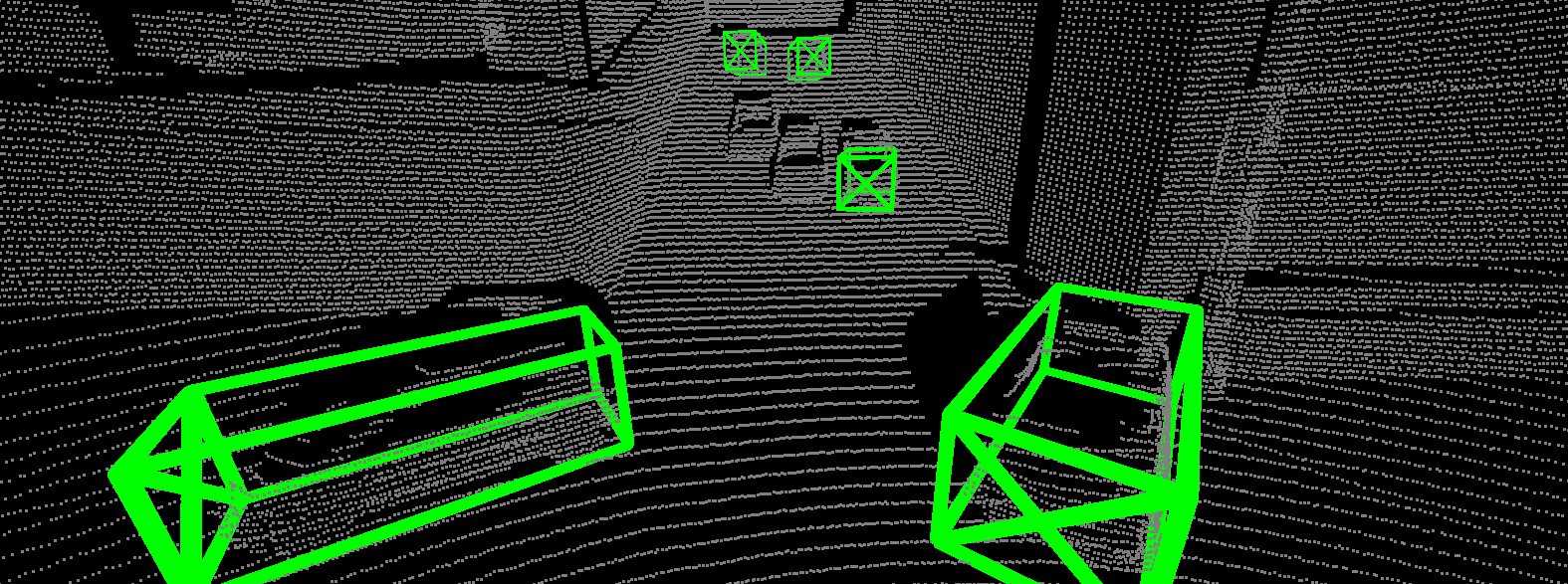}\vspace{2pt}
            \includegraphics[width=1.0\linewidth]{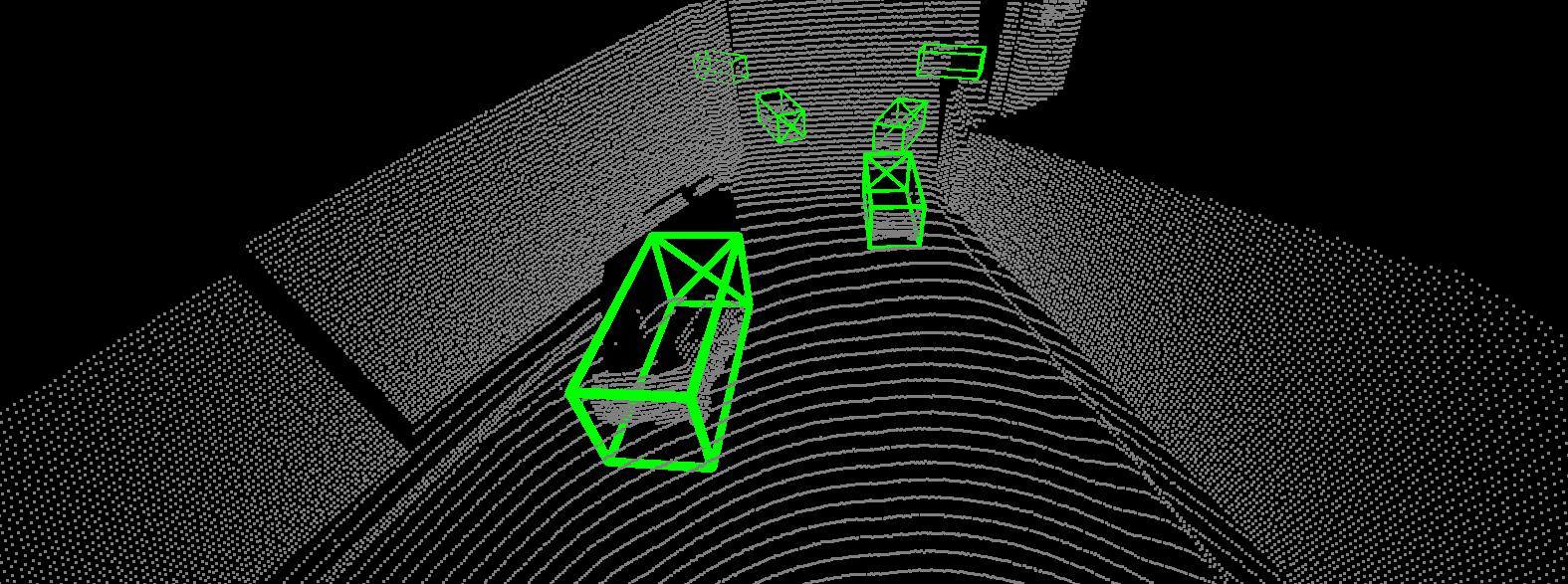}\vspace{2pt}
            \includegraphics[width=1.0\linewidth]{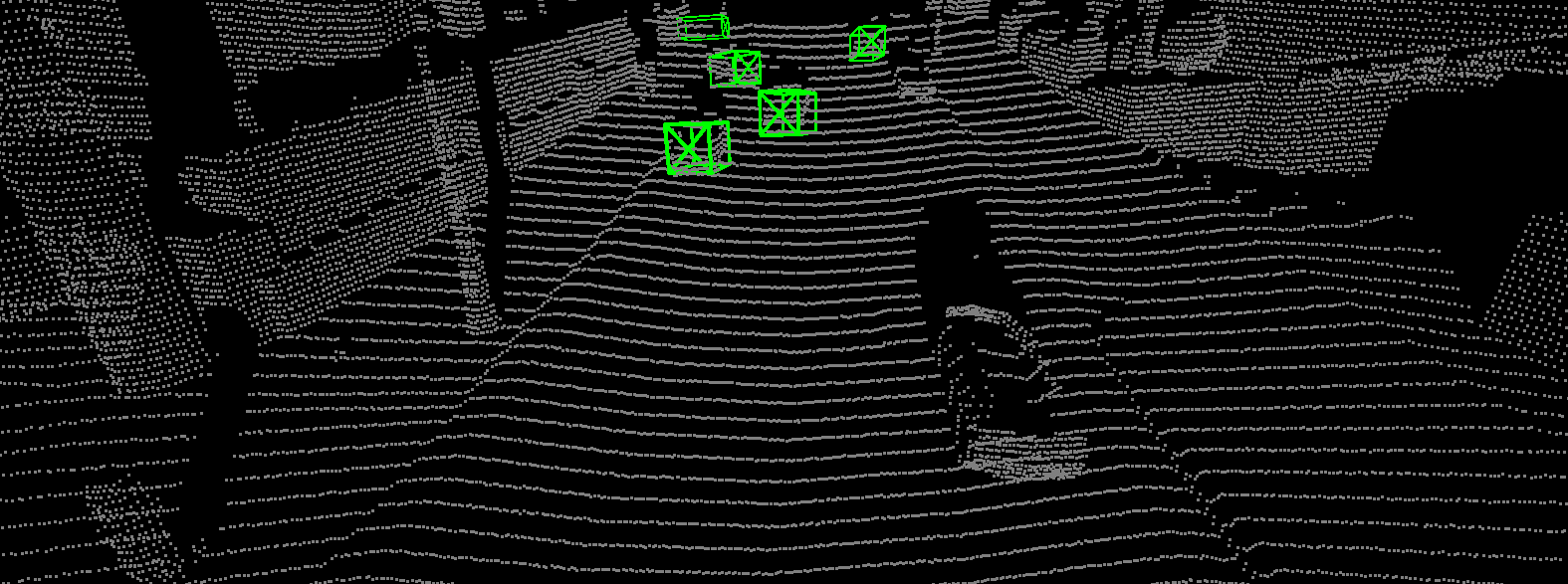}\vspace{2pt}
            \includegraphics[width=1.0\linewidth]{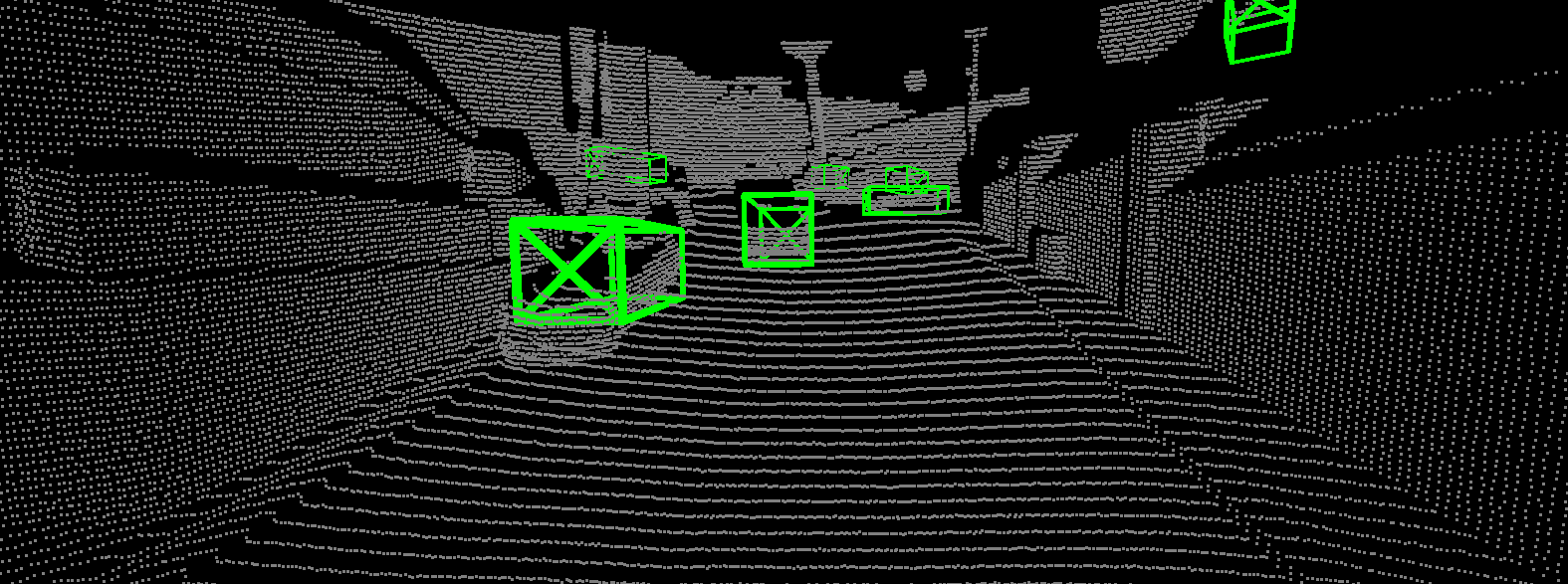}
        \end{minipage}
    }
    \subfloat{%
        \begin{minipage}[]{0.19\linewidth}
            \includegraphics[width=1.0\linewidth]{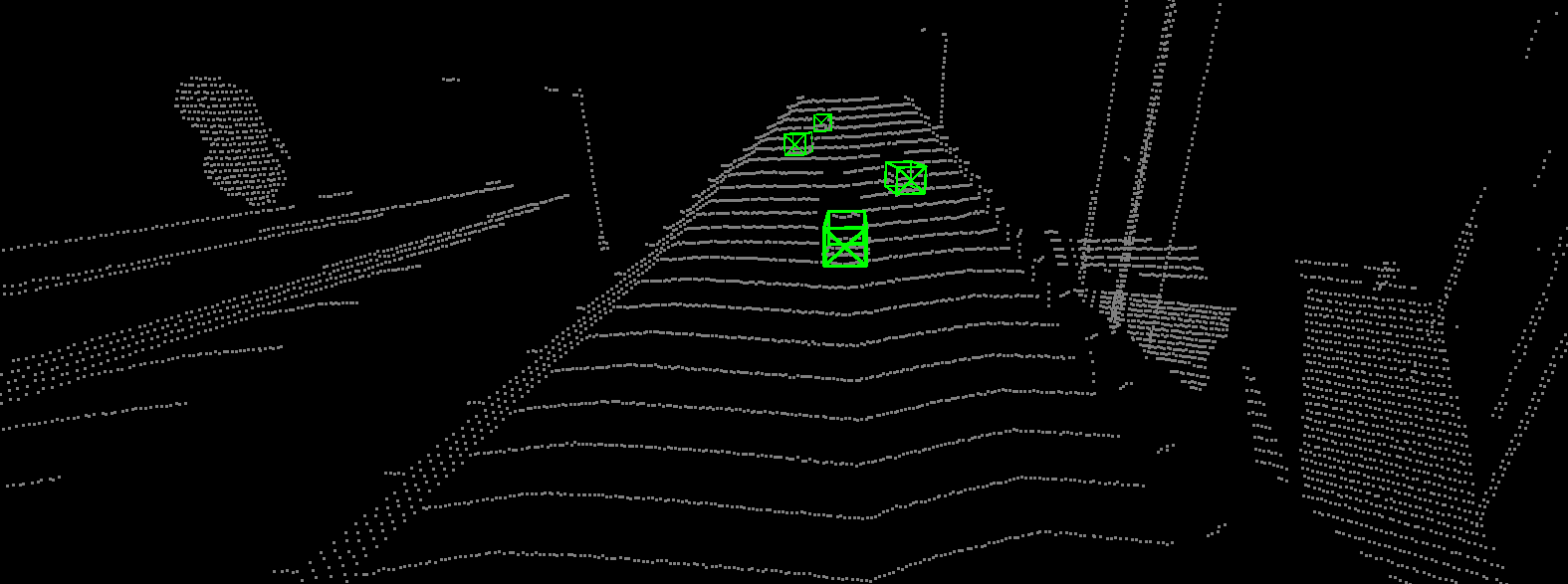}\vspace{2pt}
            \includegraphics[width=1.0\linewidth]{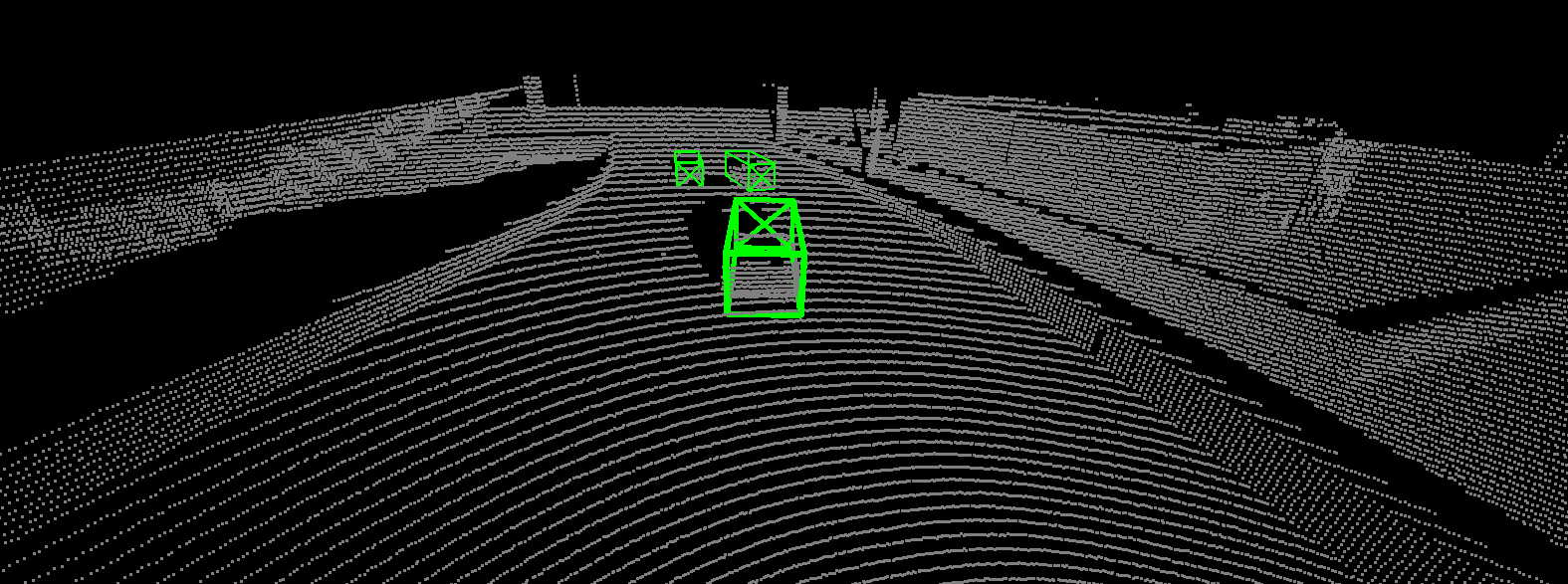}\vspace{2pt}
            \includegraphics[width=1.0\linewidth]{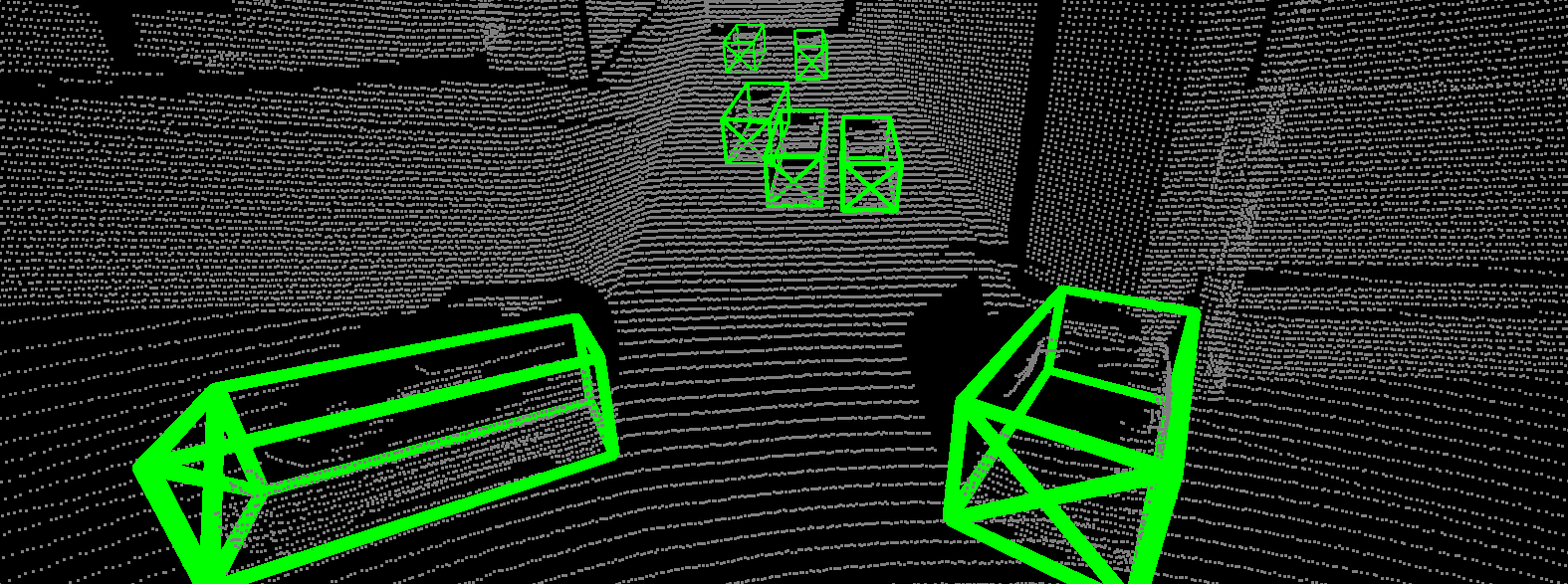}\vspace{2pt}
            \includegraphics[width=1.0\linewidth]{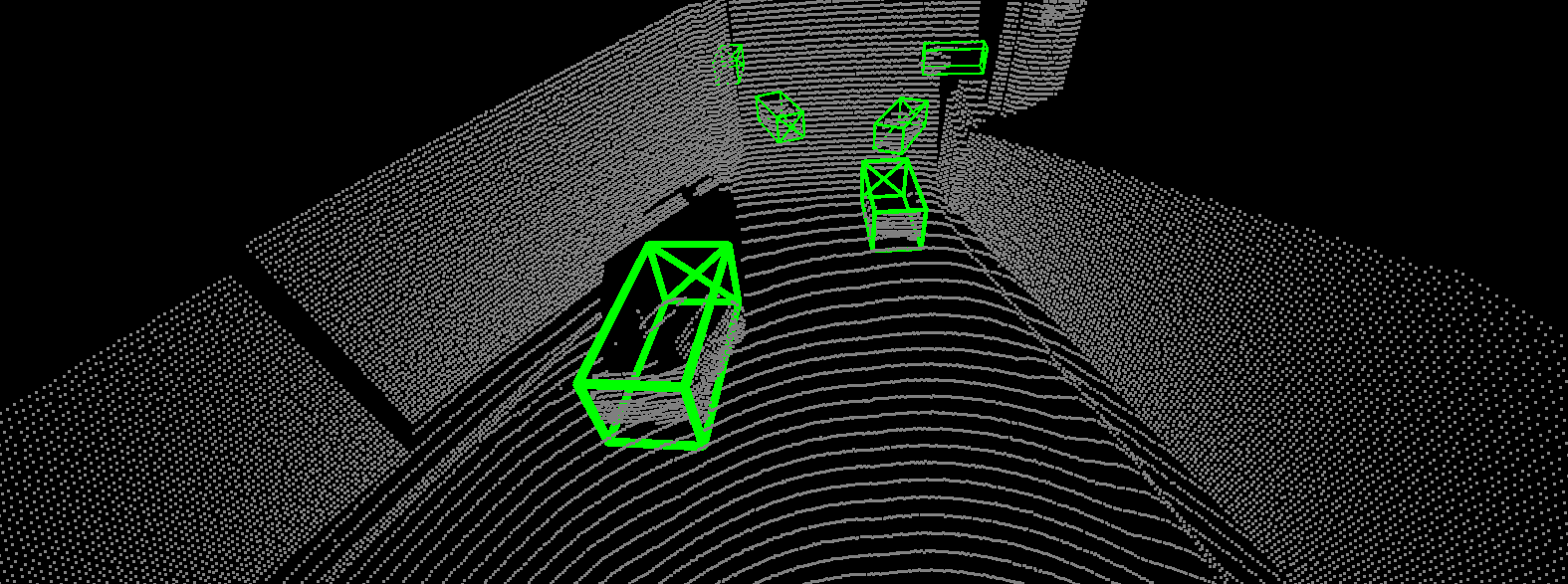}\vspace{2pt}
            \includegraphics[width=1.0\linewidth]{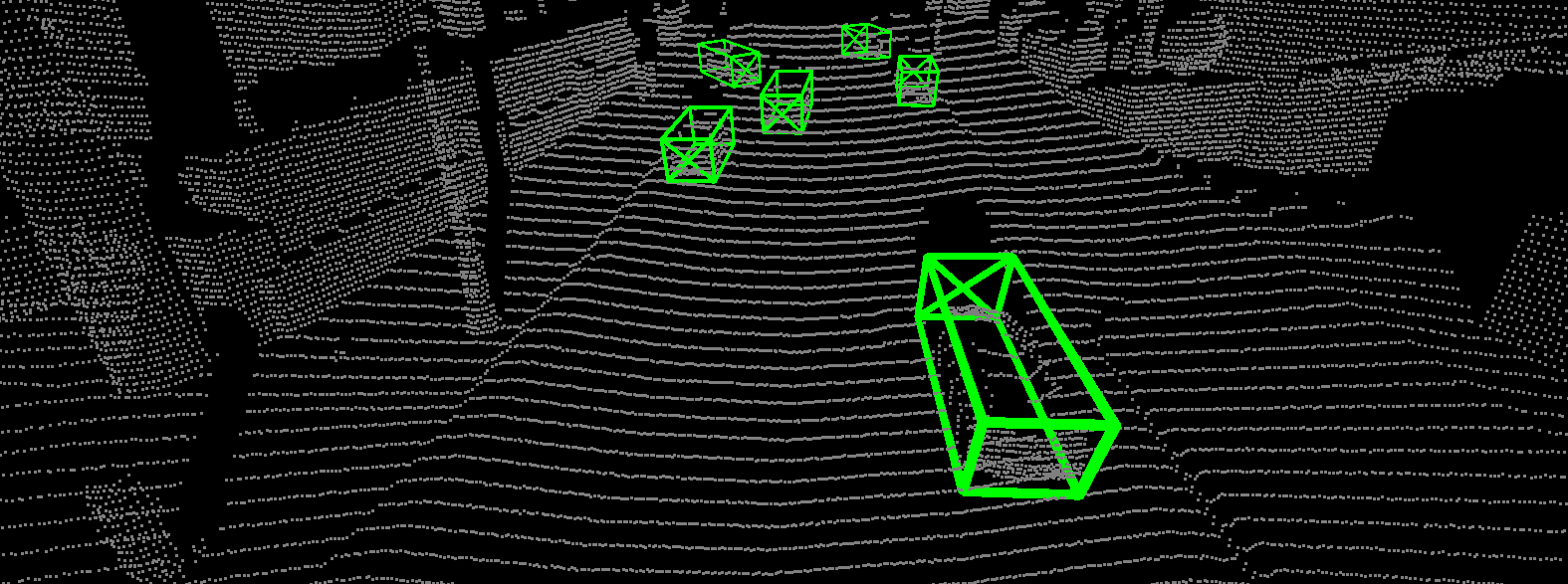}\vspace{2pt}
            \includegraphics[width=1.0\linewidth]{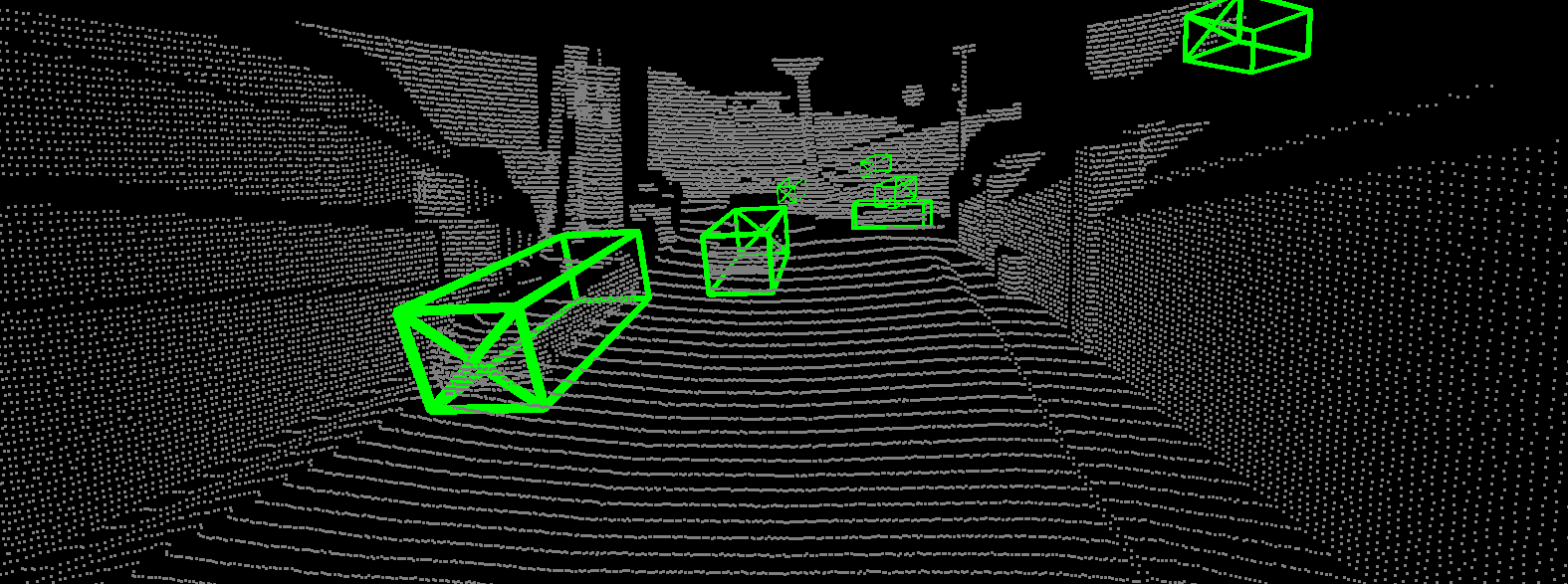}
        \end{minipage}
    }
    \subfloat{%
        \begin{minipage}[]{0\linewidth}
            \rotatebox{90}{\tiny Scene 6 \hspace{0pt} Scene 5 \hspace{0pt} Scene 4 \hspace{0pt} Scene 3 \hspace{0pt} scene 2 \hspace{0pt} Scene 1}
        \end{minipage}
    }
    \\
    \centering\hspace{-5pt}
    \subfloat{%
        \setcounter{subfigure}{0}
        \begin{minipage}[]{0.005\linewidth}
            \rotatebox{90}{\tiny GAZEBO}
        \end{minipage}
    }
    \subfloat[]{%
        \begin{minipage}[]{0.1322\linewidth}
            \includegraphics[width=1.0\linewidth]{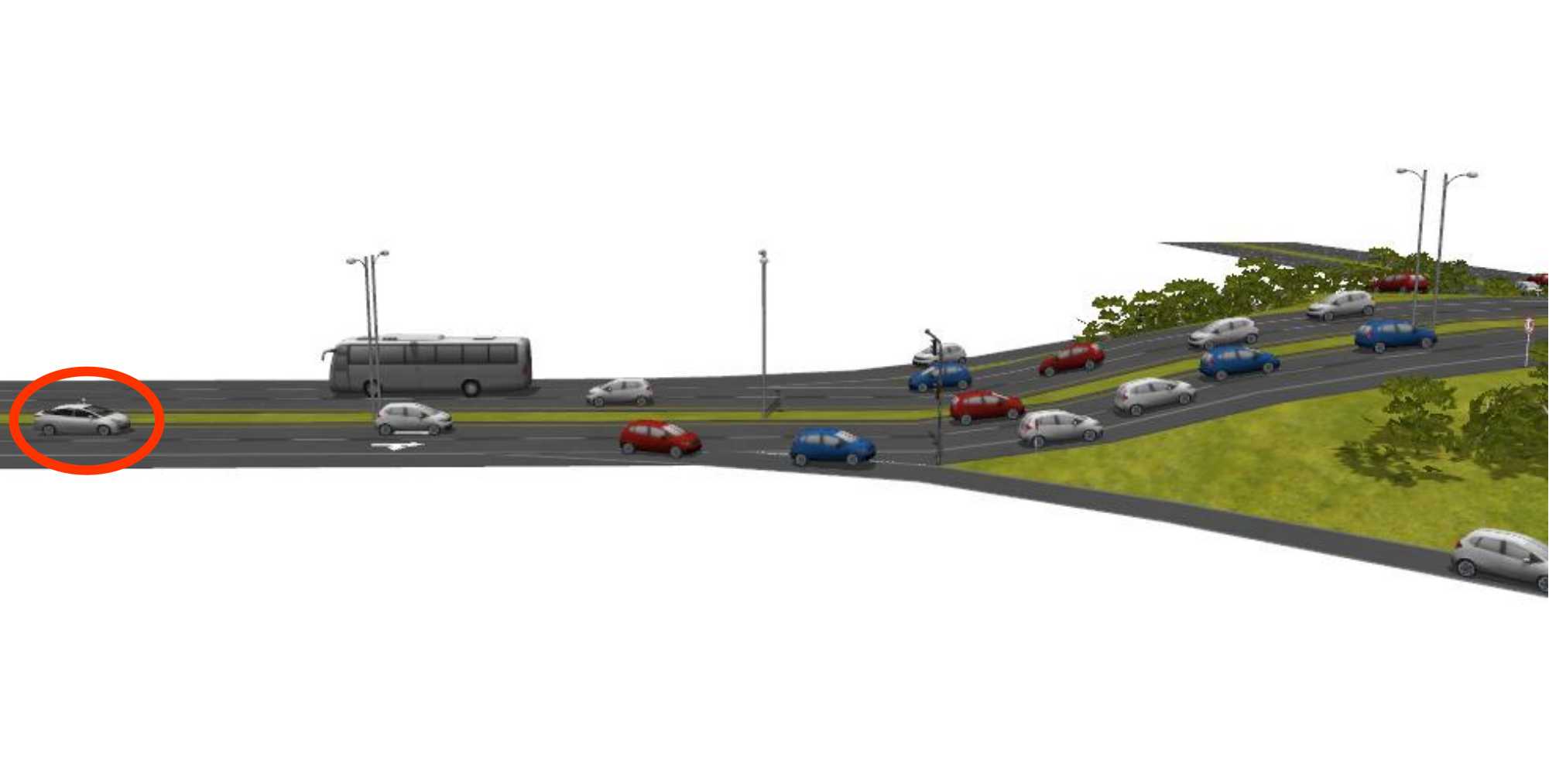}\vspace{13.8pt}
            \includegraphics[width=1.0\linewidth]{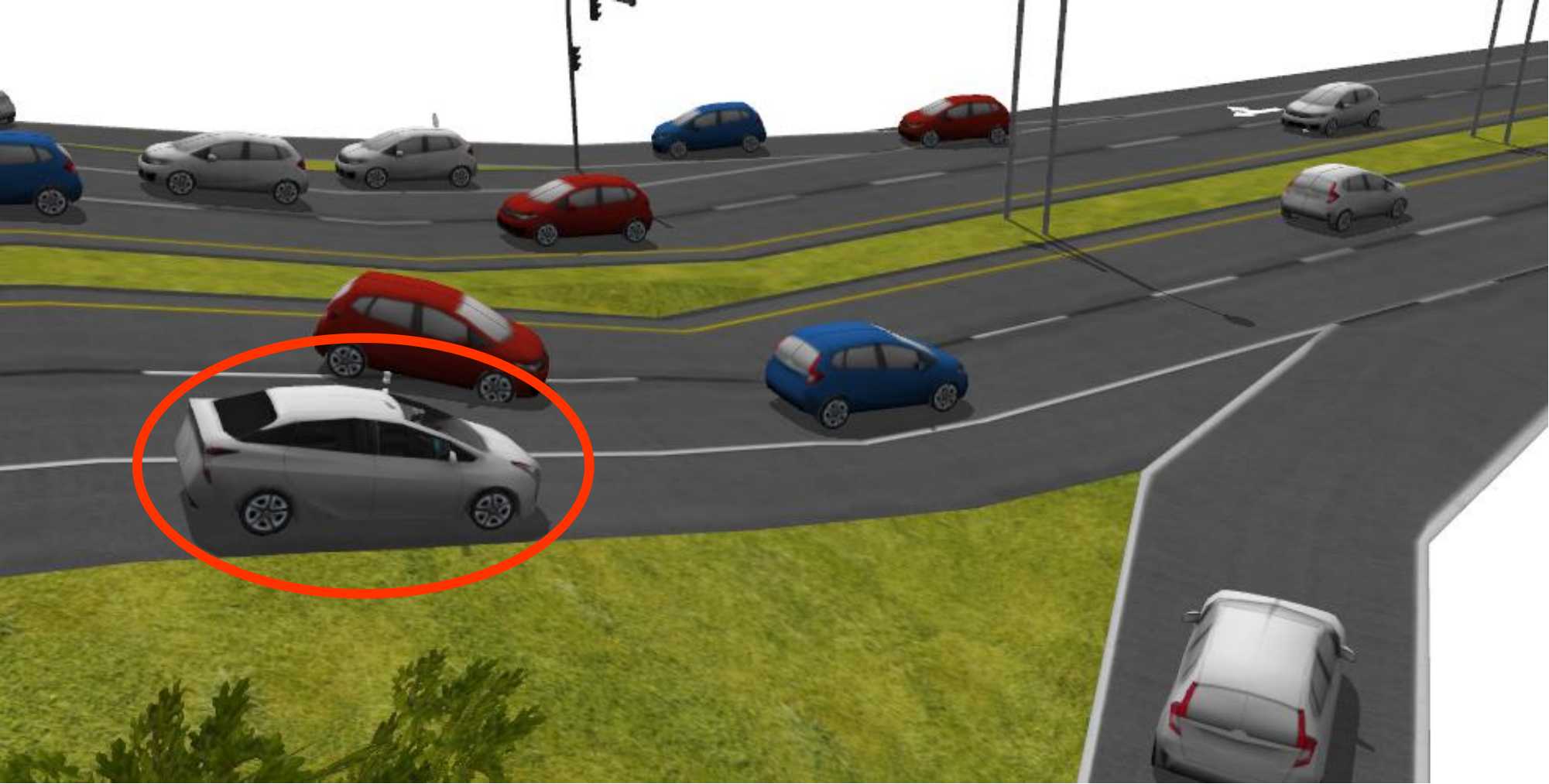}\vspace{13.8pt}
            \includegraphics[width=1.0\linewidth]{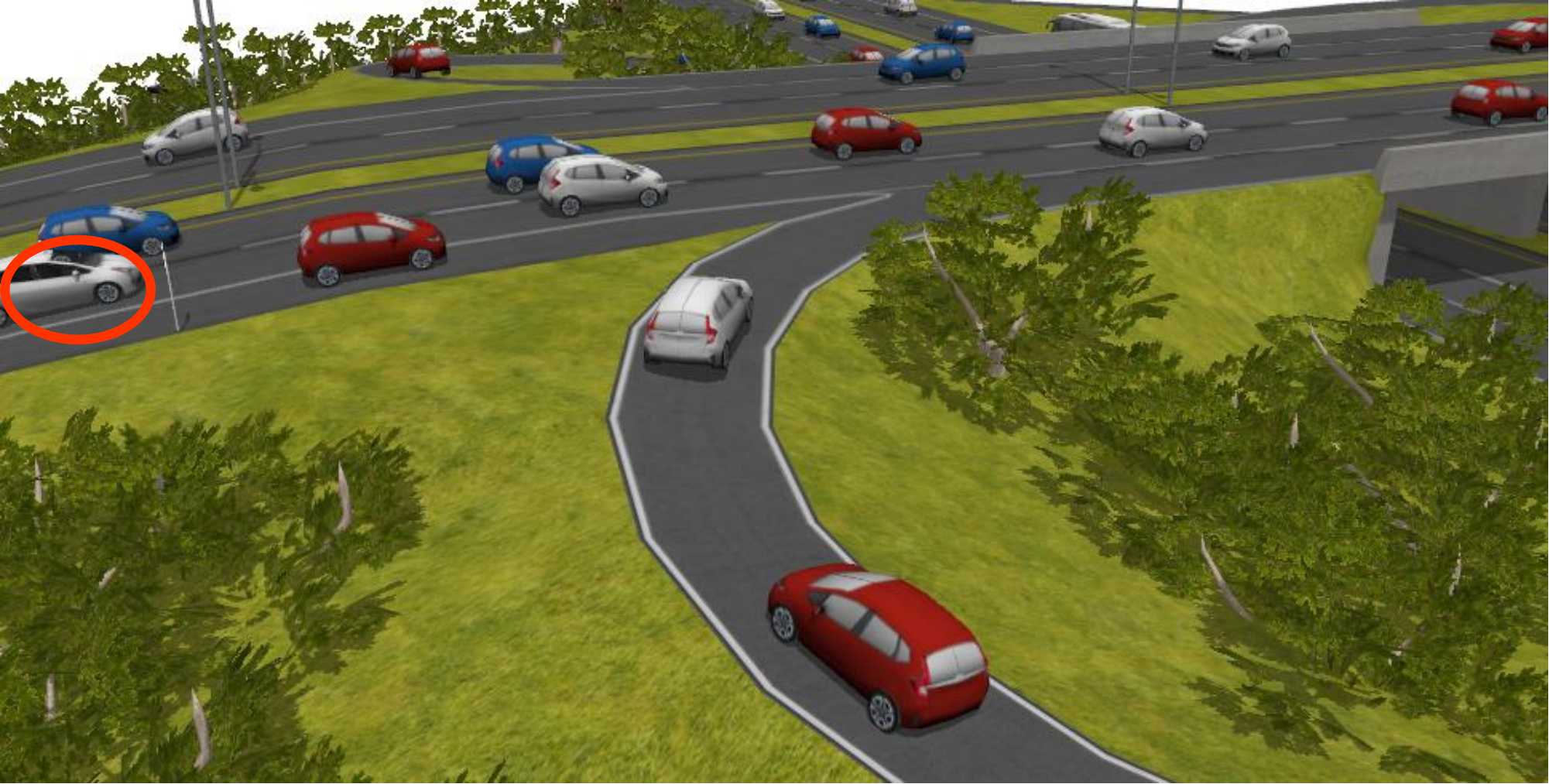}
        \end{minipage}
    }
    \subfloat[]{%
        \begin{minipage}[]{0.19\linewidth}
            \includegraphics[width=1.0\linewidth]{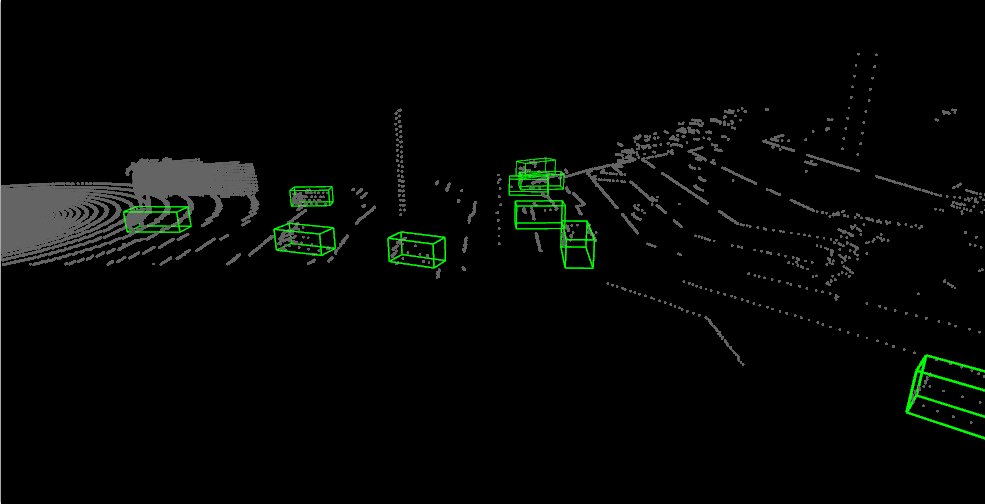}\vspace{2pt}
            \includegraphics[width=1.0\linewidth]{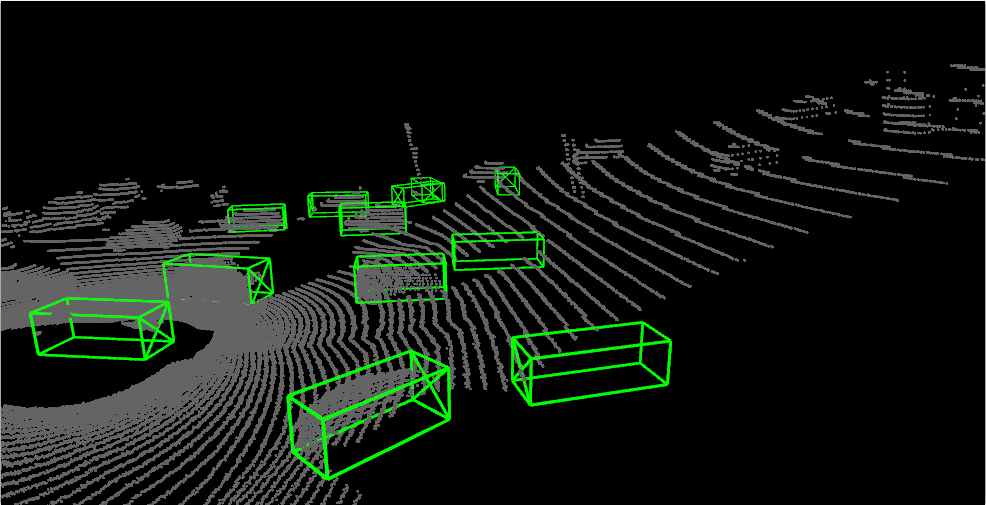}\vspace{2pt}
            \includegraphics[width=1.0\linewidth]{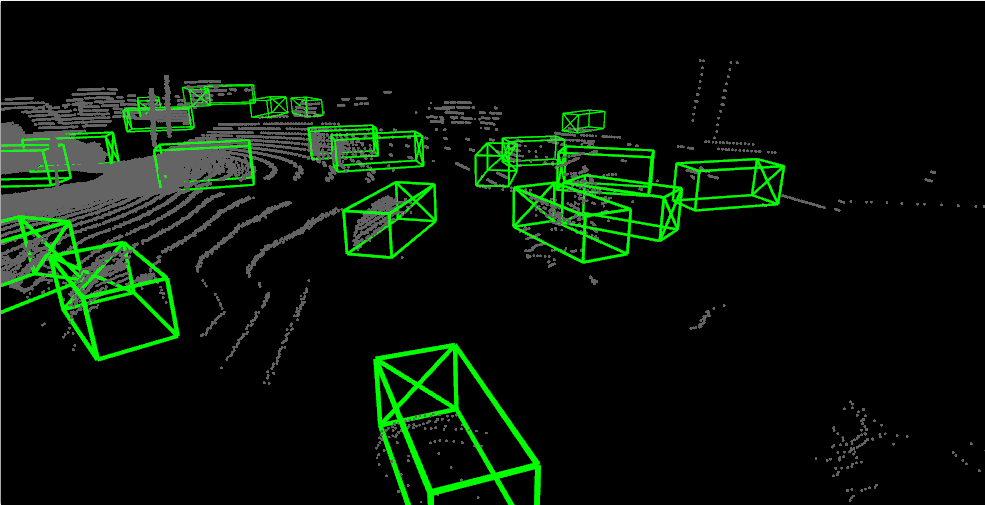}
        \end{minipage}
    }
    \subfloat[]{%
        \begin{minipage}[]{0.19\linewidth}
            \includegraphics[width=1.0\linewidth]{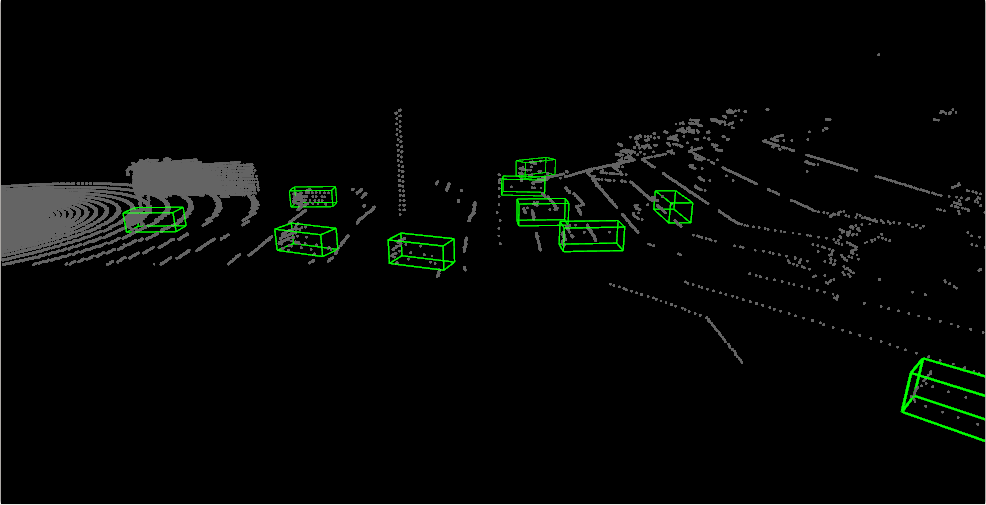}\vspace{2pt}
            \includegraphics[width=1.0\linewidth]{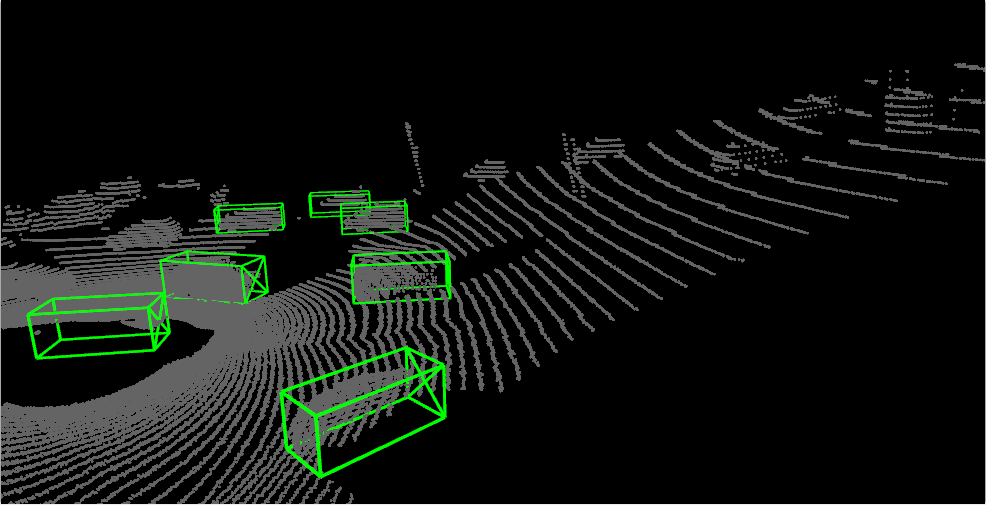}\vspace{2pt}
            \includegraphics[width=1.0\linewidth]{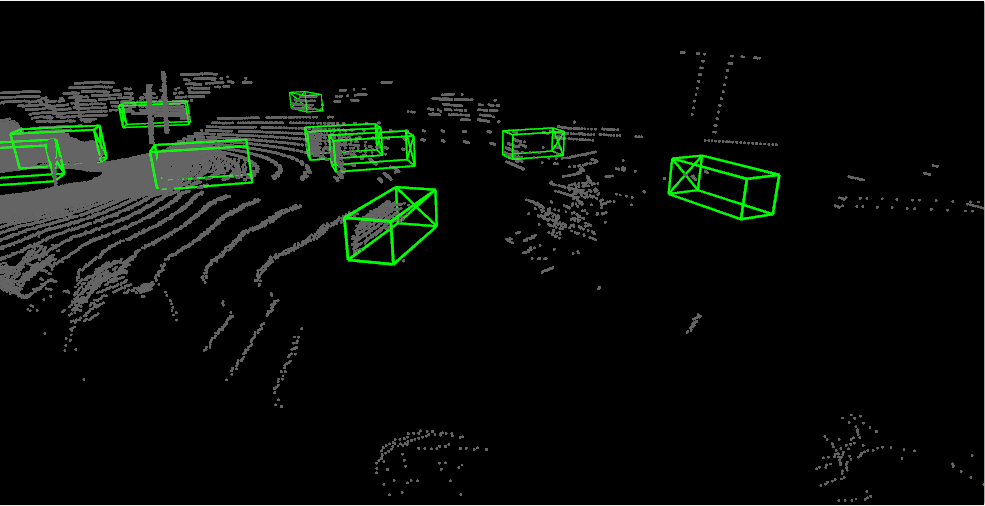}
        \end{minipage}
    }
    \subfloat[]{%
        \begin{minipage}[]{0.19\linewidth}
            \includegraphics[width=1.0\linewidth]{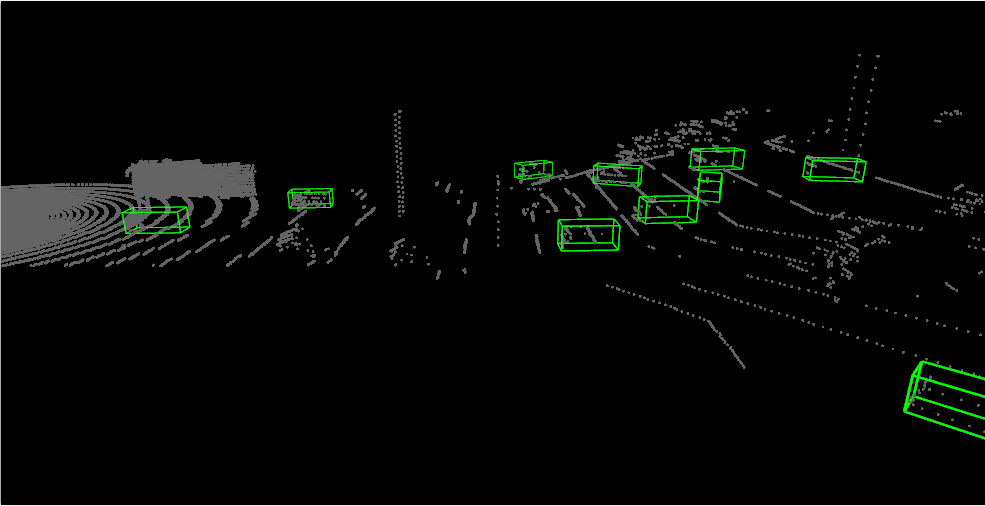}\vspace{2pt}
            \includegraphics[width=1.0\linewidth]{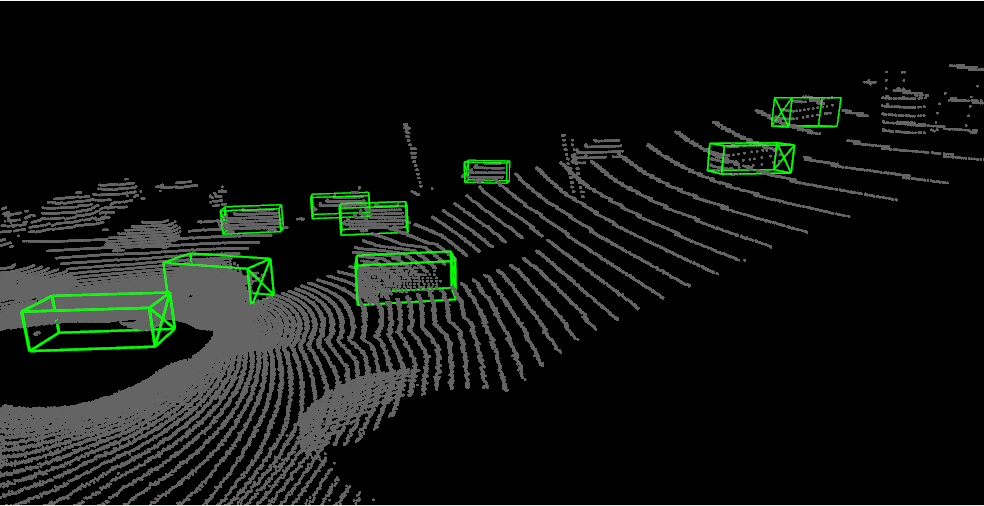}\vspace{2pt}
            \includegraphics[width=1.0\linewidth]{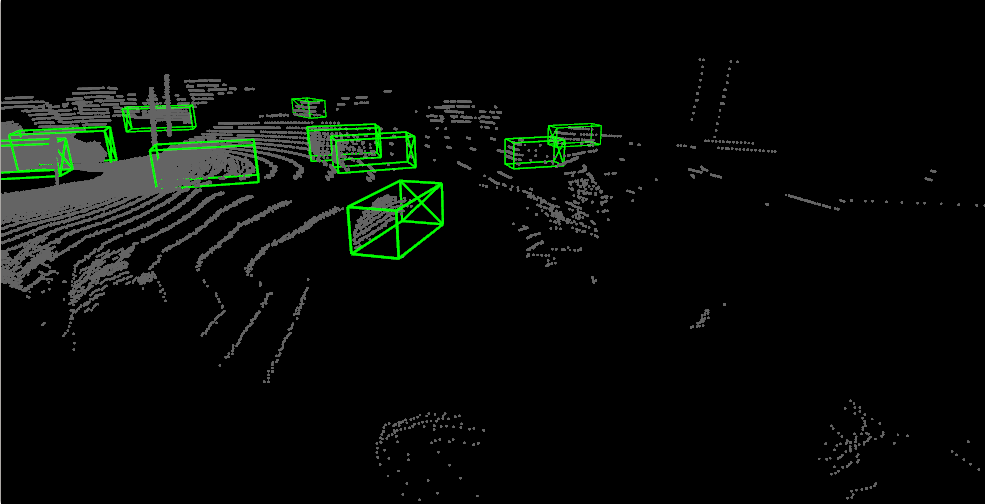}
        \end{minipage}
    }
    \subfloat[]{%
        \begin{minipage}[]{0.19\linewidth}
            \includegraphics[width=1.0\linewidth]{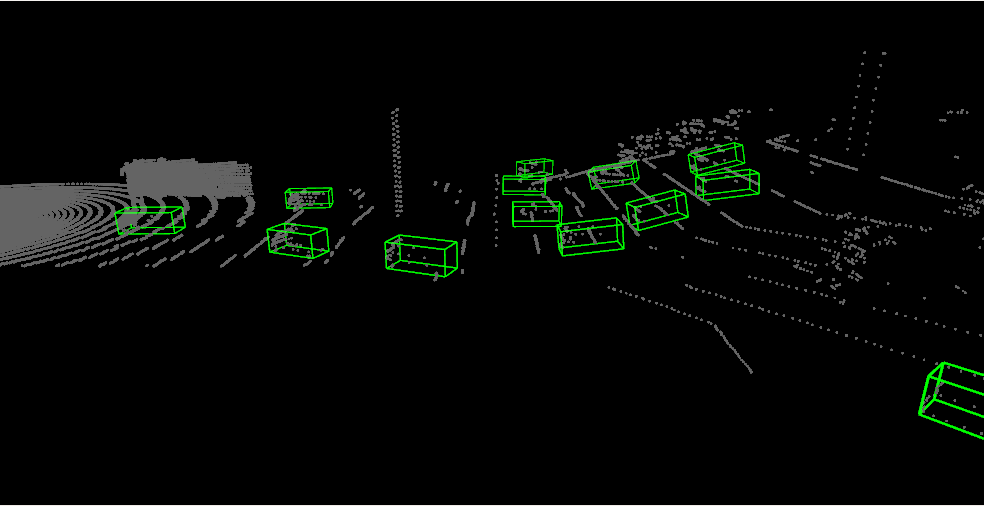}\vspace{2pt}
            \includegraphics[width=1.0\linewidth]{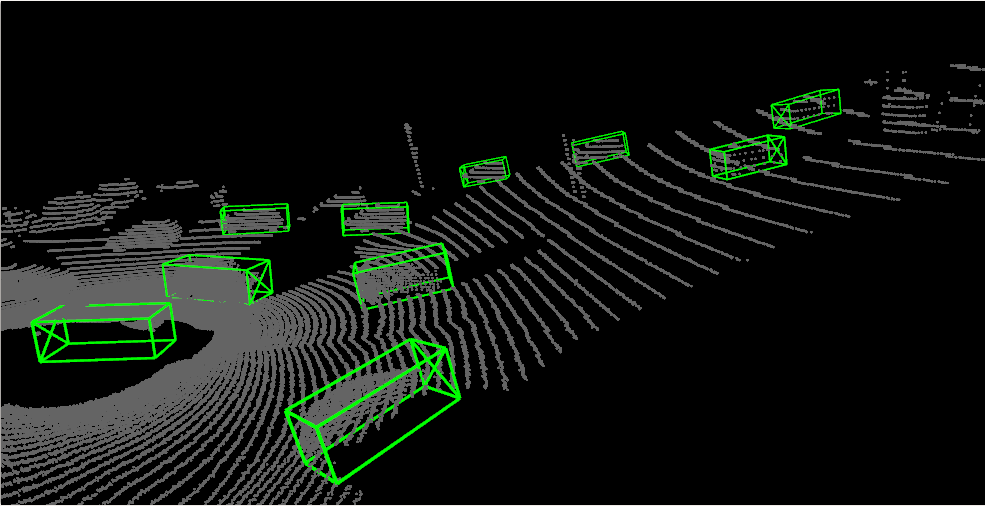}\vspace{2pt}
            \includegraphics[width=1.0\linewidth]{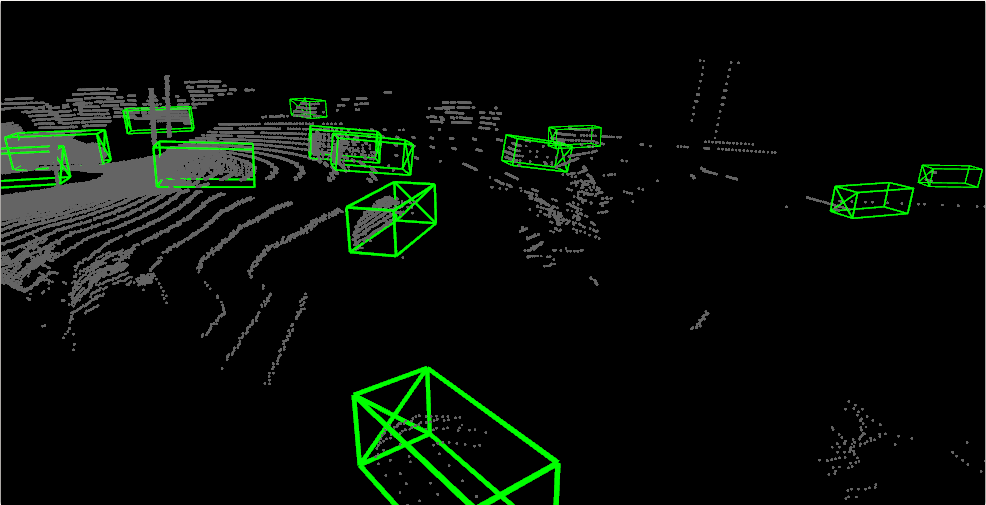}
        \end{minipage}
    }
    \subfloat{%
        \begin{minipage}[]{0\linewidth}
            \rotatebox{90}{\tiny Scene 9 \hspace{7pt} scene 8 \hspace{7pt} Scene 7}
        \end{minipage}
    }
    \caption{
    Visualization for qualitative comparisons. 
    (a) Scenes of GTA-V (first six rows, red circles mean targets) and GAZEBO (last three rows, red circle mean ego vehicle).
    (b) PointPillars\cite{pointpillars}. (c) Voxel R-CNN\cite{deng2020voxel}. (d) PointRCNN\cite{shi2019pointrcnn}. (e) Det6D(Ours). Green boxes are predicted results. }
    \label{fig:gtav_viz}
\end{figure}

%% file: _exp__fig_gazebo2.tex
\begin{figure}
    \centering
    \subfloat[car on sloped terrain\label{fig:exp_gazebo_2:1}]{
        \begin{minipage}[]{0.4\linewidth}
            \begin{minipage}[]{0.45\linewidth}
                \includegraphics[width=1\linewidth]{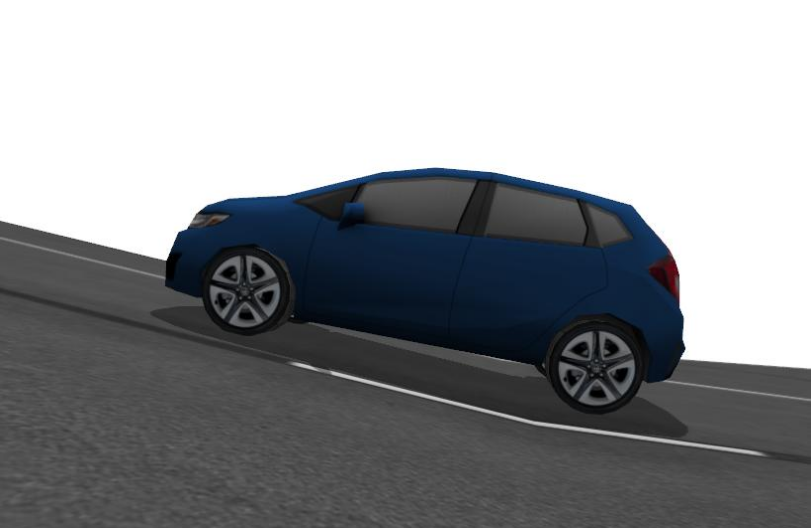}\vspace{2pt}
                \includegraphics[width=1\linewidth]{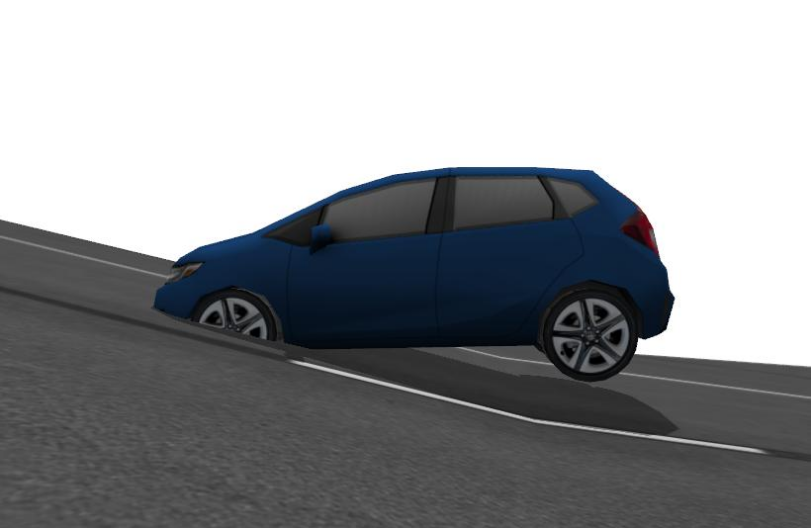}\vspace{2pt}
                \includegraphics[width=1\linewidth]{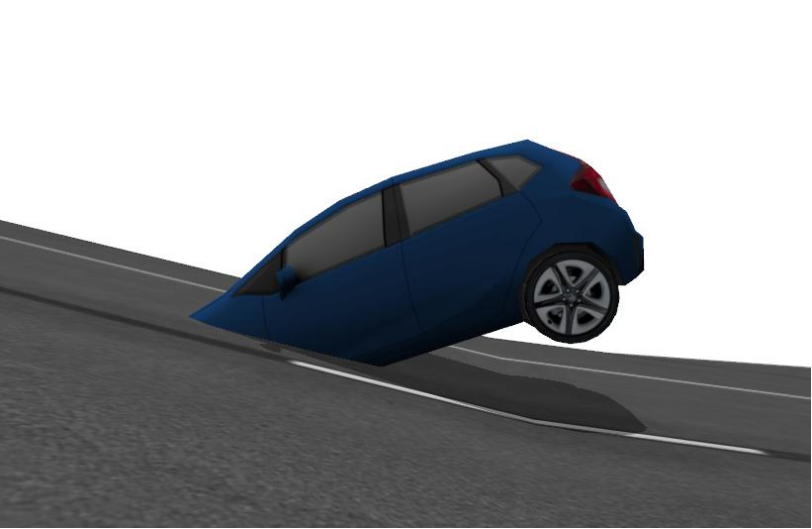}
            \end{minipage}
            \begin{minipage}[]{0.45\linewidth}
                \includegraphics[width=1\linewidth]{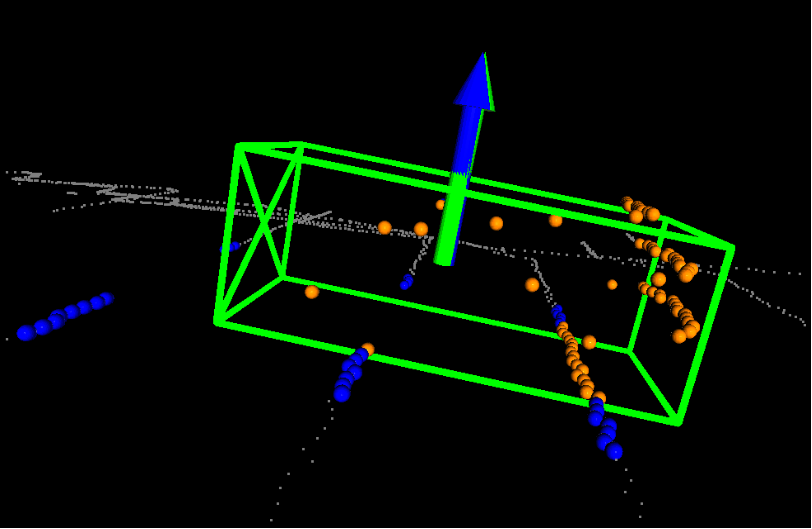}\vspace{2pt}
                \includegraphics[width=1\linewidth]{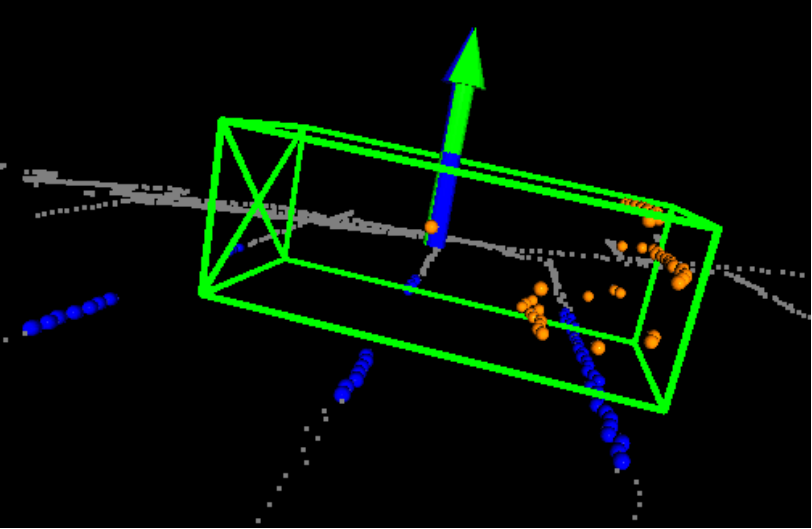}\vspace{2pt}
                \includegraphics[width=1\linewidth]{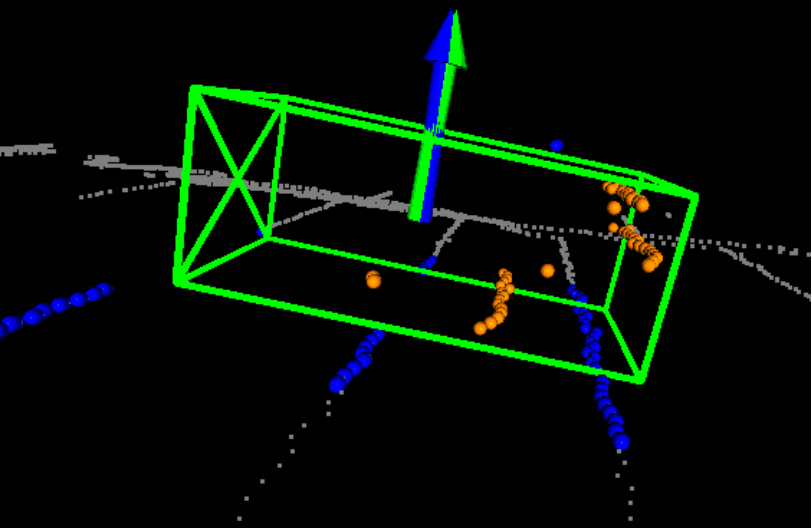}
            \end{minipage}    
        \end{minipage}
    }
    \subfloat[car on flat terrain\label{fig:exp_gazebo_2:2}]{
    \begin{minipage}[]{0.4\linewidth}
        \begin{minipage}[]{0.45\linewidth}
            \includegraphics[width=1\linewidth]{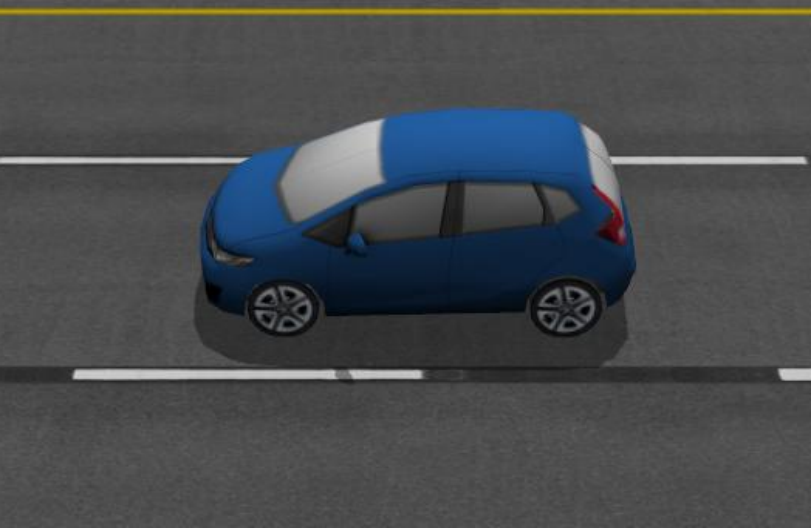}\vspace{2pt}
            \includegraphics[width=1\linewidth]{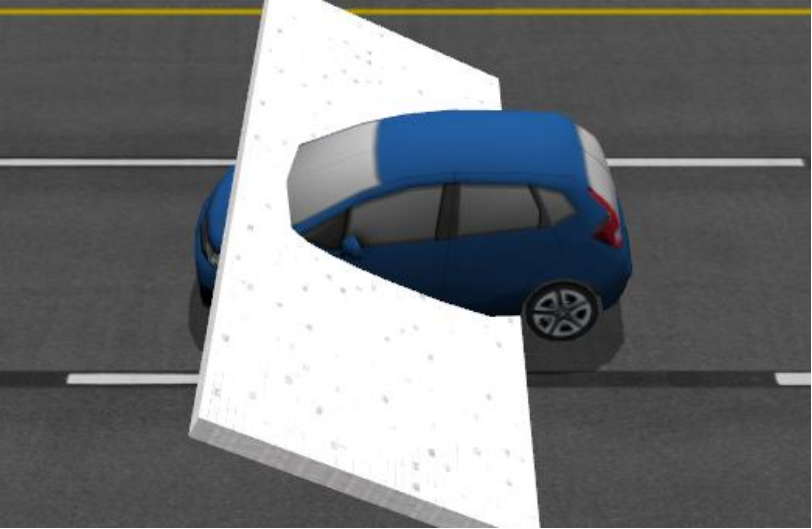}\vspace{2pt}
            \includegraphics[width=1\linewidth]{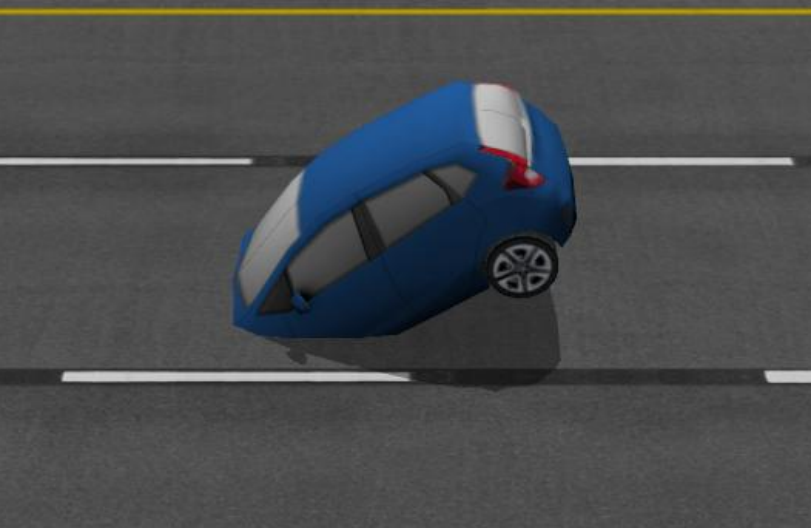}
        \end{minipage}
        \begin{minipage}[]{0.45\linewidth}
            \includegraphics[width=1\linewidth]{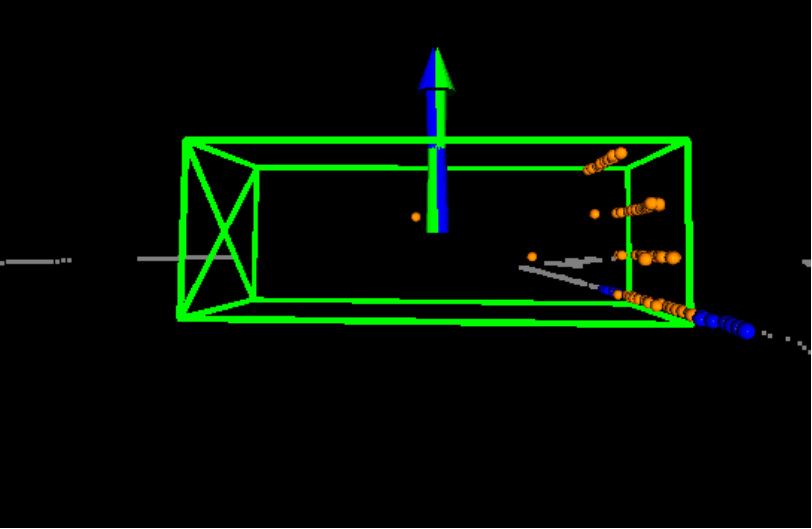}\vspace{2pt}
            \includegraphics[width=1\linewidth]{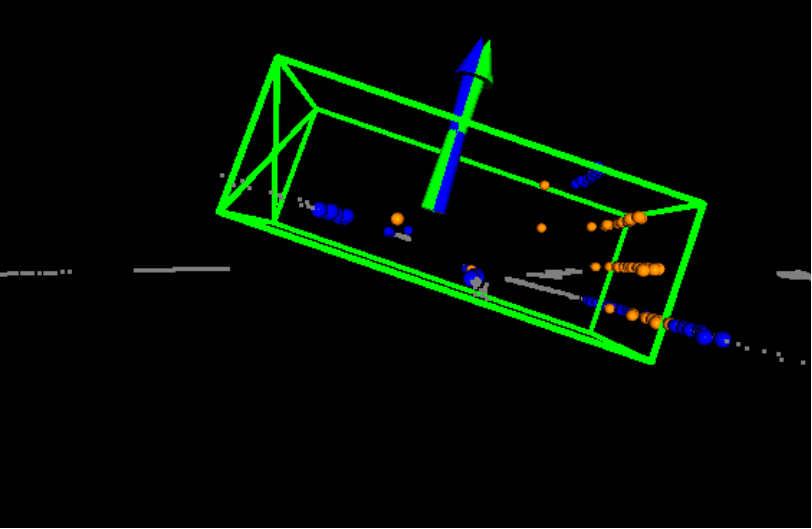}\vspace{2pt}
            \includegraphics[width=1\linewidth]{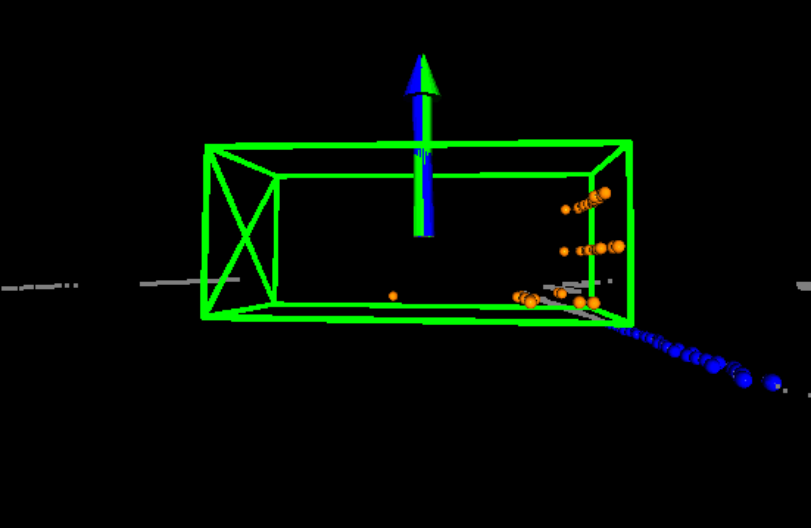}
        \end{minipage}    
    \end{minipage}
    }
\caption{
Visualization of extended experiment results in GAZEBO. The green box is the detection result. 
The green and blue arrows are the object's z-axis and the ground normal.
The orange and blue points are foreground points of objects and ground points near objects, respectively.
}
\label{fig:exp_gazebo_2}
\end{figure} 

%% file: conclusion.tex
\section{CONCLUSION}
This work introduces a new method, Det6D, the first full-space full-pose 3D object detector. 
It well addresses the performance degradation of existing methods in non-flat scenes and improves the perception robustness in autonomous driving. 

We first uncover that point-based methods have full-space detection capabilities.
Then we analyze the essential differences between different poses and propose a ground-aware orientation head following a ground segmentation module to predict full poses accurately.
Finally, we propose Slope-Aug to utilize existing datasets to synthesize non-flat scenes.
Experiments show that Det6D keeps performance in flat scenes while excelling in non-flat scenes.
For future work, we hope to address the limitations of the voxel-based method in full-space full-pose detection.
